\documentclass[11pt]{article}

\usepackage{apacite}
\usepackage{amsmath,amssymb}
\usepackage{graphicx}
\usepackage{color}
\usepackage{url}
\usepackage[left=1in, right=1in, top=1in, bottom=1in]{geometry}
\usepackage{setspace}
\usepackage{booktabs}
\usepackage{multirow}
\usepackage{lingmacros}
\usepackage{caption}
\usepackage{subcaption}
\usepackage{rotating}
\usepackage{tablefootnote}
\usepackage{fancyvrb}
\usepackage{epigraph}
\usepackage{bbm}
\usepackage{setspace}
\usepackage{authblk}
\usepackage{fancyhdr}
\usepackage{todonotes}

\fancypagestyle{firststyle}
{
  \fancyhf{}
\fancyhead[L]{RUNNING HEAD: USEFULLY REDUNDANT REFERRING EXPRESSIONS}
\fancyhead[R]{\thepage}
\textheight = 350pt
\fancyfoot[L]{Author note: The earliest precursor of this work (the core  idea of a continuous semantics RSA model and Exp.~1) was presented as a talk at the RefNet Round Table Event in 2016 and at AMLaP 2016. Exp.~2 and the corresponding model were presented as a talk at the CUNY Conference on Sentence Processing in 2017 and as a poster at the Experimental Pragmatics (XPrag) Conference in 2017. An earlier version of  Exp.~3 and an earlier version of the corresponding model appeared in the Proceedings of the 38th Conference of the Cognitive Science Society. All experiments and models have been presented by the first author in various invited talks at workshops and colloquia in Linguistics, Psychology, Philosophy, and Cognitive Science since 2016.\\
We are grateful to many audiences and individuals for discussions that have made this paper better, in particular Sarah Brown-Schmidt, Herb Clark, Michael Franke, Gerhard J\"ager, Florian Jaeger, Emiel Krahmer, Timo Roettger, Paula Rubio-Fern\'{a}ndez, Greg Scontras, Mike Tanenhaus, and our anonymous reviewers; audiences at the RefNet Round Table Event 2015, CogSci 2016, AMLaP 2016, CUNY 2017, XPrag 2017, the UC Davis Perception, Cognition, and Cognitive Neuroscience Brown Bag, the Bay Area Philosophy of Science Group, the IMBS Quantitative Approaches to Language Workshop 2018; and audiences at colloquium talks given by the first author at Berkeley Psychology, CMU Philosophy, Dartmouth Linguistics, McGill Linguistics, OSU Linguistics, Stanford Linguistics, T\"ubingen Linguistics, and UC Davis Linguistics. All remaining errors are our own.\\
Correspondence concerning this article should be addressed to Judith Degen, Department of Linguistics, Stanford University, 450 Serra Mall, Stanford, CA 94305. E-mail: jdegen@stanford.edu.}
}

\fancypagestyle{mainstyle}
{
  \fancyhf{}
\fancyhead[L]{USEFULLY REDUNDANT REFERRING EXPRESSIONS}
\fancyhead[R]{\thepage}
\textheight = 660pt
}

\usepackage{makecell}

\definecolor{Blue}{RGB}{50,50,200}

\definecolor{Red}{RGB}{255,0,0}

\definecolor{Green}{RGB}{50,200,50}

\definecolor{Purple}{RGB}{128,0,128}

\definecolor{Orange}{RGB}{255,140,0}

\newcommand{\tableref}[1]{Table \ref{#1}}
\newcommand{\figref}[1]{Figure \ref{#1}}
\newcommand{\appref}[1]{Appendix \ref{#1}}
\newcommand{\sectionref}[1]{Section \ref{#1}}

\VerbatimFootnotes

\title{When redundancy is useful: A Bayesian approach to `overinformative' referring expressions}

\author[$\bullet$]{Judith Degen}
\author[$\bullet$]{Robert X.D.~Hawkins}
\author[$\triangleright$]{Caroline Graf}
\author[$\bullet$]{Elisa Kreiss}
\author[$\bullet$]{Noah D.~Goodman}

\affil[$\bullet$]{Stanford University}
\affil[$\triangleright$]{Freie Universit\"at Berlin}

\begin{document}

\maketitle
\thispagestyle{firststyle}

\pagebreak

\pagestyle{mainstyle}

\begin{abstract}
Referring is one of the most basic and prevalent uses of language. How do speakers choose from the wealth of referring expressions at their disposal? Rational theories of language use have come under attack for decades for not being able to account for the seemingly irrational overinformativeness ubiquitous in referring expressions. Here we present a novel production model of referring expressions within the Rational Speech Act framework that treats speakers as agents that rationally trade off cost and informativeness of utterances. Crucially, we relax the assumption that informativeness is computed with respect to a deterministic Boolean semantics, in favor of a non-deterministic continuous semantics. This innovation allows us to capture a large number of seemingly disparate phenomena within one unified framework: the basic asymmetry in speakers' propensity to overmodify with color rather than size; the increase in overmodification in complex scenes; the increase in overmodification with atypical features; and the increase in specificity in nominal reference as a function of typicality. These findings cast a new light on the production of referring expressions: rather than being wastefully overinformative, reference is usefully redundant. 
\end{abstract}


\pagebreak

When redundancy is useful: A Bayesian approach to `overinformative' referring expressions

\section[]{Overinformativeness in referring expressions}
\label{sec:intro}

Reference to objects is one of the most basic and prevalent uses of language. In order to refer, speakers must choose from a wealth of referring expressions at their disposal. How does a speaker decide whether to call an object \emph{the animal}, \emph{the dog}, \emph{the dalmatian}, or \emph{the big mostly white dalmatian}? The context within which the object occurs (other non-dogs, other dogs, other dalmatians) plays a large part in determining which features the speaker chooses to include in their utterance -- speakers aim to be sufficiently informative to establish unique reference to the intended object. However, speakers' utterances exhibit what has been claimed to be \emph{overinformativeness}: referring expressions are often more specific than necessary for establishing unique reference, and they are more specific in systematic ways.

This paper is concerned with developing a unified quantitative account for these systematic patterns, which has so far proven elusive.
We formalize our account as a computational model of referring expression production within the Rational Speech Act framework \cite{frank2012, goodman2016, FrankeJaeger2016}, which treats speakers as boundedly rational agents who optimize the tradeoff between utterance cost and informativeness. Our key innovation is to relax the assumption that informativeness of utterances is computed with respect to a deterministic Boolean semantics. Under this relaxed semantics, certain terms may apply better than others to an object without strictly being true or false. This idea has its oldest modern precursor in fuzzy logic \cite{zadeh1965fuzzy}. It is similar in spirit  to recently proposed models of meaning in both computational semantics, which assign probabilities rather than truth conditions to sentences \cite{Bernardy2018}, and in NLP, which treat word and sentence meanings  as vectors of real numbers \cite{pennington2014glove, peters2018deep, devlin2018bert}. 

As we will show,  computing utterance informativeness with respect to these more graded meanings can explain a number of seemingly disparate phenomena. 
We restrict ourselves to definite descriptions of the form \emph{the (ADJ}?\emph{)}+ \emph{NOUN}, that is, noun phrases that minimally contain the definite determiner \emph{the} followed by a head noun, with any number of restrictive adjectives occurring between the determiner and the noun.\footnote{In contrast, we will not provide a treatment of pronominal referring expressions, indefinite descriptions, names, definite descriptions with post-nominal modification, or non-restrictive modifier uses, though we offer some speculative remarks on how the approach outlined here can be applied to these cases.} 
This broad class of referring expressions subsumes two domains in language production that have been typically treated as separate. 
The choice of adjectives in (purportedly) overmodified referring expressions has been a primary focus of the language production literature
\cite{herrmann1976, Pechmann1989, nadig2002, sedivy2003a, Maes2004, Engelhardt2006, Arts2011, Koolen2011, rubiofernandez2016}, while the choice of noun in simple nominal expressions has so far mostly received attention in the concepts and categorization literature \cite{Rosch1973, Rosch1976} and in the developmental literature on generalizing basic level terms (\citeNP{Xu2007}; but see \citeNP{dale1995} for a treatment of basic level terms in natural language generation). 

In \sectionref{sec:intro} we review several key overinformativeness phenomena across these literatures that have presented a puzzle for rational accounts of language use.
In \sectionref{sec:models} we introduce the basic Rational Speech Act framework with deterministic Boolean semantics and show how it can be extended to a relaxed semantics. 
In Sections 3 - 5 we evaluate the relaxed semantics RSA model on data from interactive online reference game experiments that exhibit the phenomena introduced in \sectionref{sec:intro}: asymmetries in size and color modifier choice under varying conditions of scene complexity; typicality effects in the choice of color modifier; and choice of nominal level of reference. 
In each case, our model explains why seemingly overinformative modifiers or overly specific nouns can in fact be useful and informative; not doing so might lead the listener astray, or require them to invest too much processing effort. 
We wrap up in \sectionref{sec:gd} by summarizing our findings and discussing the far-reaching implications of and further challenges for this line of work.

\subsection{Production of referring expressions: a case against rational language use?}

How should a cooperative speaker choose between competing referring expressions? Grice, in his seminal work, provided some guidance by formulating his famous conversational maxims, intended as a guide to listeners' expectations about cooperative speaker behavior \cite{grice1975}. His maxim of Quantity, consisting of two parts, requires of speakers to:

\begin{enumerate}
	\item \emph{Quantity-1:} Make your contribution as informative as is required (for the purposes of the exchange).
	\item \emph{Quantity-2:} Do not make your contribution more informative than is required.
\end{enumerate}

That is, speakers should aim to produce neither under- nor overinformative utterances. While much support has been found for the avoidance of underinformativeness \cite{brennan1996, brown1958words, olson1970language, levinson1983pragmatics, Engelhardt2006, Davies2013}, speakers seem remarkably willing to systematically violate Quantity-2. For example, they routinely produce modifiers that are not necessary for uniquely establishing reference \cite<e.g., the \emph{small blue pin} instead of \emph{the small pin} in contexts like Figure 1a;>{gatt2011, Gatt2014, Arts2011, Koolen2011} and routinely use a basic level term even when a superordinate level term would be sufficient \cite<e.g., \emph{the dog} instead of \emph{the animal} in contexts like \figref{fig:dogexamples};>{Rosch1976, hoffmann1983objektidentifikation, TanakaTaylor91_BasicLevelAndExpertise, Johnson1997, brown1958words}.

These observations have posed a challenge for theories of language production, especially those positing rational language use (including the Gricean one): why this extra expenditure of useless effort? Why this seeming blindness to the level of informativeness requirement? Many have argued from these observations that speakers are in fact not economical \cite{Engelhardt2006, Pechmann1989}. Some have appealed to a built-in preference for referring at the basic level from considerations of conceptual representation or perceptual factors such as shape \cite{Rosch1976, Rosch1973, murphy1982basic}. Others have argued for salience-driven effects on willingness to overmodify \cite{Gatt2014, Westerbeek2015}. In all cases, it is argued that informativeness itself cannot be the key factor in determining the content of speakers' referring expressions. Here we revisit this claim and show that systematically relaxing the requirement of a deterministic Boolean semantics for referring expressions also systematically changes the informativeness of utterances. This results in a reconceptualization of what have been termed \emph{overinformative referring expressions} as \emph{usefully redundant referring expressions}. We begin by reviewing the phenomena of interest that a revised theory of definite referring expressions should be able to account for. 

\subsection{Overinformativeness in modified referring expressions}
\label{sec:modified}

Most of the literature on overinformative referring expressions has been devoted to the use of overinformative modifiers in modified referring expressions. The prevalent observation is that speakers frequently do not include only the minimal modifiers required for establishing reference, but often also include redundant modifiers \cite{Pechmann1989, nadig2002, Maes2004, Engelhardt2006, Arts2011, Koolen2011}. However, not all modifiers are created equal: there are systematic differences in the overmodification patterns observed for size adjectives (e.g., \emph{big, small}), color adjectives (e.g., \emph{blue, red}), material adjectives (e.g., \emph{plastic, wooden}), and others \cite{sedivy2003a}. Furthermore, these asymmetries interact with features of the context and world knowledge about the typicality of different properties.   

\paragraph{Asymmetry in redundant use of color and size adjectives.}
\label{sec:asymmetry}

\begin{figure}
\begin{subfigure}{.5\textwidth}
\includegraphics[width=\textwidth]{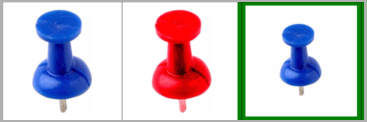}
\caption{Size sufficient.}
\label{fig:sizesufficient}
\end{subfigure}
\begin{subfigure}{.5\textwidth}
\includegraphics[width=\textwidth]{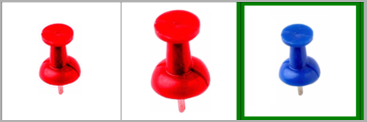}
\caption{Color sufficient.}
\label{fig:colorsufficient}
\end{subfigure}
\caption{Example contexts where (a) size only (e.g., \emph{the small pin}) or (b) color only (e.g., \emph{the blue pin}) is sufficient for unique reference. Thick border marks the intended referent.}
\label{fig:pin}
\end{figure}

In \figref{fig:sizesufficient}, distinguishing the object highlighted by the thick border requires only mentioning its size (\emph{the small pin}). 
It is now well-documented that speakers routinely include redundant color adjectives (\emph{the small blue pin}) which are not necessary for uniquely singling out the intended referent in these kinds of contexts \cite{Pechmann1989, Belke2002, gatt2011}. However, the same is not true for size: in contexts like \figref{fig:colorsufficient}, where color is sufficient for unique reference (\emph{the blue pin}), speakers overmodify much more rarely. Though there is quite a bit of variation in proportions of overmodification, this asymmetry in the propensity for overmodifying with color but not size has been documented repeatedly \cite{Pechmann1989, sedivy2003a,gatt2011, rubiofernandez2016,Westerbeek2015,Koolen2013}. 

\paragraph{Scene variation.}
\label{sec:scenevariation}

Speakers' propensity to overmodify with color is highly dependent on features of the distractor objects in the context. 
In particular, as the variation present in the scene increases, so does the probability of overmodifying.
For example \citeA{Koolen2013} consistently found higher rates of overmodification with color adjectives in high-variation scenes (28-27\%) compared to the low-variation ones (4-10\%).
Scene variation has been quantified in several different ways: the number of dimensions along which objects differ \citeA{Koolen2013}, the number of distractors present in a scene \citeA{gatt2017}, and whether objects are `simple' or `compositional' \citeA{Davies2013}. 
A model of referring expression generation should ideally capture all of these types of variation in a unified way. 

\paragraph{Feature typicality.}
\label{sec:colortypicalityintro}

\begin{figure}
\begin{subfigure}{.5\textwidth}
\includegraphics[width=\textwidth]{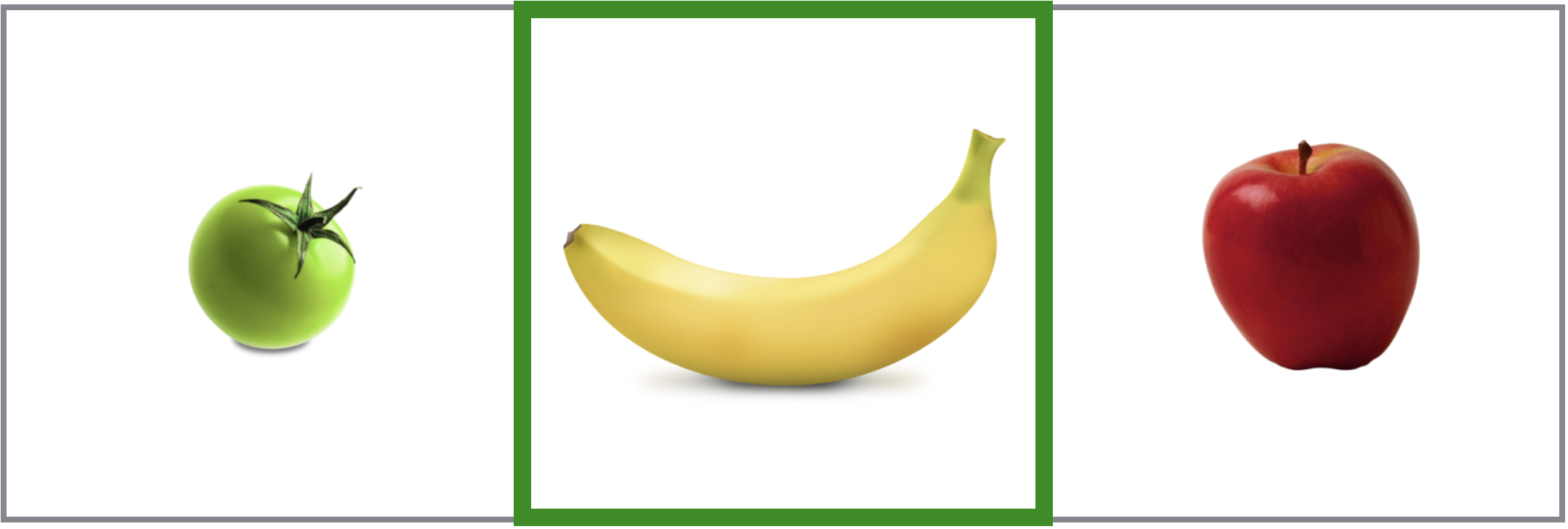}
\caption{Typical color, type sufficient.}
\label{fig:typical}
\end{subfigure}
\begin{subfigure}{.5\textwidth}
\includegraphics[width=\textwidth]{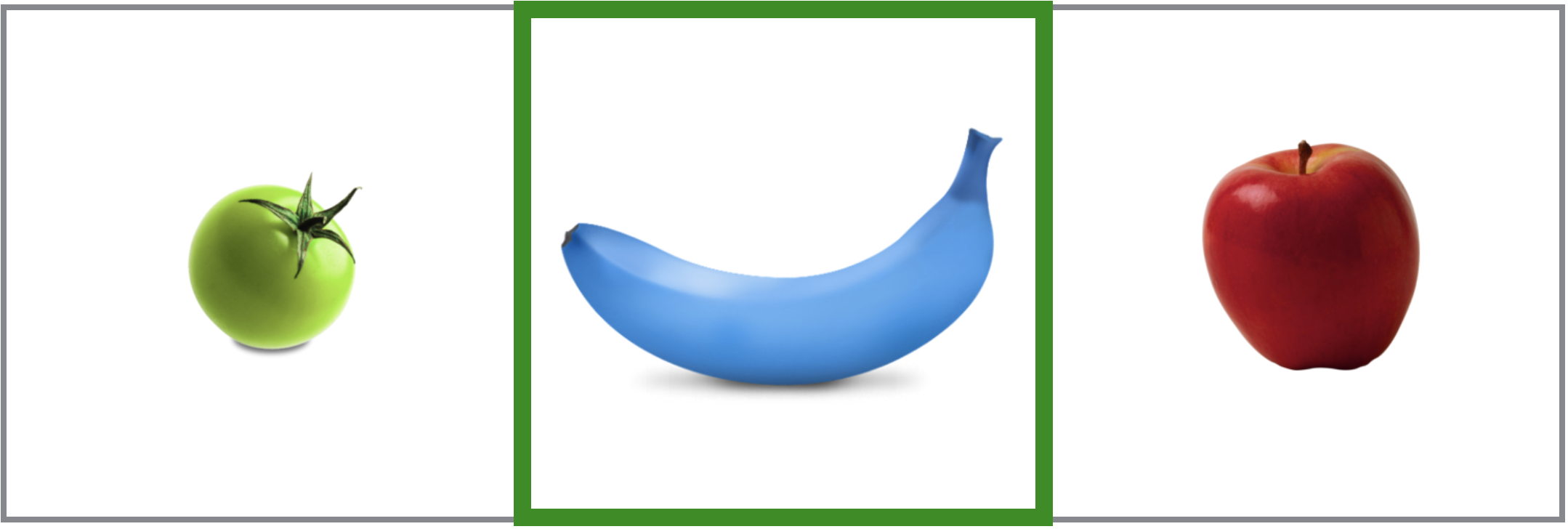}
\caption{Atypical color, type sufficient.}
\label{fig:atypical}
\end{subfigure}
\caption{Example contexts where type (\emph{banana}) is sufficient for unique reference and color is (a) typical or (b) atypical. A thick border marks the intended referent.}
\label{fig:banana}
\end{figure}

Overmodification with color has also been shown to be systematically related to the typicality of the color for the object. 
\citeA{Westerbeek2015} have shown that the more typical a color is for an object, the less likely it is to be mentioned when not necessary for unique reference \cite<see also>{sedivy2003a,rubiofernandez2016}. For example, speakers never refer to a yellow banana in the absence of other bananas as \emph{the yellow banana} (see \figref{fig:typical}), but they sometimes refer to a brown banana as \emph{the brown banana}, and they almost always refer to a blue banana as \emph{the blue banana} (see \figref{fig:atypical}). Similar typicality effects have been shown for other (non-color) properties. For example, \citeA{Mitchell2013} showed that speakers are more likely to include an atypical than a typical property (either shape or material) when referring to everyday objects like boxes when mentioning at least one property was necessary for unique reference. 

\subsection{Overinformativeness in nominal referring expressions}
\label{sec:nominalintro}

\begin{figure}
\begin{subfigure}{.5\textwidth}
\includegraphics[width=\textwidth]{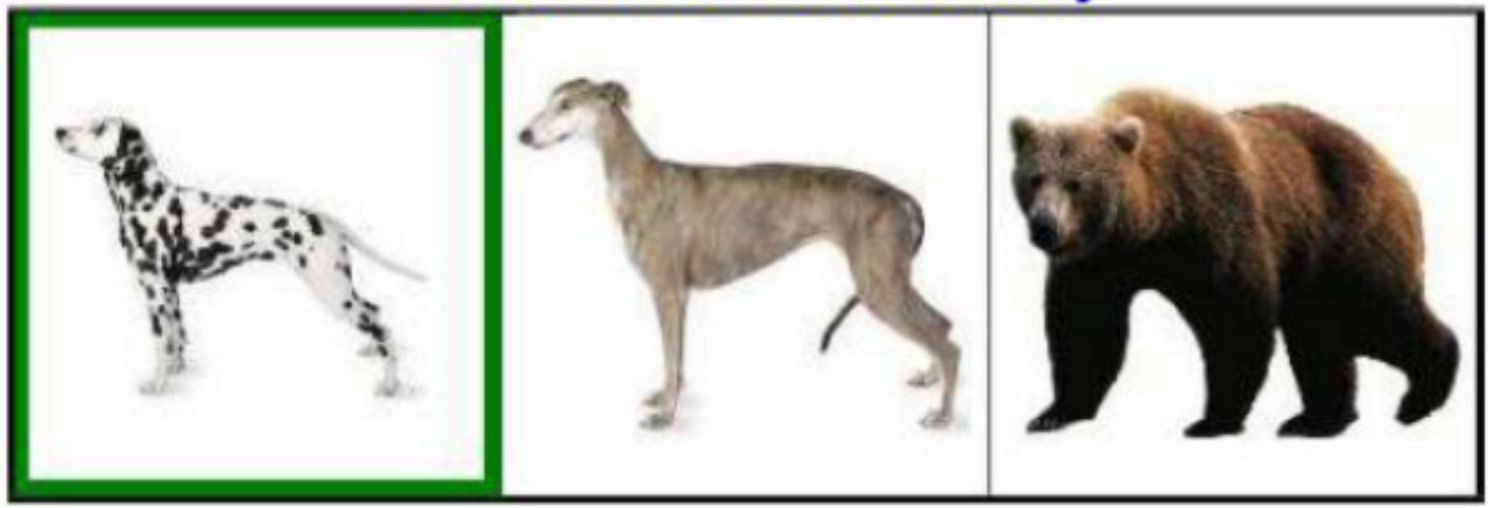}
\caption{Subordinate level term necessary.}
\label{fig:subnecessary}
\end{subfigure}
\begin{subfigure}{.5\textwidth}
\includegraphics[width=\textwidth]{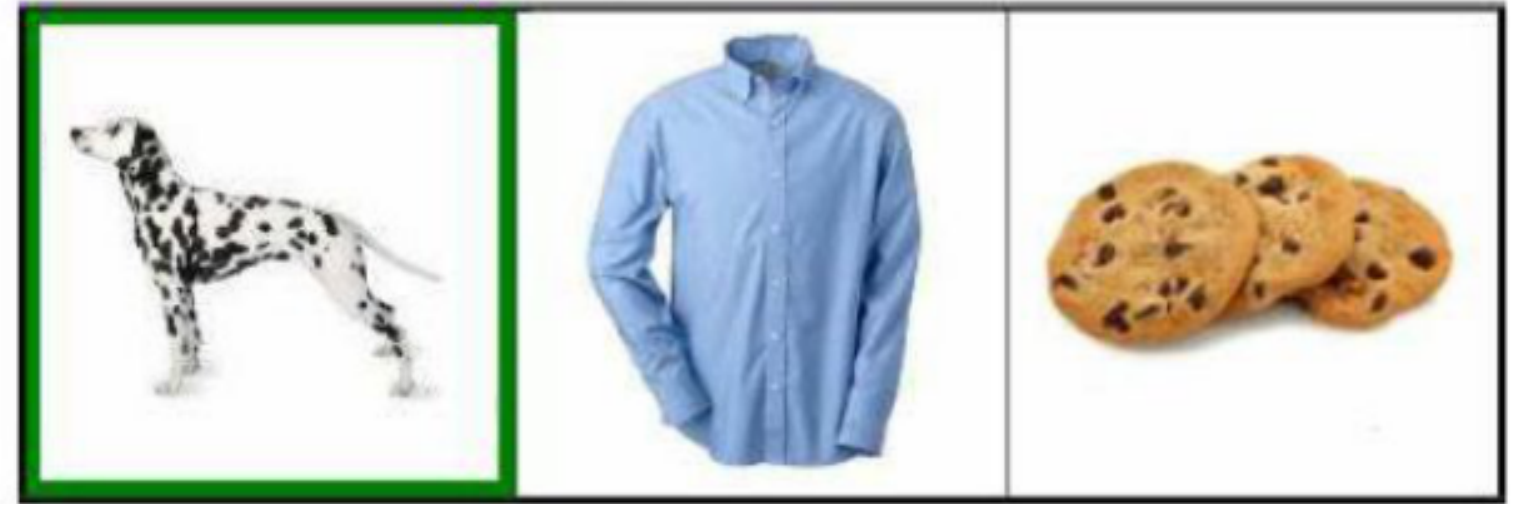}
\caption{Superordinate level term sufficient.}
\label{fig:supersufficient}
\end{subfigure}
\caption{Example contexts in which different levels of reference are necessary for establishing unique reference to the target marked with a thick border. (a) subordinate (\emph{dalmatian}) necessary; (b) superordinate (\emph{animal}) sufficient, but basic  (\emph{dog}) or subordinate (\emph{dalmatian}) possible.}
\label{fig:dogexamples}
\end{figure}

Even in the absence of modifying adjectives, a referring expression can be more or less informative: \emph{the dalmatian} communicates more information about the object in question than \emph{the dog} (being a dalmatian entails being a dog), which in turn is globally more informative than \emph{the animal}. 
Thus, this choice can be considered analogous to the choice of adding more modifiers -- in both cases, the speaker has a choice of being more or less specific about the intended referent. 
A well-documented effect from the concepts and categorization literature is that speakers prefer to refer at the \emph{basic level} \cite{Rosch1976, Tanaka1991}. That is, in the absence of other constraints, even when a superordinate level term would be sufficient for establishing reference (as in \figref{fig:supersufficient}), speakers prefer to say \emph{the dog} rather than \emph{the animal}. 
However, there are systematic exceptions: in some cases when the basic level would be sufficient, speakers prefer the \emph{subordinate} term.
For example, atypical birds like penguins are often referred to at the subordinate level rather than at the basic level \emph{bird} \cite{Jolicoeur1984}.
Indeed, children may even use expectations about such referential preferences to infer narrower categories from atypical exemplars during word learning \cite{emberson2019blowfish}.

\begin{table}
\caption{List of effects a theory of referring expression production should account for and paper section(s) in which they are treated.}
\begin{tabular}{l l l } 
\toprule
Section & Effect & Description \\ 
\midrule
2 \& 3 & Color/size asymmetry & More redundant use of color than size \tablefootnote{Reported by many \cite<e.g.,>{Pechmann1989, Engelhardt2006, gatt2011, rubiofernandez2016}}\\ 
2 \& 3 & Scene variation & More redundant use of color with increasing scene variation \tablefootnote{Multiple replications reported \cite<e.g.,>{Davies2013, Koolen2013}}\\
\midrule
4 & Color typicality & More redundant use of color with decreasing color typicality \tablefootnote{Multiple replications reported \cite<e.g.>{sedivy2003a, Westerbeek2015, rubiofernandez2016}}\\ 
\midrule
5 & Basic level preference & Preference for basic level term when superordinate  sufficient \tablefootnote{Originally reported by \citeA{Rosch1976}, dozens of replications.}\\ 
5 & Subordinate level use & Unnecessary use of subordinate level term  \tablefootnote{Reported by \citeA{Jolicoeur1984}}\\ 
\bottomrule
\end{tabular}
\label{tab:effects}
\end{table}

\section[]{Modeling speakers' choice of referring expression}
\label{sec:models}

To date, there is no theory to account for all of these different phenomena (see \tableref{tab:effects}), and no model has attempted to unify the domains of modified and nominal referring expressions. 
Here we propose an explicit computational account of how multiple factors --- including an utterance's literal semantics, its contextual informativity, its cost relative to alternative utterances, and the typicality of an object or its features --- interact in referring expression production.
We argue that this model provides a principled explanation for the phenomena reviewed in the previous section and holds promise for being generalizable to many further production phenomena related to overinformativeness, which we discuss in relation to previous accounts in \sectionref{sec:gd}.

Our model is formulated within the Rational Speech Act (RSA) framework \cite{frank2012, goodman2016}.\footnote{All RSA models and Bayesian data analyses reported in this paper were implemented in the probabilistic programming language WebPPL \cite{GoodmanStuhlmuller14_DIPPL} and can be viewed at \url{https://github.com/thegricean/RE_production}. All experimental materials and analysis scripts are available in the same repository. An interactive browser-based toy model is provided at \url{http://forestdb.org/models/overinf.html}.} 
We proceed by first presenting the general production framework in \sectionref{sec:basicrsa}, and show why the most basic model, as formulated by \citeNP{frank2012}, does not produce the phenomena outlined above due to its strong focus on speakers maximizing the informativeness of expressions under a deterministic Boolean semantics. In \sectionref{sec:modifiedmodel} we introduce our crucial innovation: relaxing the semantics. 

\subsection{Basic RSA}
\label{sec:basicrsa}

The production component of RSA aims to soft-maximize the utility of utterances, where utility is defined in terms of the contextual informativeness of an utterance, given each utterance's literal semantics. Formally, this is treated as a pragmatic speaker $S_1$ reasoning about a literal listener $L_0$, who can be described by the following formula:

\begin{equation}
\label{eq:listener}
P_{L_0}(o | u) \propto \mathcal{L}(u,o).
\end{equation}

The literal listener $L_0$ observes an utterance $u$ from the set of  utterances $U$, consisting of single adjectives denoting features available in the context of a set of objects  $O$, and returns a distribution over objects $o \in O$. Here, $\mathcal{L}(u,o)$ is the lexicon that encodes deterministic lexical meanings such that: 

\begin{equation}
\mathcal{L}(u,o) = \left\{
 \begin{array}{rl}
  1 & \text{if } u \text{ is true of } o\\
   0 & \text{otherwise}.
 \end{array} \right.
\end{equation}

 Thus, $P_{L_0}(o | u)$ returns a uniform distribution over all contextually available $o$ in the extension of $u$. For example, in the size-sufficient context shown in \figref{fig:sizesufficient}, $U = \{\textrm{\emph{big}}, \textrm{\emph{small}}, \textrm{\emph{blue}}, \textrm{\emph{red}}\}$ and $O = \{o_{\textrm{big\_blue}}, o_{\textrm{big\_red}}, o_{\textrm{small\_blue}}\}$. Upon observing \emph{blue}, the literal listener therefore assigns equal probability to  $o_{\textrm{big\_blue}}$ and $o_{\textrm{small\_blue}}$. Values of $P_{L_0}(o | u)$ for each $u$ are shown on the left in \tableref{tab:detliteral}.

The pragmatic speaker in turn produces an utterance with probability proportional to the utility of that utterance:
\begin{equation}
P_{S_1}(u | o) \propto e^{U(u,o)}
\end{equation}

The speaker's utility $U(u,o)$ is a function of both the utterance's \emph{informativeness}  with respect to the literal listener $P_{L_0}(o | u)$ and the utterance's \emph{cost} $c(u)$:

\begin{equation}
U(u,o) = \beta_{i} \ln P_{L_0}(o | u) - \beta_c c(u)
\end{equation}

Two free parameters, $\beta_i$ and $\beta_c$ enter the computation, weighting the respective contributions of informativeness and utterance cost, respectively.\footnote{\citeA{frank2012} fixed $\beta_i = 1$ and did not include cost in their formulation, because they assumed equal costs for all utterances. Subsequent work has demonstrated the importance of taking into account utterance cost in modeling  interpretation phenomena like cost-based quantity implicatures \cite{rohde2012,degenfrankejaeger2013} and M-implicature \cite{bergen2016}. We include it here because of the importance that cost has played in explanations of overinformative referring expressions, where it typically surfaces as the idea that speakers have different overall preferences for mentioning color vs.~size modifiers \cite{dale1995, Koolen2011, VanGompel2019}. At this point we remain agnostic about the factors that contribute to an utterance's cost $c(u)$. In later sections we allow cost to be a function of properties (e.g. color \& size) mentioned in the utterance, or of an utterance's empirical length and corpus frequency; our policy for these cases is to introduce free cost parameters for each linear component of the cost function.}  In order to understand the effect of $\beta_i$, it is useful to explore its effect when utterances are cost-free. In this case, as $\beta_i$ approaches infinity, the speaker increasingly only chooses utterances that maximize informativeness; if $\beta_i$ is 0, informativeness is disregarded and the speaker chooses randomly from the set of all available utterances; if $\beta_i$ is 1, the speaker probability-matches, i.e., chooses utterances proportional to their informativeness \cite<equivalent to Luce's choice rule,>{luce1959}. Applied to the example in \tableref{tab:detliteral}, if the speaker wants to refer to $o_{\textrm{small\_blue}}$ they have two semantically possible utterances, \emph{small} and \emph{blue}, where \emph{small} is twice as informative as \emph{blue}. They produce \emph{small} with probability 1 when $\beta_i \rightarrow \infty$, probability 2/3 when $\beta_i = 1$ and probability 1/4 when $\beta_i = 0$.\footnote{Note that instead of a $\beta_i$ parameter weighting informativeness \emph{inside} the utility function, other recent formulations have used an $\alpha$ parameter modulating the entire utility function, i.e. $P_{S_1}(u |o)\propto \exp{\alpha U(u,o)}$. These parameterizations are equivalent. In the present work, where informativeness and cost both play important roles, we chose the `flattened' linear combination with independent weights for simplicity.}
Conversely, disregarding informativeness and focusing only on cost, any asymmetry in costs will be exaggerated with increasing $\beta_c$, such that the speaker will choose the least costly utterance with higher and higher probability as $\beta_c$ increases.

As has been pointed out by \citeA{VanGompel2019}, the basic Rational Speech Act model described so far \cite{frank2012} does not generate overinformative referring expressions for two reasons. 
One of these is trivial: $U$ only contains one-word utterances. We can ameliorate this easily by allowing complex two-word utterances. We assume an intersective semantics for complex utterances $u_{\textrm{complex}}$ that consist of a two adjective sequence $u_{\textrm{size}} \in \{\textrm{\emph{big}}, \textrm{\emph{small}}\}$ and $u_{\textrm{color}} \in \{\textrm{\emph{blue}}, \textrm{\emph{red}}\}$, such that the meaning of a complex two-word utterance is defined as
\begin{equation} 
\label{eq:prodcomp}
\mathcal{L}(u_{\text{complex}},o) = \mathcal{L}(u_{\text{size}},o) \times \mathcal{L}(u_{\text{color}},o).
\end{equation} 
The resulting renormalized literal listener distributions for our example size-sufficient context in \figref{fig:sizesufficient} are shown in the left columns in \tableref{tab:detliteral},\footnote{`Normalization' refers to the process of turning a set of numbers into a probability distribution by dividing each number by the sum of all the numbers in the set, such that they add up to 1.} and the concomitant pragmatic speaker distributions are shown in the left columns in \tableref{tab:speaker}.\footnote{An interactive toy version of this model is provided at \url{http://forestdb.org/models/overinf.html}.}

\begin{table}
\centering
\caption{Row-wise literal listener distributions $P_{L_0}(o | u)$ for each utterance $u$ in the size-sufficient context depicted in \figref{fig:sizesufficient}, under a deterministic Boolean semantics (left) or under a continuous semantics (right) with  $x_{\text{size}} = .8$, $x_{\text{color}} = .99$. Bolded numbers indicate crucial comparisons between literal listener probabilities in correctly selecting the intended referent $o_{\text{small\_blue}}$ in response to observing the sufficient \emph{small} and the redundant \emph{small blue} utterances.}
\begin{tabular}{l r r r r r r}
\toprule
& \multicolumn{3}{c}{Boolean} & \multicolumn{3}{c}{continuous}\\ 
& $o_{\textrm{big\_blue}}$ & $o_{\textrm{big\_red}}$ & $o_{\textrm{small\_blue}}$
& $o_{\textrm{big\_blue}}$ & $o_{\textrm{big\_red}}$ & $o_{\textrm{small\_blue}}$ \\
\midrule
\emph{big} & .5 & .5 & 0 & .39 & .39 & .22 \\
\emph{small} & 0 & 0 & \textbf{1} & .26 & .26 & \textbf{.48} \\
\emph{blue} & .5 & 0 & .5 & .42 & .16 & .42 \\
\emph{red} & 0 & 1 & 0 & .21 & .57 & .21 \\
\emph{big blue}  & 1 & 0 & 0 & .50 & .23 & .27 \\
\emph{big red}  & 0 & 1 & 0 & .24 & .52 & .24 \\
\emph{small blue} & 0 & 0 & \textbf{1} & .27 & .23 & \textbf{.50} \\
\bottomrule
\end{tabular}
\label{tab:detliteral}
\end{table}

\begin{table}
\centering
\caption{Column-wise pragmatic speaker distributions $P_{S_1}(u | o)$ for each object $o$ in the size-sufficient context depicted in \figref{fig:sizesufficient}, under a deterministic Boolean semantics (left) or under a continuous semantics (middle, right) with  $x_{\text{size}} = .8$, $x_{\text{color}} = .99$, with $\beta_i$ set to 1 (middle) or 30 (right). Bolded numbers indicate the relevant speaker probabilities for the minimal (\emph{small}) and redundant (\emph{small blue}) utterances when intending to communicate referent $o_{\text{small\_blue}}$.}
\small
\begin{tabular}{l r r r r r r r r r}
\toprule
& \multicolumn{3}{c}{Boolean} & \multicolumn{3}{c}{continuous ($beta_i = 1$)} & \multicolumn{3}{c}{continuous ($\beta_i = 30$)}\\ 
& $o_{\textrm{big\_blue}}$ & $o_{\textrm{big\_red}}$ & $o_{\textrm{small\_blue}}$
& $o_{\textrm{big\_blue}}$ & $o_{\textrm{big\_red}}$ & $o_{\textrm{small\_blue}}$
& $o_{\textrm{big\_blue}}$ & $o_{\textrm{big\_red}}$ & $o_{\textrm{small\_blue}}$ \\
\midrule
\emph{big} & .25 & .2 & 0 & .17 & .17 & .09 & 0 & 0 & 0 \\
\emph{small} & 0 & 0 & \textbf{.4} & .11  & .11 & \textbf{.20} & 0 & 0 & \textbf{.21} \\
\emph{blue} & .25 & 0 & .2 & .18 & .07 & .18 & .01 & 0 & 0 \\
\emph{red} & 0 & .4 & 0 & .09 & .24 & .09 & 0 & .93 & 0 \\
\emph{big blue}  & .5 & 0 & 0 & .22 & .10 & .12 & .99 & 0 & 0 \\
\emph{big red}  & 0 & .4 & 0 & .10 & .22 & .10 & 0 & .07 & 0 \\
\emph{small blue} & 0 & 0 & \textbf{.4} & .12 & .10 & \textbf{.21} & 0 & 0 & \textbf{.79} \\
\bottomrule
\end{tabular}
\label{tab:speaker}
\end{table}

Unfortunately, simply including complex utterances in the set of alternatives does not solve the problem. We turn again to the case where the speaker wants to communicate the small blue object. There are now two useful utterances, \emph{small} and \emph{small blue}, for referring to this object. 
Because they are equally informative (see bolded numbers in  \tableref{tab:detliteral}, column 3), the pragmatic speaker is equally likely to produce them (see bolded numbers in \tableref{tab:speaker}, column 3). The only way for the more complex utterance to be chosen with greater probability than the simple utterance is if it was the \emph{cheaper} one. While this would achieve the desired mathematical effect, the cognitive plausibility of complex utterances being cheaper than simple utterances is highly dubious\footnote{See also the discussion of cost functions in \citeA{Krahmer2003}, who explicitly introduce this monotonicity constraint as a constraint on the search space of possible referring expressions within a graph-based framework.}. 
Thus we must look elsewhere to break the symmetry and account for overinformativeness. We propose that the place to look is the computation of informativeness itself.

\subsection{RSA with continuous semantics}
\label{sec:modifiedmodel}

Here we introduce the crucial innovation: rather than assuming a deterministic Boolean semantics that returns true (1) or false (0) for any combination of expression and object, we relax to a continuous semantics that returns real values in the interval $[0,1]$. 
Formally, the only change is in the values that the lexicon can return:
\begin{equation}
\mathcal{L}(u,o) \in [0, 1] \subset \mathbb{R}
\end{equation}
That is, rather than assuming that an object is unambiguously big (or not) or unambiguously blue (or not), this continuous semantics captures that objects count as big or blue to  varying degrees \cite<similar to approaches in fuzzy logic, prototype theory, and recent developments in NLP;>{zadeh1965fuzzy, Rosch1973, Bernardy2018}.

Another approach to relaxing the deterministic Boolean semantics would be to relax the determinism. This can be done either by assuming a semantics which is fundamentally Boolean, but whose truth-values contain an element of randomness; or by assuming a fully deterministic Boolean semantics with intensional parameters that are themselves random variables. This is appealing because if would preserve the existing machinery of standard truth-functional compositional semantics. It can be shown that using continuous semantic values in the RSA model is equivalent to using Boolean values that are chosen non-deterministically. Conversely, marginalizing over the randomness in a Boolean semantics yields a probability of truth, which is a value between 0 and 1. For this reason we will sometimes refer to the relaxed semantics as a ``noisy'' semantics, and the deviation of the semantic value from 0 or 1 as the degree of noise. We will generally treat the relaxed semantics in its continuous value guise, as it simplifies exposition and development.

We now show via simulations that this model can qualitatively account both for speakers' asymmetric propensity to overmodify with color rather than with size (in \sectionref{sec:colorsizesimulation}) and  for speakers' propensity to overmodify more with increasing scene variation (in \sectionref{sec:modelkoolen}). 
The intuition, using the example from \figref{fig:sizesufficient}, is that \emph{blue} and \emph{small} do not apply equally well to all roughly blue, roughly small objects, and that a speaker might opt to include more modifiers when any one alone might not be a perfectly apt descriptor. Assuming that \emph{blue} is more precise than \emph{small} leads the speaker to overmodify more with color than with size -- and further, the more variability is present in the scene, the more the precision of color helps weed out non-intended referents, i.e., the more color overmodification  occurs. 

\subsubsection{Simulation 1: color-size asymmetry}
\label{sec:colorsizesimulation}

To see the basic effect of switching to a continuous semantics, and to see how far we can get in capturing overinformativeness patterns with this change, let us explore a simple semantics in which all colors are treated the same, all sizes are as well, and the two compose via a product rule.
That is, when an object $o$ is in the extension of a size adjective under a Boolean semantics -- i.e., when the size can be truthfully predicated of $o$ -- we take $\mathcal{L}(u,o) = x_{\text{size}}$, a constant; when it is not in the extension of the adjective  -- i.e., when the size cannot be truthfully predicated of $o$ --  $\mathcal{L}(u,o) = 1 - x_{\text{size}}$. 
Similarly for color adjectives. 
This results in two free model parameters, $x_{\text{size}}$ and $x_{\text{color}}$, that can take on different values, capturing that size and color adjectives may apply more or less well/reliably to objects.
Together with the product composition rule, Eq.~\ref{eq:prodcomp}, this fully specifies a relaxed semantic function for our reference domain.\footnote{An interactive toy version of this model is provided at \url{http://forestdb.org/models/overinf.html}.}

Now consider the RSA literal listener, Eq.~\ref{eq:listener}, who uses these relaxed semantic values.
Given an utterance, the listener simply normalizes over potential referents. 
As an example, the resulting renormalized literal listener distributions for the size-sufficient example context in \figref{fig:sizesufficient} are shown for values  $x_{\text{size}} = .8$ and $x_{\text{color}} = .99$ on the right in \tableref{tab:detliteral}.\footnote{These values were chosen for the demonstration because they are the ones that result in the best approximation of the proportion of redundant referring expressions reported in \citeA{VanGompel2019}: 79\% in size-sufficient contexts; 7\% in color-sufficient contexts.} Recall that in this context, the speaker intends for the listener to select the small blue pin. To see which would be the best utterance to produce for this purpose, we compare the literal listener probabilities in the $o_{\text{small\_blue}}$ column. The two best utterances under both the Boolean and the continuous semantics are bolded in the table: under the Boolean semantics, the two best utterances are \emph{small} and \emph{small blue}, with no difference in listener probability. In contrast, under the continuous semantics \emph{small} has a smaller literal listener probability (.48) of retrieving the intended referent than the redundant \emph{small blue} (.50). While this difference may appear small, it is enough to break the symmetry in utterance informativeness. Consequently, the pragmatic speaker will be more likely to produce \emph{small blue} than \emph{small}, though the precise probabilities depend on the cost and informativeness parameters $\beta_c$ and $\beta_i$. \tableref{tab:speaker} shows the resulting pragmatic speaker probabilities under a low and a high $\beta_i$ with no utterance cost. 

Crucially, the reverse is not the case when color is the distinguishing dimension. Consider the speaker in the same context wanting to communicate the big red pin. The two best utterances for this purpose are \emph{red} (.57) and \emph{big red} (.52). In contrast to the results for the small blue pin, the redundant  utterance does not increase the literal listener probability of inferring the intended referent. The reason for this is that we defined color to be almost noiseless, with the result that the literal listener distributions in response to utterances containing color terms are more similar to those obtained via a Boolean semantics than the distributions obtained in response to utterances containing size terms. The reader is encouraged to verify this by comparing the row-wise distributions under the  Boolean and continuous semantics in \tableref{tab:detliteral}.

\begin{figure}
\centering
\includegraphics[width=.8\textwidth]{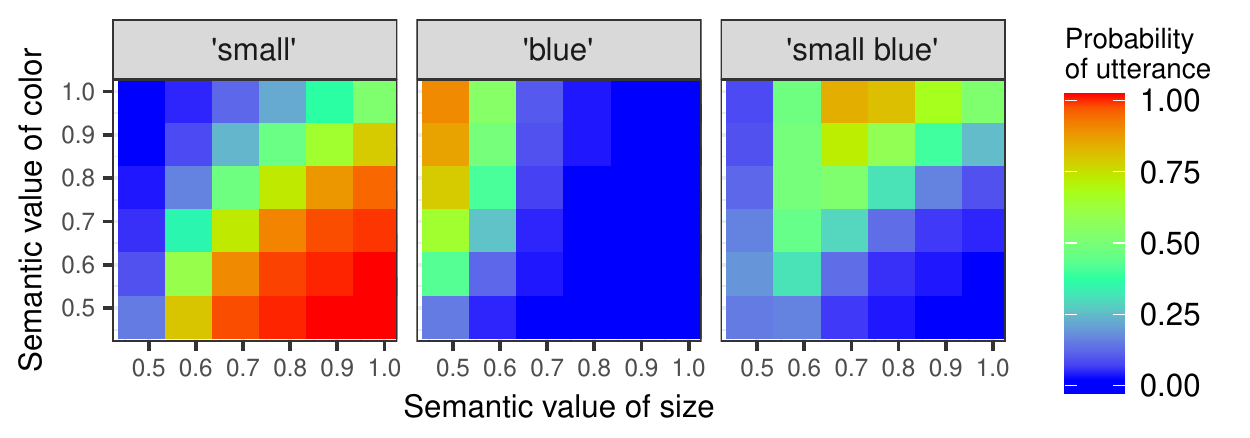}

\caption{Probability of producing sufficient \emph{small pin}, insufficient \emph{blue pin}, and redundant \emph{small blue pin} in contexts as depicted in \figref{fig:sizesufficient}, as a function of semantic value of color and size utterances (for $\beta_i = 30$ and $ \beta_c = 0$). For a visualization of model behavior under varying $\beta_i$, see \appref{app:modelexploration}.}
\label{fig:basicasymmetry}
\end{figure}

To better understand the consequences of continuous meanings in contexts like that depicted in \figref{fig:sizesufficient}, we visualize the results of varying  $x_{\text{size}}$ and $x_{\text{color}}$ in \figref{fig:basicasymmetry}. 
The deterministic Boolean semantics of utterances is approximated where the  semantic values of both size and color utterances are close to 1 (.999, top right-most point in graph).  In this case, the simple sufficient (\emph{small pin}) and complex redundant utterance (\emph{small blue pin}) are equally likely because they are both equally informative and utterances are assumed to have 0 cost. All other utterances are highly unlikely. The interesting question is under which circumstances, if any,  the standard color-size asymmetry emerges. This asymmetry is found in the warmer region of the `small blue' facet, characterized by values of $x_{\text{size}}$ that are lower than $x_{\text{color}}$, with high values for $x_{\text{color}}$. That is, redundant utterances are more likely than sufficient utterances when the redundant dimension (in this case color) is less noisy than the sufficient dimension (in this case size) and overall is close to noiseless. 
Thus, when size adjectives are noisier than color adjectives, the model produces overinformative referring expressions with color, but not with size -- precisely the pattern observed in the literature \cite{Pechmann1989, gatt2011}. Note also that no difference in adjective \emph{cost} is necessary for obtaining the overinformativeness asymmetry, though assuming a greater cost for size than for color does further increase the observed asymmetry (see \sectionref{sec:modifiermodeleval} for further discussion).

\subsubsection{Simulation 2: scene variation}
\label{sec:modelkoolen}

In the previous section, we showed that extending RSA with continuous adjective semantics gives rise to color-size asymmetries when the semantics of color adjectives is closer to deterministic Boolean truth-functions than size adjectives. When modifiers are noisy, adding `stricter' modifiers adds information. From this perspective, these additional modifiers are not \emph{over}informative; they are usefully redundant given the needs of the listener. 
Next, we show how the same mechanism accounts for why increased scene variation increases the probability that referring expressions are overmodified with color. 

\begin{figure}
\begin{subfigure}{.5\textwidth}
\includegraphics[width=\textwidth]{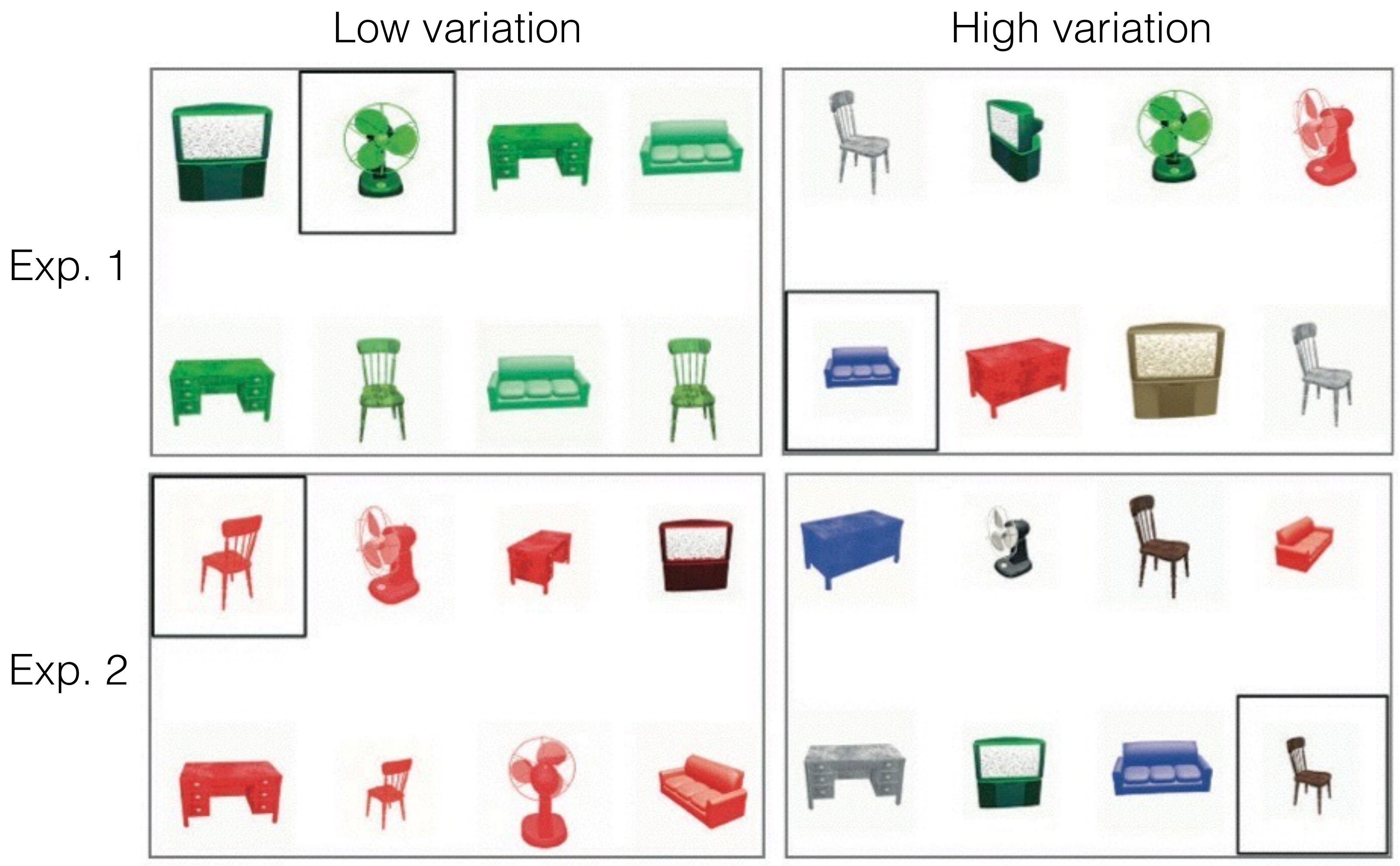}
\caption{Contexts from Koolen et al.~(2013)'s low variation (left column) and high variation (right column) conditions in Exp.~1 (top row) and Exp.~2 (bottom row).}
\label{fig:koolencontexts}
\end{subfigure}
\begin{subfigure}{.5\textwidth}
\centering
\includegraphics[width=.75\textwidth]{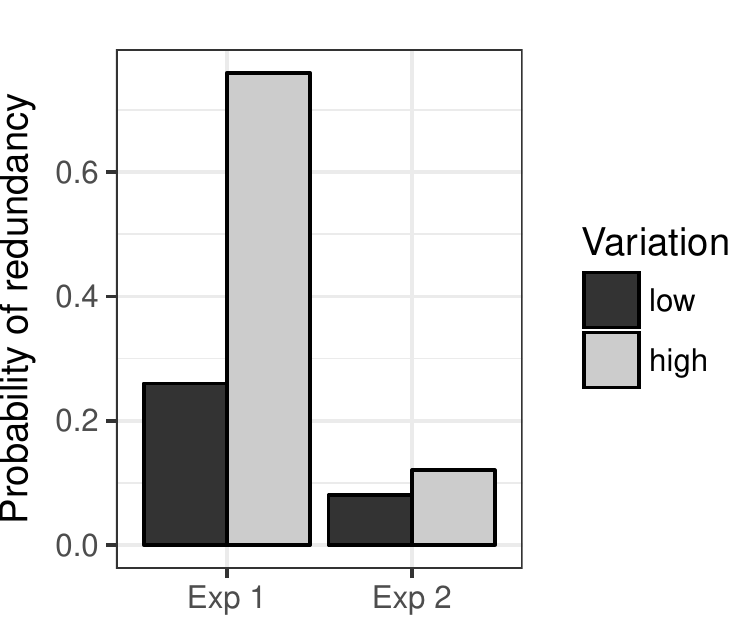}
\caption{Predicted probability of redundant color utterance in Koolen et al.~(2013) conditions for $\beta_i = 30$, $ \beta_c = c(u_{\textrm{size}}) = c(u_{\textrm{color}}) = 1$, $x_{\text{size}} = .8$, $x_{\text{color}} = .999$, $x_{\text{type}} = .9$.}
\label{fig:koolensimulationresults}
\end{subfigure}
\caption{Visual contexts employed in experiments by Koolen et al.~(2013) alongside RSA model predictions for the use of redundant modifiers in those contexts.}
\end{figure}

\citeA{Koolen2013}  quantified scene variation as the number of feature dimensions along which pieces of furniture in a scene varied: type (e.g., chair, fan), size (big, small), and color (e.g., red, blue).\footnote{They also included orientation (left-facing, right-facing) as a dimension along which objects could vary in certain cases. We ignore this dimension here for the sake of simplicity.} 
Scene variation was manipulated across two experiments, which differed in the dimension necessary for unique reference (color was always redundant). 
In Exp.~1, only type was necessary (\emph{fan} and \emph{couch} in the low and high variation conditions in \figref{fig:koolencontexts}, respectively). 
In Exp.~2, size and type were necessary (\emph{big chair} and \emph{small chair} in \figref{fig:koolencontexts}, respectively). 
Across both experiments, lower rates of redundant color use were found in the low variation conditions (4\% and 9\%) than in the high variation conditions (24\% and 18\%).
Here, we use simulations to explore the predictions that continuous semantics RSA  -- henceforth \emph{cs-RSA} -- makes for these situations.

Following \citeA{Koolen2013}, we considered any mention of color as a redundant mention. In Exp.~1, this includes the simple redundant utterances like \emph{blue couch} as well as complex redundant utterances like \emph{small blue couch}. In Exp.~2, where size was necessary for unique reference, only the complex redundant utterance \emph{small brown chair} was truly redundant (\emph{brown chair} was insufficient, but still included in counts of color mention). 
Because object type was a distinguishing dimension, we introduce an additional semantic value $x_{\text{type}}$, which encodes how noisy nouns are.
The results of simulating these conditions with parameters $\beta_i = 30$, $ \beta_c = c(u_{\textrm{size}}) = c(u_{\textrm{color}}) = 1$, $x_{\text{size}} = .8$, $x_{\text{color}} = .999$, and $x_{\text{type}} = .9$ are shown in \figref{fig:koolensimulationresults}, under the assumption that the cost of a two-word utterance $c(u)$ is the sum of the costs of the one-word sub-utterances.\footnote{These parameter values were chosen merely for convenience in illustrating the qualitative model predictions. We reused values from the previous example, where possible, but also included a cost per word.}
For both experiments, the model exhibits the empirically-observed qualitative effect of variation on the probability of redundant color mention: when variation is greater, redundant color mention is more likely. 
Indeed, this effect of scene variation is predicted by the model anytime the semantic values for size, type, and color are ordered as: $x_{\text{size}} \leq  x_{\text{type}} < x_{\text{color}}$. If, on the other hand, $x_{\text{type}}$ is greater than $x_{\text{color}}$, the probability of redundantly mentioning color is close to zero and does not differ between variation conditions (in those cases, color mention reduces, rather than adds, information about the target). 

To further explore the scene variation effect predicted by RSA, we turn again to \figref{fig:sizesufficient}. Here, the target item is the small blue pin and there are two distractor items: a big blue pin and a big red pin. Thus, for the purpose of establishing unique reference, size is the sufficient dimension and color the insufficient dimension. 
We can measure scene variation as the proportion of distractor items that do not share the value of the insufficient feature with the target, that is, as the number of distractors $n_{\textrm{diff}}$ that differ in the value of the insufficient feature divided by the total number of distractors $n_{\textrm{total}}$:
\begin{equation*}
	\textrm{scene variation} = \frac{n_{\textrm{diff}}}{n_{\textrm{total}}}
\end{equation*}
In \figref{fig:sizesufficient}, there is one distractor that differs from the target in color (the big red pin) and there are two distractors in total.  Thus, $\textrm{scene variation} = \frac{1}{2} = .5$. In general, this measure of scene variation is minimal when all distractors are of the same color as the target, in which case it is 0. Scene variation is maximal when all distractors except for one (in order for the dimension to remain insufficient for establishing reference) are of a different color than the target. That is, scene variation may take on values between 0 and $\frac{n_{\textrm{total}} - 1}{n_{\textrm{total}}}$.\footnote{Some readers might find this unintuitive: shouldn't scene variation be maximal when there is an equal number of same and different colors? Or when the different colors are also all different from one another? As discussed in \sectionref{sec:modified}, there are many ways of quantifying (different aspects of) scene variation. We choose to explore this aspect of variation here as a reasonable first step; RSA makes predictions for other kinds of variation that would be equally straightforward to test.}

\begin{figure}
\centering
\includegraphics[width=.9\textwidth]{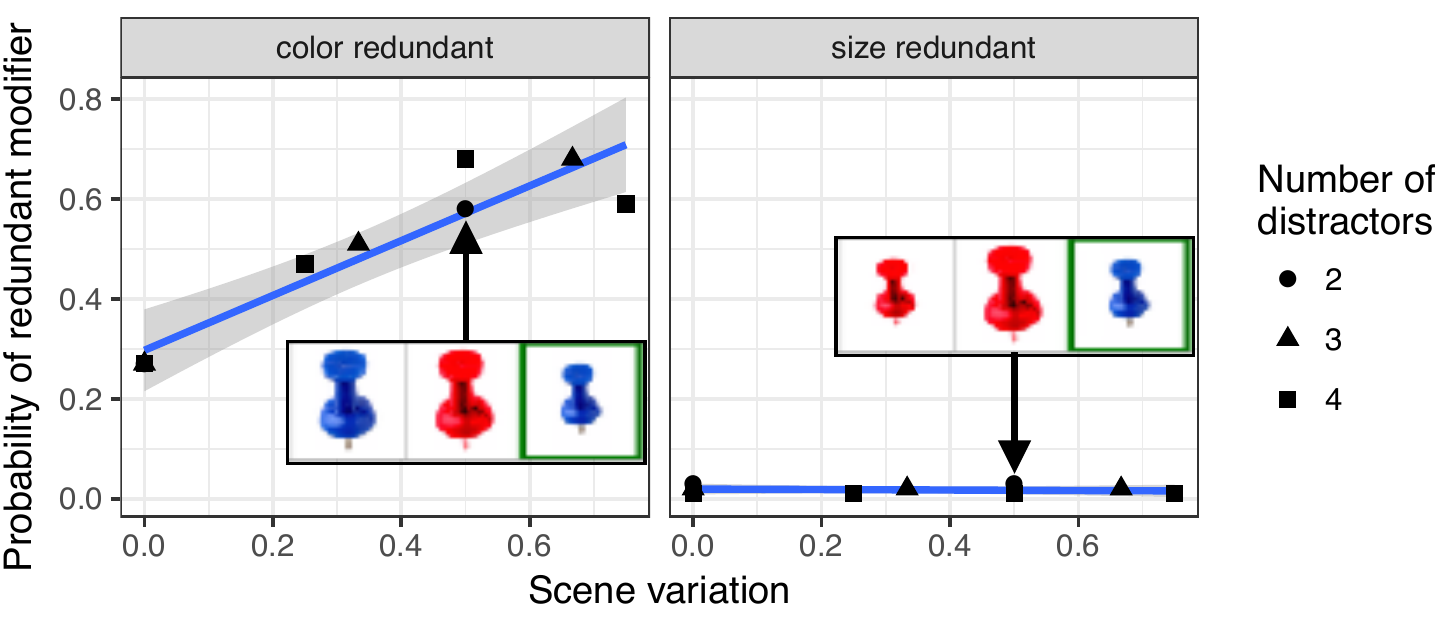}
\caption{Predicted probability of redundant utterance (\emph{small blue pin}) as a function of scene variation when size is sufficient (and color redundant, left) and when color is sufficient (and size redundant, right), for $\beta_i = 30$, $ \beta_c = c(u_{\textrm{size}}) = c(u_{\textrm{color}}) = 1$, $x_{\text{size}} = .8$, $x_{\text{color}} = .999$. Linear smoothers overlaid.}
\label{fig:numdistractors}
\end{figure}

Using the same parameter values as above, 
we generate model predictions for size-sufficient and color-sufficient contexts, manipulating scene variation by varying number of distractors (2, 3, or 4) and number of distractors that don't share the insufficient feature value. The resulting model predictions are shown in \figref{fig:numdistractors}. The predicted probability of redundant adjective use is largely (though not completely) correlated with scene variation.
Redundant adjective use increases with increasing scene variation when size is sufficient (and color redundant), but not when color is sufficient (and size redundant). 
The latter prediction depends, however, on the actual semantic value of color---with slightly lower semantic values for color, the model predicts small increases in redundant size use. 
In general: increased scene variation is predicted to lead to a greater increase in redundant adjective use for less noisy adjectives.

RSA with a continuous semantics thus captures the qualitative effects of color-size asymmetry and scene variation in production of redundant expressions, and it makes quantitative predictions for both. Testing these quantitative predictions, however, will require more data. 
In the remainder of the paper, we quantitatively evaluate cs-RSA on new datasets capturing the phenomena described in the Introduction (\tableref{tab:effects}): modifier type and scene variation effects on modified referring expressions, typicality effects on color mention, and the choice of taxonomic level of reference in nominal choice.

\section[]{Experiment 1: size and color modifiers under different scene variation conditions}
\label{sec:rsaevaluationbasicscene}

Adequately assessing the explanatory value of RSA with continuous semantics requires evaluating how well it does at predicting the probability of various types of utterances occurring in large datasets of naturally produced referring expressions. 
While we showed in \sectionref{sec:modelkoolen} that cs-RSA \emph{qualitatively} predicts the pattern of overmodification under scene variation, we now test the model's \emph{quantitative} predictions more rigorously in an interactive web-based reference game paradigm.
We then perform a Bayesian data analysis to both assess how likely the model is to generate the observed data -- i.e., to obtain a measure of model quality -- and to explore the posterior distribution of parameter values -- i.e., to understand whether the asymmetries in adjectives' semantic values and/or costs explored in the previous section are validated by the data.

\subsection{Method}

\paragraph{Participants}

We recruited 58 pairs of participants (116 participants total) over Amazon's Mechanical Turk who were each paid \$1.75 for their participation.\footnote{We aim to pay Mechanical Turk workers at a rate of \$12 - \$14.} Data from another 7 pairs who prematurely dropped out of the experiment and who could therefore not be compensated for their work, were also included. Here and in all other experiments reported in this paper, participants' IP address was limited to US addresses and only participants with a past work approval rate of at least 95\%  were accepted. 

\paragraph{Procedure}

\begin{figure}
\begin{subfigure}{.5\textwidth}
\includegraphics[width=\textwidth]{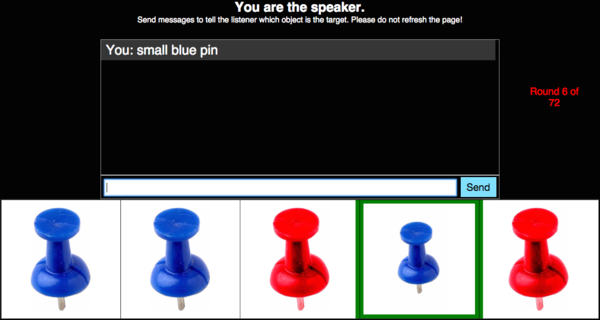}
\caption{Speaker's perspective.}
\label{fig:speakerpersp}
\end{subfigure}
\begin{subfigure}{.5\textwidth}
\includegraphics[width=\textwidth]{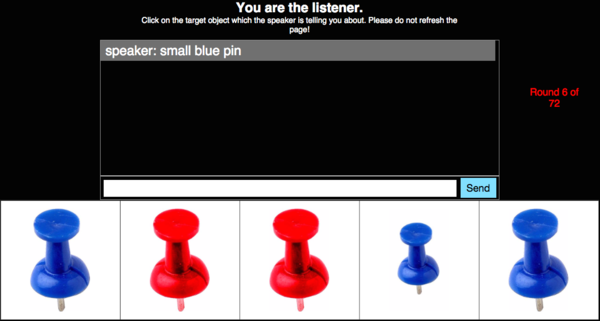}
\caption{Listener's perspective.}
\label{fig:listenerpersp}
\end{subfigure}
\caption{Example displays from the  (a) speaker's and the (b) listener's perspective on a \emph{size-sufficient 4-2} trial.}
\label{fig:speakerlistenerperspective}
\end{figure}

Participants were paired up through a real-time multi-player interface \cite{Hawkins15_RealTimeWebExperiments}. 
One participant was assigned the speaker role and one the listener role. 
Before continuing to the experiment, participants were required to correctly answer a series of questions about the experimental procedure (see \appref{app:numdistractors}). 
On each trial, both participants saw the same array of objects in independently randomized locations.
One of these objects was privately designated as the \emph{target} object to the speaker, and marked by a thick border (see \figref{fig:speakerlistenerperspective}).
The speaker's task was to use an unrestricted chat box to send a message communicating the target to the listener, who subsequently clicked an object to make a response. 
Both participants then received feedback about whether the intended referent was selected and advanced to the next trial.
They were explicitly told that using locative modifiers (like \emph{left} or \emph{right}) would be useless because the order of objects on their partner's screen would be different than on their own screen.
For natural interaction, we allowed both speakers and listeners to write freely in the chat window at any point, but listeners could only click on an object to advance to the next trial after the speaker sent an initial message. 
At the end of the experiments, participants completed a questionnaire in which they indicated whether their native language was English, whether they thought their partner was human, and how much they liked their partner.

\paragraph{Materials}

Participants proceeded through 72 trials. Of these, half were critical trials of interest and half were filler trials. On critical trials, we varied which feature was sufficient for uniquely establishing reference, the total number of objects in the array, and  the number of objects that shared the insufficient feature with the target. 

Objects varied in color and size. On 18 trials, color was sufficient for establishing reference. On the other 18 trials, size was sufficient. \figref{fig:speakerlistenerperspective} shows an example of a size-sufficient trial. We further varied the amount of variation in the scene by varying the number of distractor objects in each array (2, 3, or  4) and the number of distractors that did share the redundant feature value with the target. That is, when size was sufficient, we varied the number of distractors that shared the same color as the target. This number had to be at least one, since otherwise the redundant property would have been sufficient for uniquely establishing reference, i.e.~mentioning it would not have been redundant. Each total number of distractors was crossed with each possible number of distractors that shared the redundant property, leading to the following nine conditions: \emph{2-1, 2-2, 3-1, 3-2, 3-3, 4-1, 4-2, 4-3,} and \emph{4-4}, where the first number indicates the total number and the second number the shared number of distractors. Each condition occurred twice with each sufficient dimension. Objects never differed in type within one array (e.g., all objects are pins in \figref{fig:speakerlistenerperspective}) but always differed in type across trials. Each object type could occur in two different sizes and two different colors. We used photo-realistic objects of intuitively fairly typical colors. The 36 different object types and the colors they could occur with are listed in \appref{app:itemtypes}.

Fillers were target trials from Exp.~2, a replication of \citeA{GrafEtAl2016}. Each filler item contained a three-object grid. None of the filler objects occurred on target trials. Objects stood in various taxonomic relations to each other and required neither size nor color mention for unique reference. See \sectionref{sec:nominal} for a description of these materials.

\paragraph{Data pre-processing and exclusion}

We collected data from 2177 critical trials. Because we did not restrict participants' utterances in any way, they produced many different kinds of referring expressions. Testing the model's predictions required, for each trial, classifying the produced utterance as an instance of a \emph{color}-only mention (e.g., \emph{blue pin}), a \emph{size}-only mention (e.g., \emph{big pin}), or a redundant \emph{color-and-size} mention (e.g., \emph{big blue pin}). To this end we applied a semi-automatic data pre-processing procedure in which a script first checked whether the speaker's utterance contained a color or size term. In a second step,  one of the authors (CG)  manually checked and, if necessary, corrected the automatic classification. If no classification was possible, the trial was excluded. After exclusions, 2076 cases entered the analysis.  See \appref{sec:preprocessing} for details on the pre-processing procedure.

\subsection{Behavioral results}
\label{sec:modelempiricalresults}

\begin{figure}
\centering
\includegraphics[width=.8\textwidth]{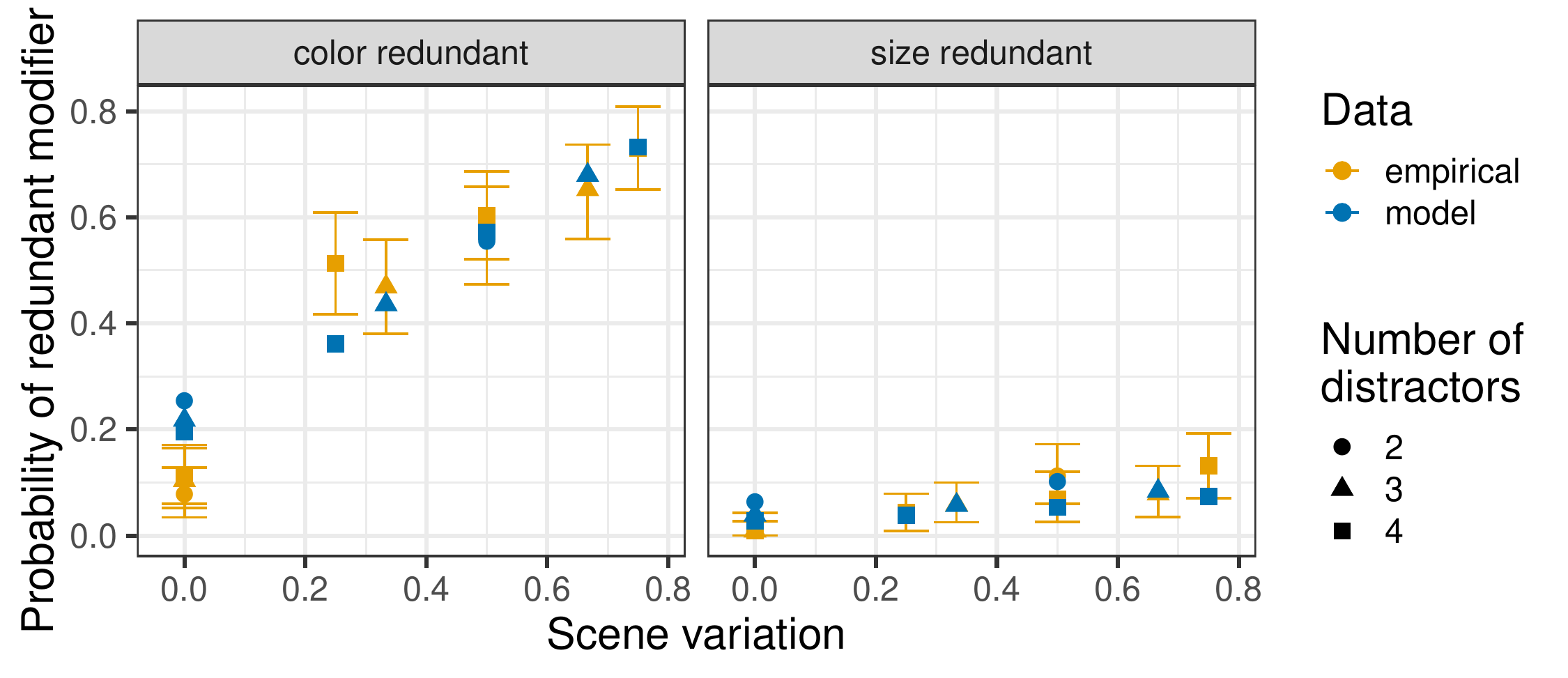}
\caption{Empirical redundant utterance proportions  (orange)  alongside point-wise maximum a posteriori (MAP) estimates of the RSA model's posterior predictives for redundant utterance probability (blue) as a function of scene variation in the color redundant (left) and size redundant (right) condition. Here and in all following plots, error bars indicate 95\% bootstrapped confidence intervals.}
\label{fig:exp1results}
\end{figure}

Proportions of redundant \emph{color-and-size} utterances are shown in \figref{fig:exp1results} alongside model predictions (to be explained further in \sectionref{sec:modifiermodeleval}). There are three main questions of interest: first, do we replicate the color/size asymmetry in probability of redundant adjective use? Second, do we replicate the previously established effect of increased redundant color use with increasing scene variation? Third, is there an effect of scene variation on redundant size use and if so, is it smaller compared to that on color use, as is predicted under asymmetric semantic values for color and size adjectives?

We addressed all of these questions by conducting a single mixed effects logistic regression analysis predicting redundant over minimal adjective use from fixed effects of sufficient property (color vs.~size), scene variation (proportion of distractors that do not share the insufficient property value with the target), and the interaction between the two.\footnote{All mixed effects analyses reported in this paper were conducted with the \verb+lme4+ package \cite{lme4} in R \cite{R}.} All predictors were centered before entering the analysis. The model included the most sophisticated random effects structure that allowed the model to converge: by-speaker and by-item random intercepts.

We observed a main effect of sufficient property, such that speakers were more likely to redundantly use color than size adjectives ($\beta = 3.54$, $SE = .22$, $p < .0001$), replicating the much-documented color-size asymmetry. We further observed a main effect of scene variation, such that redundant adjective use increased with increasing scene variation ($\beta = 4.62$, $SE = .38$, $p < .0001$). Finally, we also observed a significant interaction between sufficient property and scene variation ($\beta = 2.26$, $SE = .74$, $p < .003$). Simple effects analysis revealed that the interaction was driven by the scene variation effect being smaller in the \emph{color-sufficient} condition ($\beta = 3.49$, $SE = .65$, $p < .0001$) than in the \emph{size-sufficient} condition ($\beta = 5.75$, $SE = .38$, $p < .0001$), as predicted if size modifiers are noisier than color modifiers. That is, while the \emph{color-sufficient} condition indeed showed a scene variation effect---and as far as we know, this is the first demonstration of an effect of scene variation on redundant size use---this effect was tiny compared to that of the \emph{size-sufficient} condition.\footnote{In order to address convergence issues with \verb+lmer+ when specifying the maximal random effects structure -- i.e., by-speaker and by-item random intercepts and slopes for all fixed effects and their interactions -- we ran a Bayesian binomial mixed effects model with weakly informative priors using the \verb+brms+ package \cite{brms} that included the same fixed effects structure as the lmer model and the maximal random effects structure. The results were qualitatively identical, yielding  evidence for main effects of redundant feature (posterior mean $\beta$ = 5.91, $95\%$ CI = $[$4.15,8.10$]$, $p(\beta > 0)$ = .98), scene variation (posterior mean $\beta$ = 6.18, $95\%$ CI = $[$4.30,8.24$]$, $p(\beta > 0)$ = 1), and their interaction (posterior mean $\beta$ = 3.31, $95\%$ CI = $[$-0.54,7.23$]$, $p(\beta > 0)$ = .96).}

\subsection{Model evaluation}
\label{sec:modifiermodeleval}

In order to evaluate RSA with continuous semantics we conducted a Bayesian data analysis.  This allowed us to  simultaneously generate model predictions and infer likely parameter values, by conditioning on the observed production data (coded into \emph{size}, \emph{color}, and \emph{size-and-color} utterances as described above) and integrating over the five free parameters. To allow for differential costs for size and color, we introduce separate cost weights ($\beta_{c(\textrm{size})}, \beta_{c(\textrm{color})}$) applying to size and color mentions, respectively, in addition to semantic values for color and size ($x_{\textrm{color}}$, $x_{\textrm{size}}$) and an informativeness parameter $\beta_i$. We assumed uniform priors for each parameter: $x_{\textrm{color}}, x_{\textrm{size}} \sim \mathcal{U}(0,1)$,  $\beta_{c(\textrm{size})}, \beta_{c(\textrm{color})} \sim \mathcal{U}(0,40)$, $\beta_i  \sim \mathcal{U}(0,40)$.
Inference for the cognitive model was exact. We used Markov Chain Monte Carlo (MCMC) with a burn-in of 10000 and lag of 10 to draw 2000 samples from the joint posteriors on the five free parameters.

\begin{figure}
\centering
\includegraphics[width=.6\textwidth]{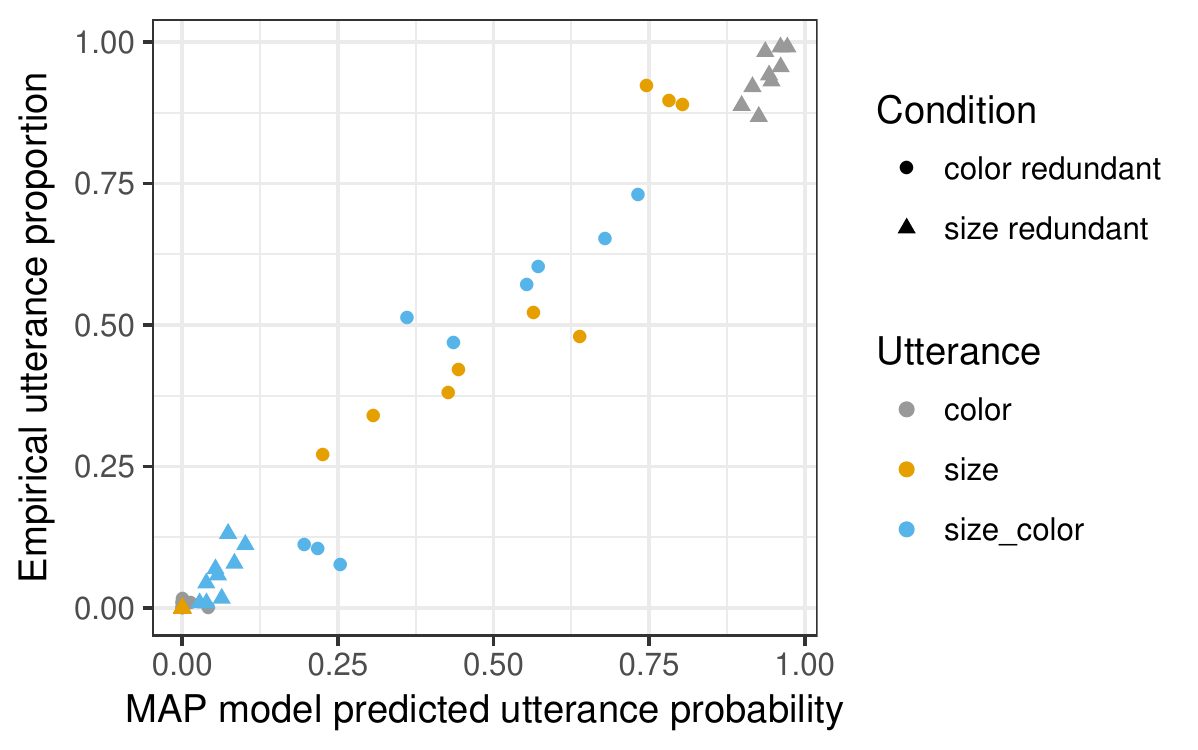}
\caption{Scatterplot of empirical utterance proportions against point-wise maximum a posteriori (MAP) estimates of the RSA model's posterior predictives. Each dot represents a condition mean.}
\label{fig:exp1predictives}
\end{figure}

Point-wise maximum a posteriori (MAP) estimates of the model's posterior predictives for just redundant utterance probabilities are shown alongside the empirical data in \figref{fig:exp1results}. In addition, MAP estimates of the model's posterior predictives for each combination of utterance, sufficient dimension, number of distractors, and number of different distractors (collapsing across different items) are plotted against all empirical utterance proportions in \figref{fig:exp1predictives}. At this level, the model achieves a correlation of $r = .99$. Looking at results additionally on the by-item level yields a correlation of $r = .85$ (this correlation is expected to be lower both because each item contains less data, and because we did not provide the model any means to refer differently to, e.g., \emph{combs} and \emph{pins}). The model thus does a very good job of capturing the quantitative patterns in the data. 

\begin{figure}
\centering
\includegraphics[width=\textwidth]{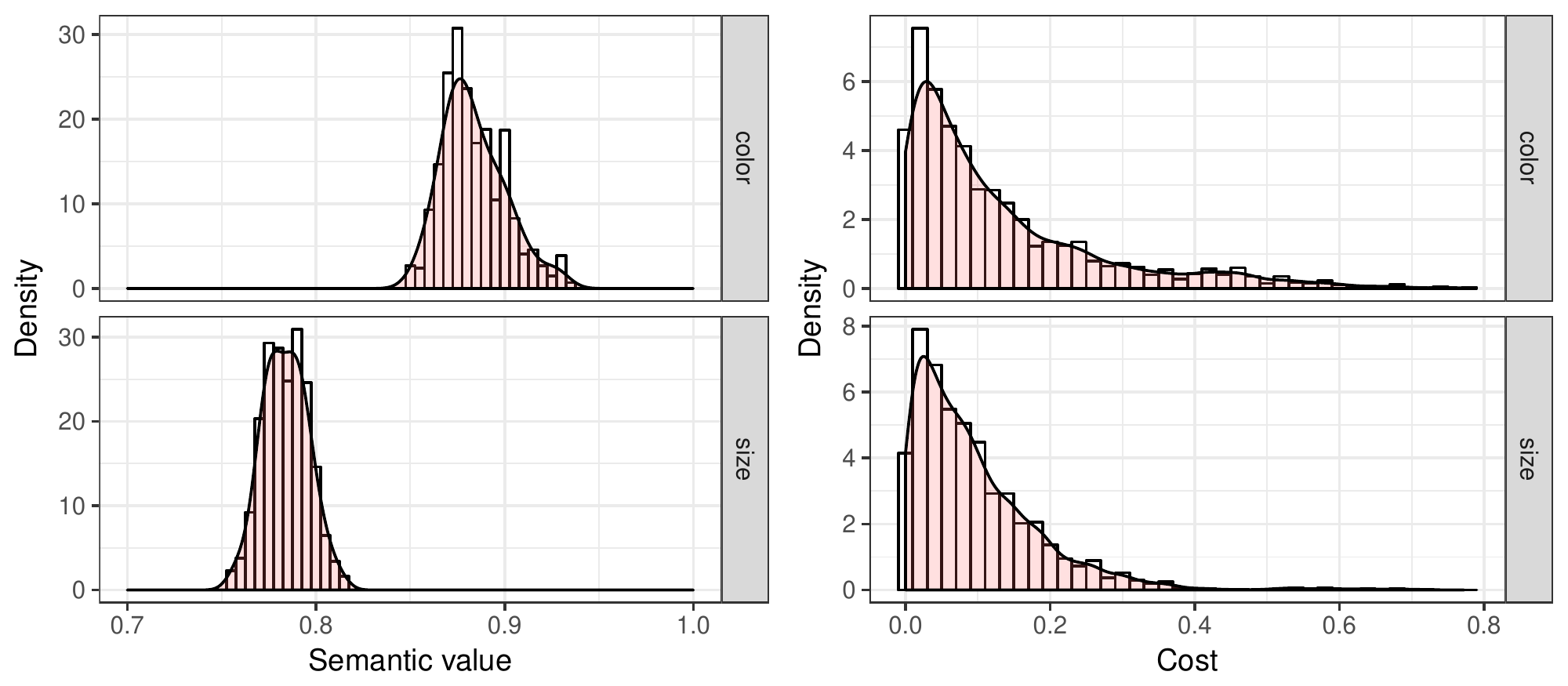}
\caption{Posterior model parameter distributions for semantic value (left column) and cost (right column), separately for color (top row) and size (bottom row) modifiers. Maximum a posteriori (MAP)  $x_{\textrm{size}}$ = 0.79, 95\% highest density interval (HDI) = [0.76,0.80]; MAP $x_{\textrm{color}}$ = 0.88, HDI = [0.85,0.92]; MAP $\beta_{c(\textrm{size})}$ = .02, HDI = [0, 0.26]; MAP $\beta_{c(\textrm{color})}$ = 0.03, HDI = [0,0.45].}
\label{fig:modifierparamposteriors}
\end{figure}

Posteriors over parameters are shown in \figref{fig:modifierparamposteriors}. Crucially, the semantic value of color is inferred to be higher than that of size -- there is no overlap between the 95\% highest density intervals (HDIs) for the two parameters. That is, size modifiers are inferred to be noisier than color modifiers. The  high inferred $\beta_i$ (MAP $\beta_i$ = 31.4, HDI = [30.7,34.5]) suggests that this difference in semantic value contributes substantially to the observed color-size asymmetries in redundant adjective use and that speakers are maximizing quite strongly. As for cost, there is a lot of overlap in the inferred weights of size and color modifiers, which are both skewed very close to zero, suggesting that a cost difference (or indeed any cost at all) is neither necessary  to obtain the color-size asymmetry and the scene variation effects, nor justified by the data. Recall further that we already showed in \sectionref{sec:modifiedmodel}  that the color-size asymmetry in redundant adjective use requires an asymmetry in semantic value and cannot be reduced to cost differences. An asymmetry in cost only serves to further enhance the asymmetry brought about by the  asymmetry in semantic value, but cannot carry the redundant use asymmetry on its own.

\subsection{Discussion}
\label{sec:modifierdiscussion}

In this section we reported a new dataset of freely produced referring expressions that replicated the well-documented color-size asymmetry in redundant adjective use, the effect of scene variation on redundant color use, and showed a novel effect of scene variation on redundant size use. We also showed that cs-RSA provides an excellent fit to these data. In particular, the crucial element in obtaining the color-size asymmetry in overmodification is that size adjectives be noisier than  color adjectives, captured in RSA via a lower semantic value for size compared to color. The effect is that color adjectives are more informative than size adjectives when controlling for the number of distractors that each would rule out under a Boolean semantics. Asymmetries in the cost of the adjectives were not attested, and would only serve to further enhance the modification asymmetry resulting from the asymmetry in semantic value. In addition, we showed that asymmetric effects of scene variation on overmodification straightforwardly fall out of cs-RSA: scene variation leads to a greater increase in overmodification with less noisy modifiers because these modifiers (colors) on average provide more information about the target.

While we defer a broader discussion of the possible psychological and linguistic interpretations of continuous semantic values to the General Discussion in \sectionref{sec:gd}, it is worth reflecting on \emph{why} size adjectives may be inherently noisier than color adjectives. Color adjectives are typically treated as  \emph{absolute adjectives} while size adjectives are inherently \emph{relative} \cite{Pechmann1989,kennedy2005}. That is, while both size and color adjectives are vague, size adjectives are arguably context-dependent in a way that color adjectives are not -- whether an object is big depends inherently on its comparison class; whether an object is red does not.\footnote{This is not entirely true, as has been repeatedly pointed out  \cite<e.g.,>{cohen1984models}: red hair has a very different color than red wine, which in turn has a different color from a red bell pepper. If presented out of context, only the last red is likely to be judged as red. For discussion of the complex semantics of color terms, see \citeNP{Kennedy2010, Rothschild2009, Szabo2001}. 
For our purposes, it suffices that one can give a color judgment but not a size judgment for an object presented in isolation.} In addition, color as a property has been claimed to be inherently salient in a way that size is not \cite{Arts2011, VanGompel2019}. Finally, color adjectives are rated as less subjective than size adjectives \cite{scontras2017}. 
All of this evidence suggests that the use of size adjectives may be more likely to vary across speakers and contexts than color.

Critically, our explanation of these phenomena departs from those offered by previous theories. 
\citeA{Pechmann1989} was the first to take the color-size asymmetry as evidence for speakers following an \emph{incremental} strategy of object naming.
That is, speakers initially start to articulate an adjective denoting a feature that listeners can quickly and easily recognize (i.e., color) before they have fully inspected the display and extracted the sufficient dimension. 
Another explanation appeals to \emph{saliency} considerations: speakers may produce modifiers that denote features that are reasonably easy for the listener to perceive, so that, even when a feature is not fully distinguishing in context, it at least serves to restrict the number of objects that could plausibly be considered the target. Indeed, there has been some support for the idea that overmodification can be beneficial to listeners by facilitating target identification \cite{Arts2011, rubiofernandez2016, Paraboni2007}. 
The effect of scene variation on propensity to overmodify has typically been explained as the result of the demands imposed on visual search: in low-variation scenes, it is easier to discern the discriminating dimensions than in high-variation scenes, where it may be easier to simply start naming features of the target that are salient \cite{Koolen2013}. 

Finally, there have been various attempts to capture the color-size asymmetry in computational natural language generation models. The earliest contenders for models of definite referring expressions like the Full Brevity algorithm \cite{Dale1989} or the Greedy algorithm \cite{Dale1989} focused only on discriminatory value -- that is, an utterance's informativeness -- in generating referring expressions. This is equivalent to the very simple interpretation of Grice's Quantity maxim, and consequently these models demonstrated the same inability to capture the color-size asymmetry: they only produced the minimally specified expressions. Subsequently, the Incremental algorithm \cite{dale1995} incorporated a preference order on features, with color ranked higher than size. The order is traversed and each encountered feature included in the expression if it serves to exclude at least one further distractor. This results in the production of overinformative color but not size adjectives. However, the resulting asymmetry is much greater than that evident in human speakers, and is deterministic rather than exhibiting the probabilistic production patterns that human speakers exhibit. 

More recently, the PRO model \cite{VanGompel2019} has sought to integrate the observation that speakers seem to have a preference for including color terms with the observation that a preference does not imply the deterministic inclusion of said color term. In PRO, the uniquely distinguishing property (if there is one) is  first selected deterministically. In additional steps, additional properties are added probabilistically, depending on both a salience parameter associated with the additional property and a parameter capturing speakers' eagerness to overmodify. If both properties are uniquely distinguishing, a property is selected probabilistically depending on its associated salience parameter. The second step proceeds as before. This model successfully captures speakers' overmodification patterns in contexts with one target and two distractors, in the choice of two properties (color, size) and three properties (color, size, border presence). 
While the PRO model -- the most state-of-the-art computational model of human production of modified referring expressions -- can capture the basic color-size asymmetry, it does not straightforwardly  account for the more subtle systematicity with which the preference to overmodify with color changes based on scene variation or object typicality, which we turn to next. 

\section[]{Experiment 2: color typicality in modified referring expressions}
\label{sec:colortypicality}

Our modeling results in Experiment 1 raise interesting questions regarding the status of the inferred semantic values: do color modifiers have inherently higher semantic values than size modifiers? Is the difference constant? What if the color modifier is a less well known one like \emph{mauve}? The way we have formulated the model thus far, there would indeed be no difference in semantic value between \emph{red} and \emph{mauve}. Moreover, the model is not equipped to handle potential object-level idiosyncracies such as the typicality effects discussed in \sectionref{sec:colortypicalityintro}: speakers are more likely to redundantly produce modifiers that denote atypical rather than typical object features, i.e., they are more likely to refer to a blue banana as a \emph{blue banana} rather than as a \emph{banana}, and they are more likely to refer to a yellow banana as a \emph{banana} than as a \emph{yellow banana} \cite{sedivy2003a, Westerbeek2015}. 

A natural first step toward explaining typicality effects is to introduce a more nuanced semantics for nouns in our model. 
In particular, we could imagine a continuous semantics in which \emph{banana} fits better (i.e. has a semantic value closer to 1 for) the yellow banana than the brown, and fits the brown better than the blue; specific such hypothetical values are shown in the first row of \tableref{tab:colorobjectfidelities}. Let us further assume that modifying the noun with a color adjective leads to uniformly high semantic values close to 1 for those objects that a simple truth-conditional semantics would return `true' for (see diagonal in \tableref{tab:colorobjectfidelities}) and a very low semantic value close to 0 for any utterance applied to any object that a simple truth-conditional semantics would return `false' for.

\begin{table}
\centering
\caption{Hypothetical semantic values for utterances (rows) as applied to objects (columns). Values where a Boolean semantics would return `true' are bolded.}
\begin{tabular}{l l l l l}
\toprule
 & yellow banana & brown banana & blue banana & other\\
\midrule
\emph{banana} & \bf{.9} & \bf{.35} & \bf{.1} & .01 \\
\midrule
\emph{yellow banana} & \bf{.99} & .01 & .01 & .01 \\
\emph{brown banana} & .01 & \bf{.99} & .01 & .01 \\
\emph{blue banana} & .01 & .01 & \bf{.99} & .01 \\
\midrule
other & .01 & .01 & .01 & \bf{.99} \\
\bottomrule
\end{tabular}
\label{tab:colorobjectfidelities}
\end{table}

\begin{figure}[bt!]
	\begin{subfigure}{.33\textwidth}
		\centering
		\includegraphics[width=\textwidth]{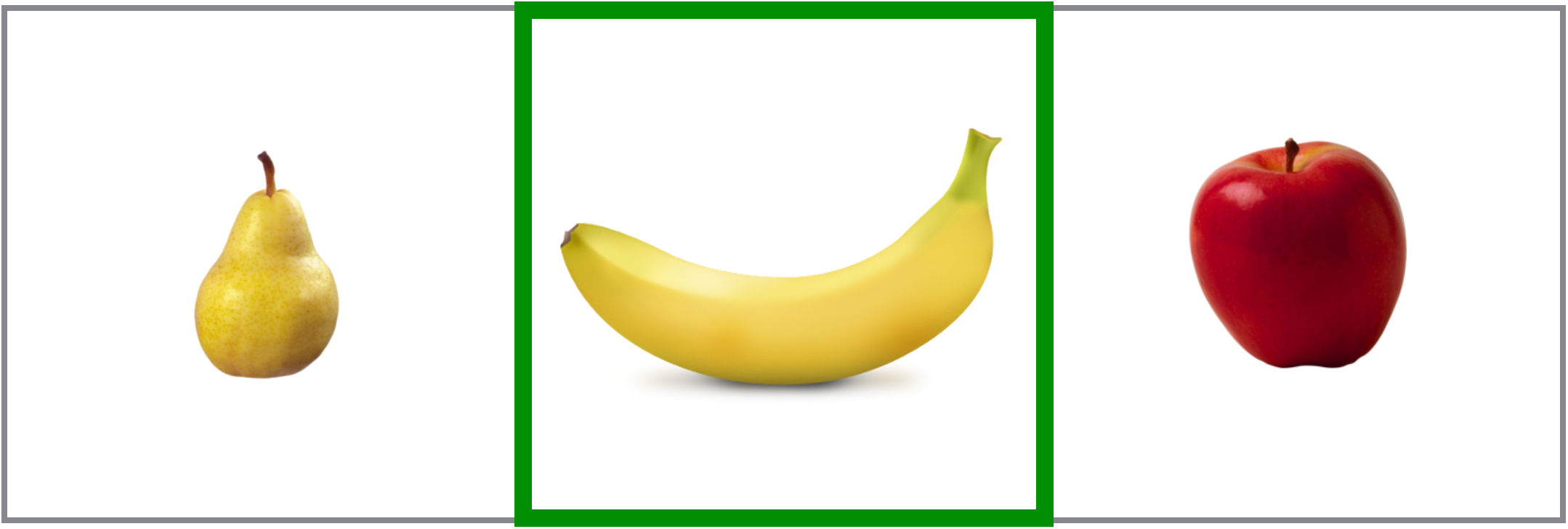}
		\caption{Typical color.}
		\label{fig:bananayellow}
	\end{subfigure}
	\begin{subfigure}{.33\textwidth}
		\centering
		\includegraphics[width=\textwidth]{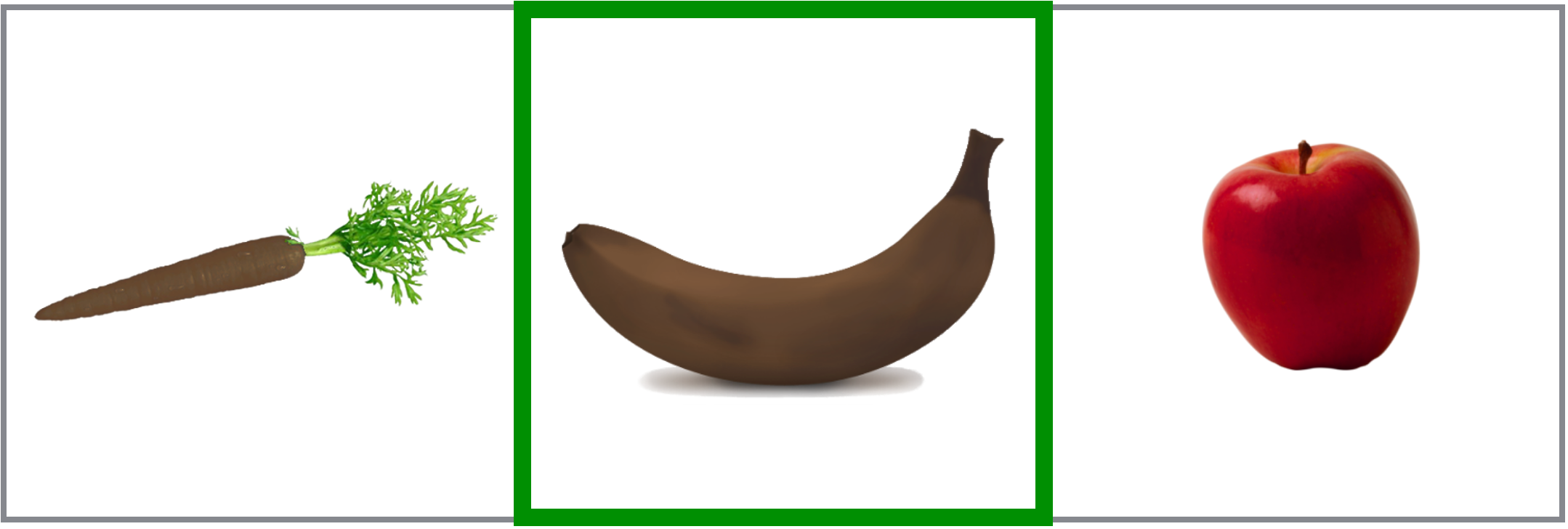}
		\centering
		\caption{Mid-typical color.}
		\label{fig:bananabrown}
	\end{subfigure}
	\begin{subfigure}{.33\textwidth}
		\centering
		\includegraphics[width=\textwidth]{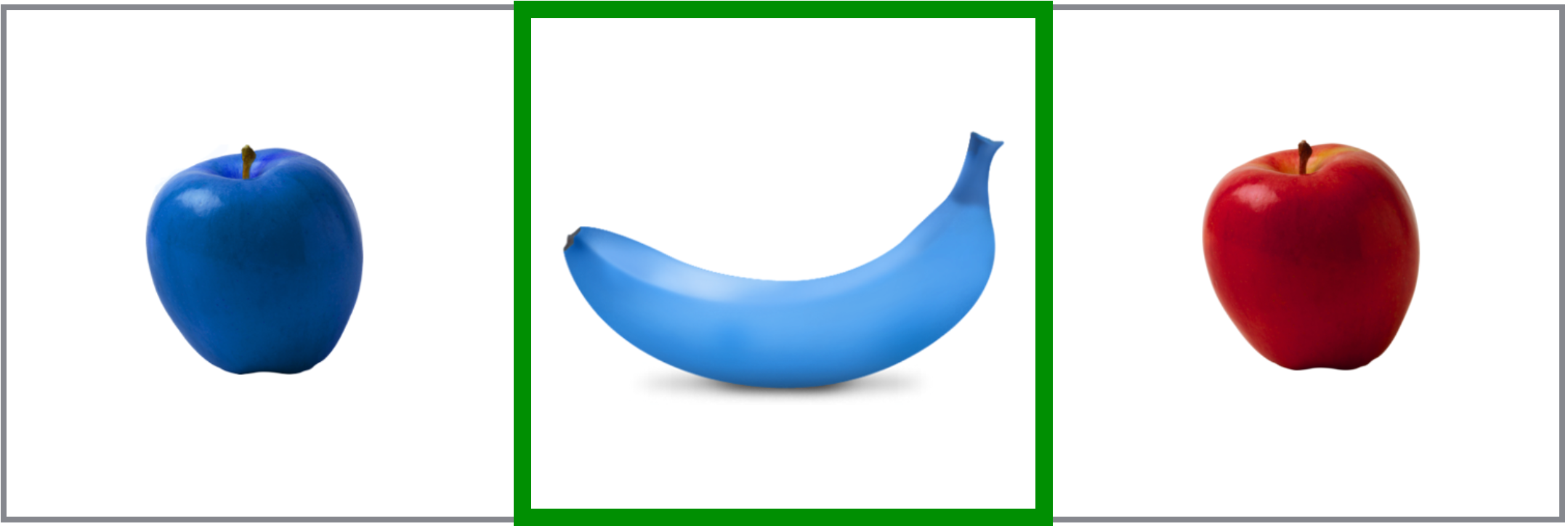}
		\caption{Atypical color.}
		\label{fig:bananablue}
	\end{subfigure}
	\caption{Three hypothetical contexts where color is redundant for referring to the target banana. Banana varies in typicality from left to right. Each context contains one distractor of the same color as the target, and one of a different color.}
	\label{fig:bananaexamples}
\end{figure}

The effect of running the speaker model forward with the standard literal listener treatment of the values in \tableref{tab:colorobjectfidelities} for the three contexts in \figref{fig:bananaexamples}, where \emph{banana} is the strictly sufficient utterance for unique reference (i.e., color is redundant under the standard view) is as follows:  with $\beta_i$ = 12 and $\beta_c$ = 5,\footnote{The results hold qualitatively for any informativeness weight $> 1$ and any cost weight $> 0$.} the resulting speaker probabilities for the minimal utterance \emph{banana} are .95, .29, and .04, to refer to the yellow banana, the brown banana, and the blue banana, respectively. In contrast, the resulting speaker probabilities for the redundant \emph{yellow banana}, \emph{brown banana}, and \emph{blue banana} are .05, .71, and .96, respectively. That is, redundant color mention increases with decreasing semantic value of the simple \emph{banana} utterance. 

This shows that cs-RSA can predict typicality effects if the semantic fit of the noun (and hence also of color-noun compounds) to an object is modulated by typicality. The reason the typicality effect arises is that, with the hypothetical values we assumed, the gain in informativeness between using the unmodified \emph{banana} and the modified \emph{COLOR banana} is greater in the blue than in the yellow banana case. 

This example is somewhat oversimplified. In practice, speakers sometimes mention an object's color without mentioning the noun. In the contexts presented in \figref{fig:bananaexamples} this does not make much sense because there is always a competitor of the same color present. In contrast, in the contexts in \figref{fig:condInf} and \figref{fig:condOverinf}, color alone disambiguates the target. This suggests that we should consider among the set of utterance alternatives not just the simple type mentions (e.g., \emph{banana}) and color-and-type mentions (e.g., \emph{yellow banana}), but also simple color mentions (e.g., \emph{yellow}). The dynamics of the model proceed as before.

An additional, more theoretically fraught, simplification concerns where typicality can enter into the semantics and how compositions proceeds. In the above, we have assumed that the semantic value of the modified expression is uniformly high, which is qualitatively what is necessary (and, as we will see below, empirically correct) in order for the typicality effects to emerge. However, there is no straightforward way to compositionally derive such uniformly high values from the semantic values of the nouns and the semantic values of the color modifiers, which we have not yet discussed. Indeed, compositional semantics of graded meanings is a well known problem for theories of modification \cite{kamp1995,Osherson1981}. Rather than try to solve it here, we note that RSA works at the level of whole utterances. Hence, if we can reasonably measure the semantic fit of each utterance to each possible referent, then cs-RSA will make predictions for production without the need to derive the semantic values compositionally. That is, if we can measure the typicality of the phrase \emph{blue banana} for a banana, we don't need to derive it from \emph{blue}, \emph{banana}, and a theory of composition. This separates pragmatic aspects of reference, which are the topic of this paper, from issues in compositional semantics, which are not; hence we will take this approach for experimentally testing the predictions of relaxed semantics RSA for typicality effects.

The stimuli for Exp.~1 were specifically designed to be realistic objects with low color-diagnosticity, so they did not include objects with low typicality values or large degrees of variation in typicality. This makes the dataset from Exp.~1 not well-suited for investigating typicality effects.\footnote{We did elicit typicality norms for the items in Exp.~1 and replicated the previously documented typicality effects on the four items that did exhibit variation in typicality. See \appref{sec:exp1typicality} for details.} We therefore conducted a separate production experiment in the same paradigm but with two broad changes: first, objects' color varied in typicality; and second, we did not manipulate object size, focusing only on color mention. This allows us to ask three questions: first, do we replicate the typicality effects reported in the literature -- that is, are less color-typical objects more likely to lead to redundant color use than more color-typical objects? Second, does cs-RSA with empirically elicited typicality values as proxy for a continuous semantics capture speakers' behavior? Third, does the semantic value depend only on typicality, or is there still a role for modifier type noise of the kind we investigated in the previous section? In addition, we can investigate the extent to which utterance cost, which we found not to play a role in the previous section, affects the choice of referring expression.

\subsection{Method}

\paragraph{Participants}
We recruited 61 pairs of participants (122 participants total) over Amazon's Mechanical Turk who were each paid \$1.70 for their participation.

\paragraph{Procedure}

The procedure of the reference game was identical to that of Exp.~1. 

\paragraph{Materials}

Each participant completed 42 trials. In this experiment, there were no filler trials, since pilot studies with and without fillers delivered very similar results. Each array presented to the participants consisted of three objects that could differ in type and color. One of the three objects functioned as a target and the other two as its distractors.

The stimuli were selected from seven color-diagnostic food items (apple, avocado, banana, carrot, pear, pepper, tomato), which all occurred in a typical, mid-typical and atypical color for that object. For example, the banana appeared in the colors yellow (typical), brown (midtypical), and blue (atypical). 
All items were presented as targets and as distractors. Pepper additionally occurred in a fourth color, which only functioned as a distractor due to the need for a green color competitor (as explained in the following paragraph). 

\begin{figure}[bt!]
	\begin{subfigure}{.5\textwidth}
		\centering
		\includegraphics[width=.8\textwidth]{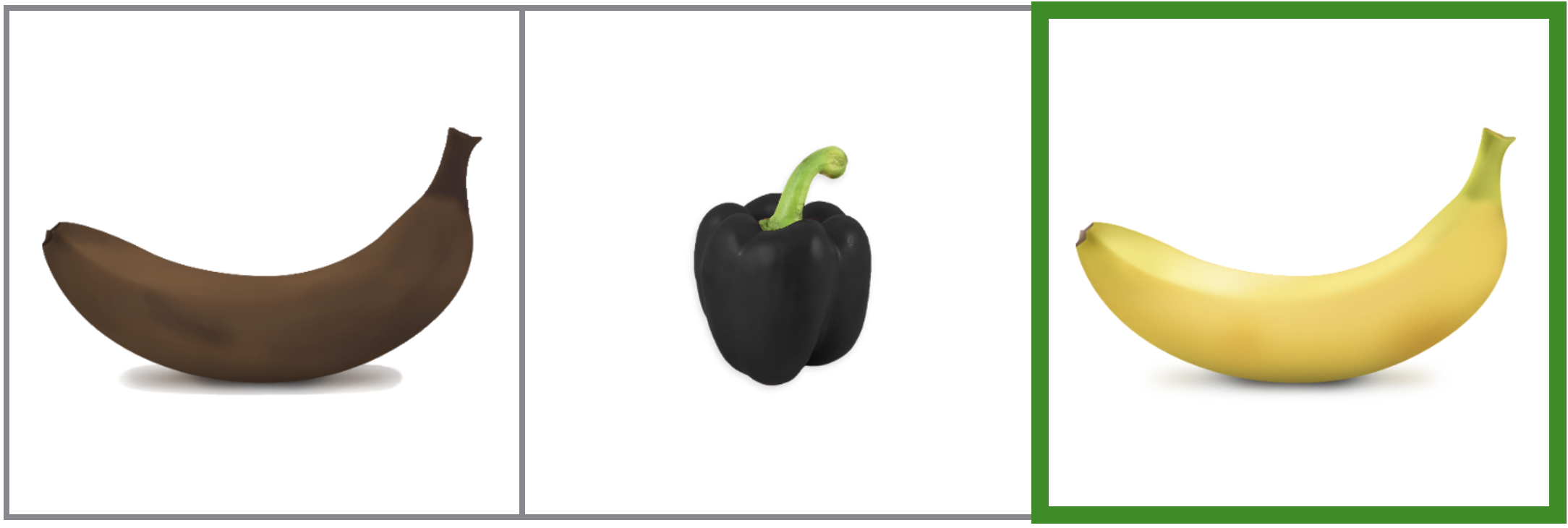}
		\caption{informative (without color competitor)}
		\label{fig:condInf}
	\end{subfigure}
	\begin{subfigure}{.5\textwidth}
		\centering
		\includegraphics[width=.8\textwidth]{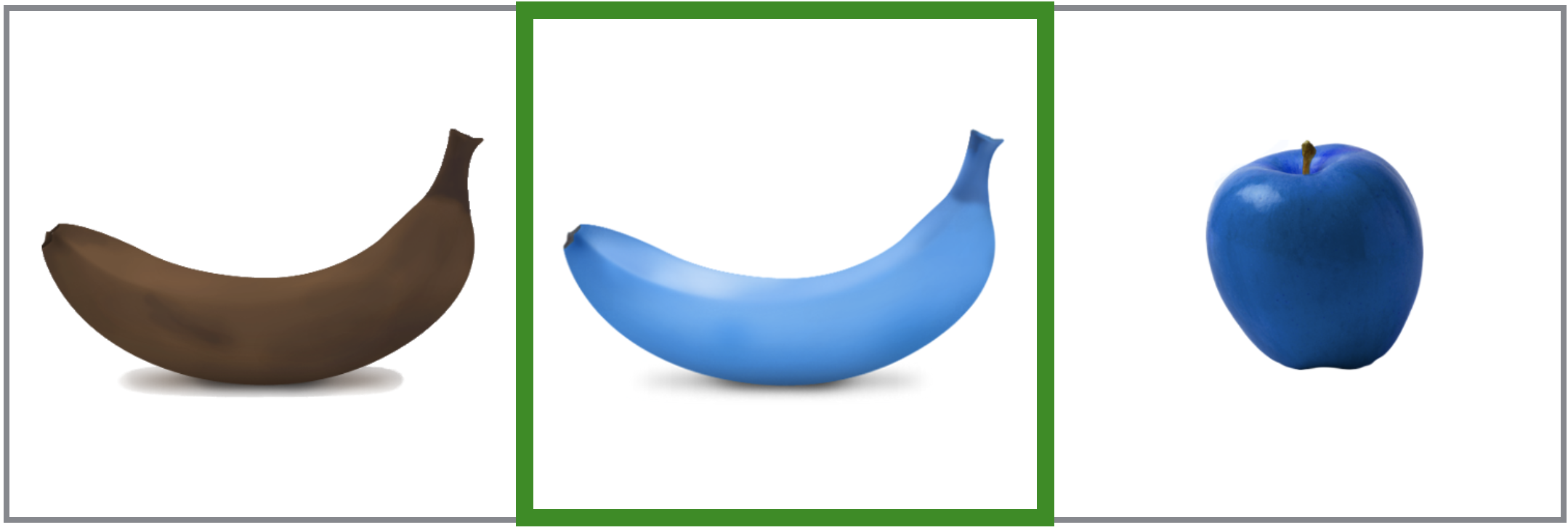}
		\centering
		\caption{informative-cc (with color competitor)}
		\label{fig:condInfcc}
	\end{subfigure}
	\begin{subfigure}{.5\textwidth}
		\centering
		\includegraphics[width=.8\textwidth]{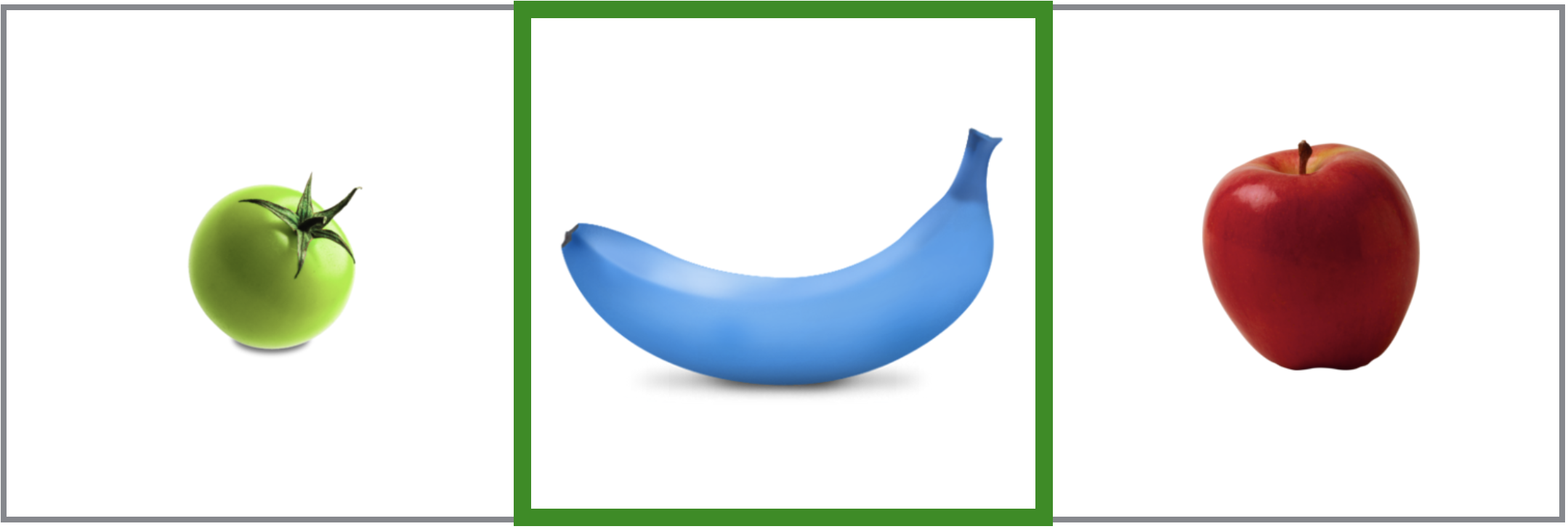}
		\caption{overinformative (without color competitor)}
		\label{fig:condOverinf}
	\end{subfigure}
	\begin{subfigure}{.5\textwidth}
		\centering
		\includegraphics[width=.8\textwidth]{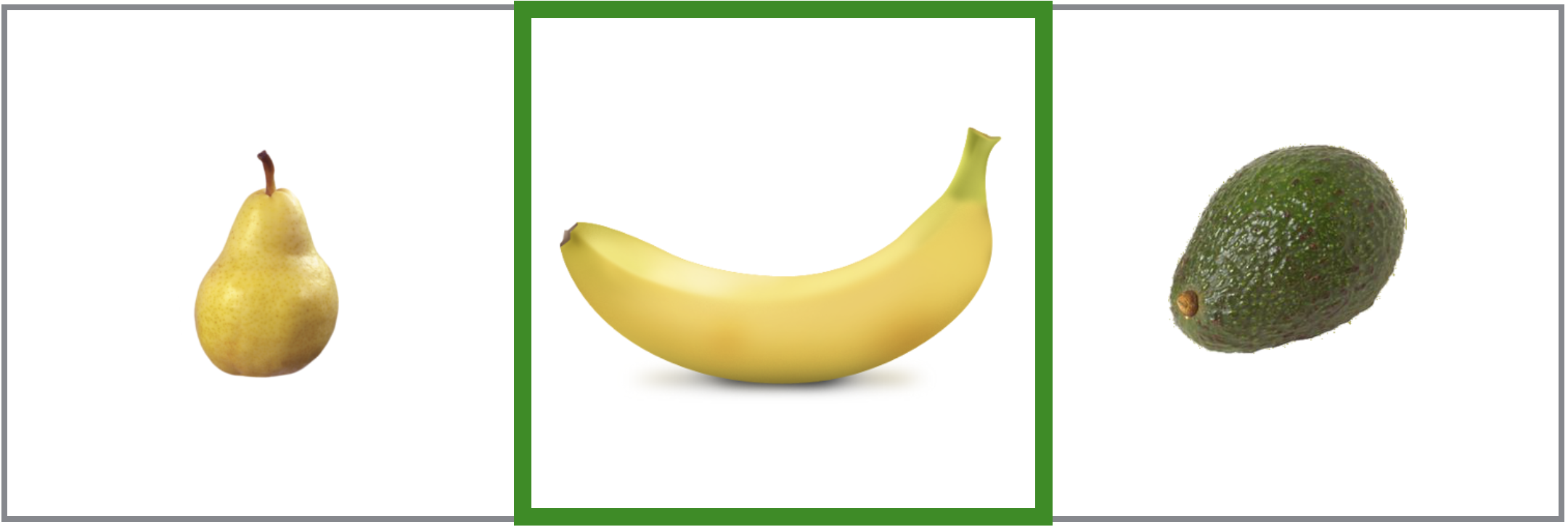}
		\centering
		\caption{overinformative-cc (with color competitor)}
		\label{fig:condOverinfcc}
	\end{subfigure}
	\caption{Examples of the four different context conditions in Exp.~2. They differed in the presence of an object of the same type (informative vs.~overinformative) and in the presence of another object of the same color as the target (with color competitor vs.~without color competitor). The thick border marks the intended referent.}
	\label{fig:conditions}
\end{figure}

We refer to the different context conditions  as ``informative'', ``informative-cc'', ``overinformative'', and ``overinformative-cc'' (see \figref{fig:conditions}). A context was ``overinformative'' (\figref{fig:condOverinf}) when mentioning the type of the item, e.g., banana, was sufficient for unambiguously identifying the target.  In this condition, the target never had a color competitor. This means that mentioning color alone (without a noun) was also unambiguously identifying. In contrast, in the overinformative condition with a color competitor (``overinformative-cc'', \figref{fig:condOverinfcc}), color alone was not sufficient. In the informative conditions, color and type mention were necessary for unambiguous reference. Again, one context type did (\figref{fig:condInf}) and one did not (\figref{fig:condOverinfcc}) include a color competitor among its distractors.

Each participant saw 42 different contexts. Each of the 21 items (color-type combinations) was the target exactly twice, but the context in which they occurred was drawn randomly from the four possible conditions mentioned above. In total, there were 84 different possible configurations (seven target food items, each of them in three colors, where each could occur in four contexts). Trial order was randomized.

\paragraph{Data pre-processing and exclusion}

We collected data from 1974 trials. The utterance produced on each trial was classified as belonging to one of the following categories: \textit{type-only} (e.g., \emph{banana}), \textit{color-and-type} (e.g., \emph{yellow banana}), and \textit{color-only} (e.g., \emph{yellow}). Referring expressions that could not be classified were excluded. See \appref{sec:preprocessing} for further details on exclusion criteria and the data pre-processing procedure. Overall, 1827 utterances entered the analysis.

\subsection{Typicality norming}
\label{sec:typicalitynormingcolor}

\begin{figure}[bt!]
	\subcaptionbox{color-and-type norming. \label{fig:colorobj}} {\includegraphics[width=2.1in]{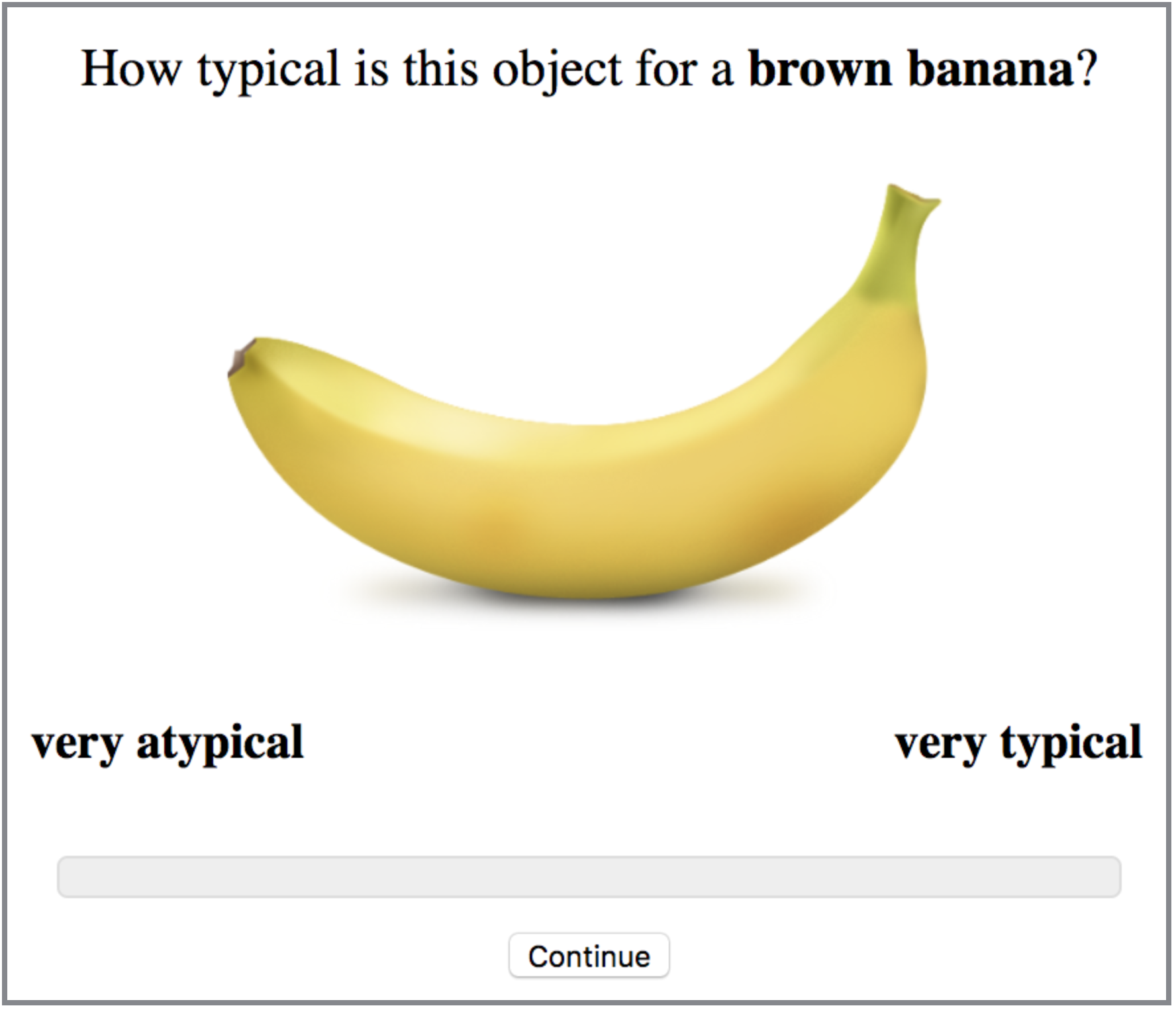}} \hfill
	\subcaptionbox{type-only norming. \label{fig:obj}} 
	{\includegraphics[width=2in]{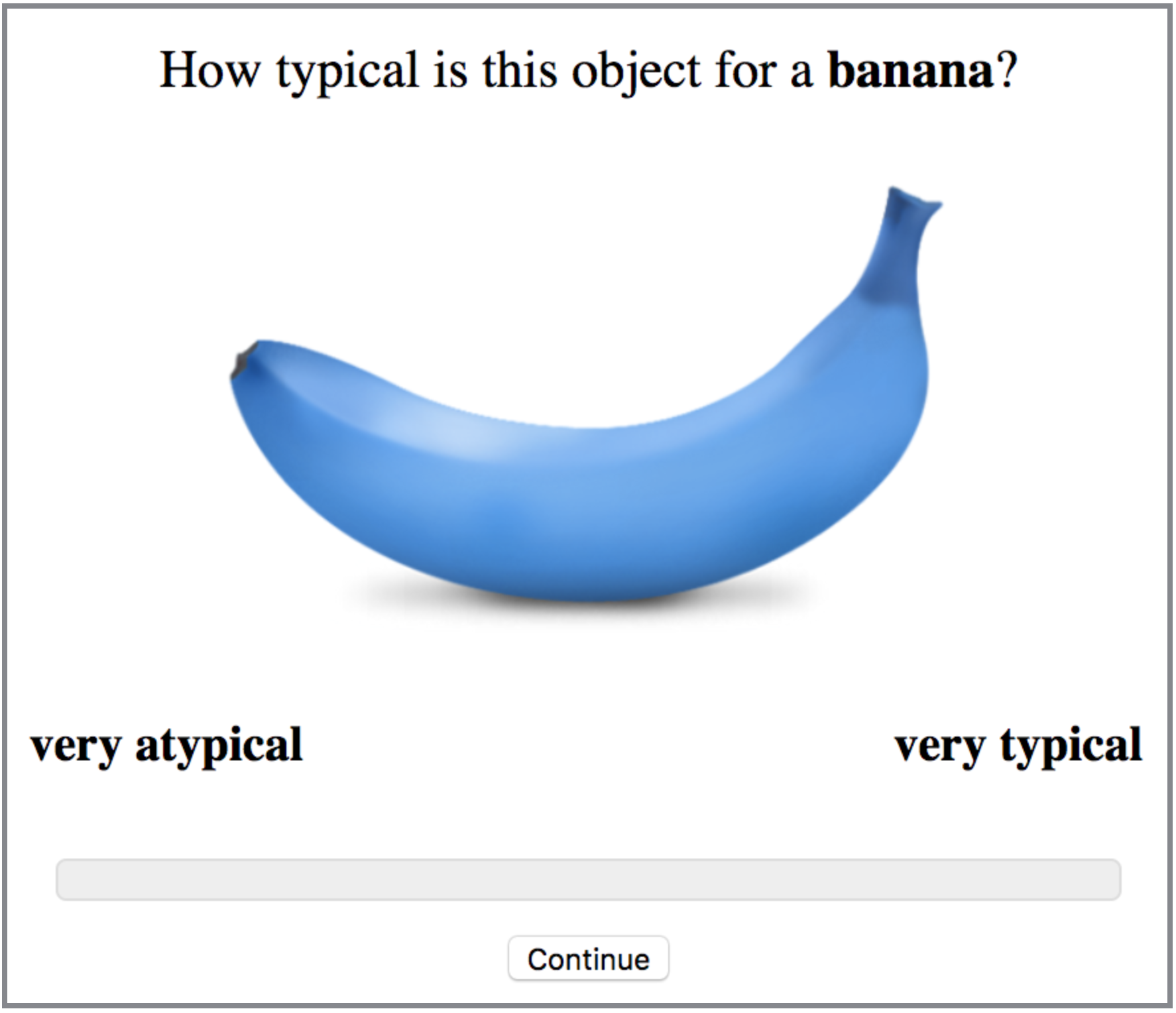}} \hfill
	\subcaptionbox{color-only norming. \label{fig:colorpatch}} {\includegraphics[width=2in]{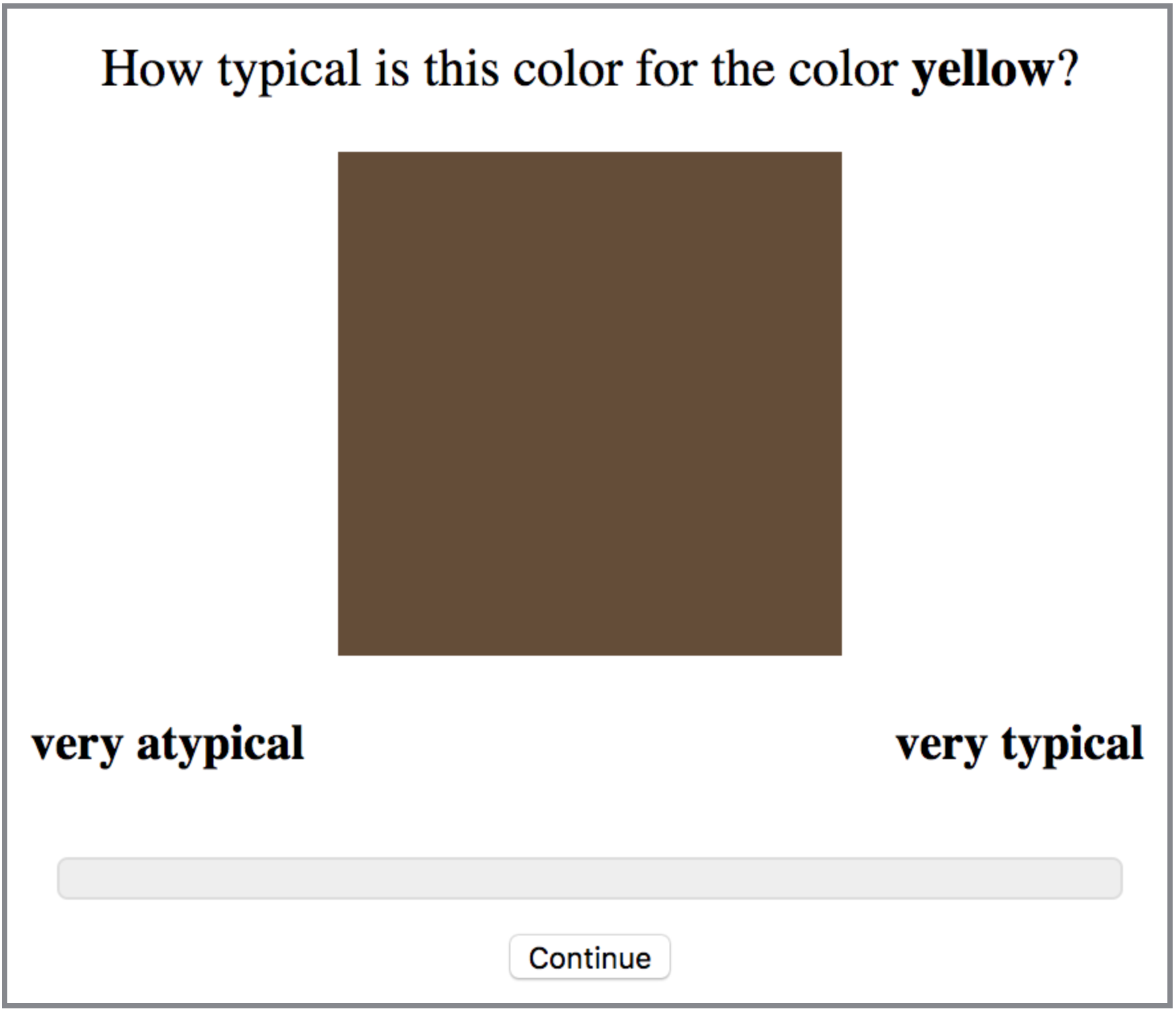}}
	\caption{Example stimuli exemplifying the three different typicality norming studies.}
	\label{fig:typ_norm}
\end{figure}

In order to test for typicality effects on the production data and to evaluate cs-RSA's performance, we collected empirical typicality values for each utterance/object pair in three separate studies. The first study collected typicalities for \emph{color-and-type}/object pairs (e.g., \emph{yellow banana} as applied to a yellow banana, a blue banana, an orange pear, etc., see \figref{fig:colorobj}). The second study collected typicalities for \emph{type-only}/object pairs (e.g., \emph{banana} as applied to a yellow banana, a blue banana, an orange pear, etc., \figref{fig:obj}). The third study collected typicalities for \emph{color}/color pairs (e.g., \emph{yellow} as applied to a color patch of the average yellow from the yellow banana stimulus or to a color patch of the average orange from the orange pear stimulus, and so on, for all other colors, \figref{fig:colorpatch}). 

On each trial of the \emph{type} or \emph{color-and-type} studies, participants saw one of the stimuli used in the production experiment in isolation and were asked: ``How typical is this object for a \emph{utterance}'', where \emph{utterance} was replaced by an utterance of interest. In the \emph{color} typicality study, they were asked ``How typical is this color for the color \emph{color}?'', where \emph{color} was replaced by one of the relevant color terms. They then adjusted a continuous sliding scale with endpoints labeled ``very atypical'' and ``very typical'' to indicate their response. A summary of the the three typicality norming studies is shown in Table~\ref{tab:normingoverview}.\footnote{The typicality elicitation procedure we employed here is somewhat different from that employed by \citeA{Westerbeek2015}, who asked their participants ``How typical is this color for this object?'' We did this because the semantic values that enter into the RSA model are best conceptualized as the typicality of an object as an instance of an utterance, rather than a feature-category relation.
See \appref{sec:exp1typicality} for a comparison of our question and the Westerbeek question as applied to typicality norms for the items in Exp.~1. In general, the \textsc{Type}-object values are highly correlated with the Westerbeek question values.} 

\begin{table}[bt!]
	\begin{tabular}{l l l l l l l}
		\toprule
		Utterances & Example & Images & Participants & Trials & Items & Excluded participants\\
		\midrule
		Adj Noun & \emph{yellow banana} & object & 174 & 110 & 484 & 14\\ 
		Noun & \emph{banana} & object & 75 & 90 & 154 & 1\\
		Adj & \emph{yellow} & color patch & 110 & 90 & 176 & None\\
		\bottomrule
	\end{tabular}
	\vspace{2mm}
	\caption{Overview of the typicality norming studies for Exp.~2. Column `Items' contains the number of unique utterance-object pairs that we elicited responses for.} 
	\label{tab:normingoverview}
\end{table}

\begin{table}[bt!]
	\caption{Mean typicalities for banana items. Combinations where Boolean semantics would return `true' are marked in boldface.}
\centering
	\begin{tabular}{r r r r r}
		\toprule
		& \multicolumn{3}{c}{Banana items} & Other \\
		Utterance & yellow & brown  & blue & \\ 
		\midrule
		\emph{banana} & \textbf{.98} & \textbf{.66} & \textbf{.42} & .05  \\
		\midrule
		\emph{yellow banana} & \textbf{.97} & .30 & .15 & .05 \\
		\emph{brown banana} & .22 & \textbf{.91} & .15 & .04\\
		\emph{blue banana} & .16 & .15 & \textbf{.92} & .06\\
		\midrule
		\emph{yellow} & \textbf{.77} & .05 & .06 & .09 \\
		\emph{brown} & .11 & \textbf{.87} & .01 & .12\\
		\emph{blue} & .06 & .06 & \textbf{.92} & .07\\		
		\bottomrule
	\end{tabular}
	\vspace{5mm}
	\label{tab:bananatypicalities}
\end{table}

\begin{figure}
	\centering
	\includegraphics[width=\textwidth]{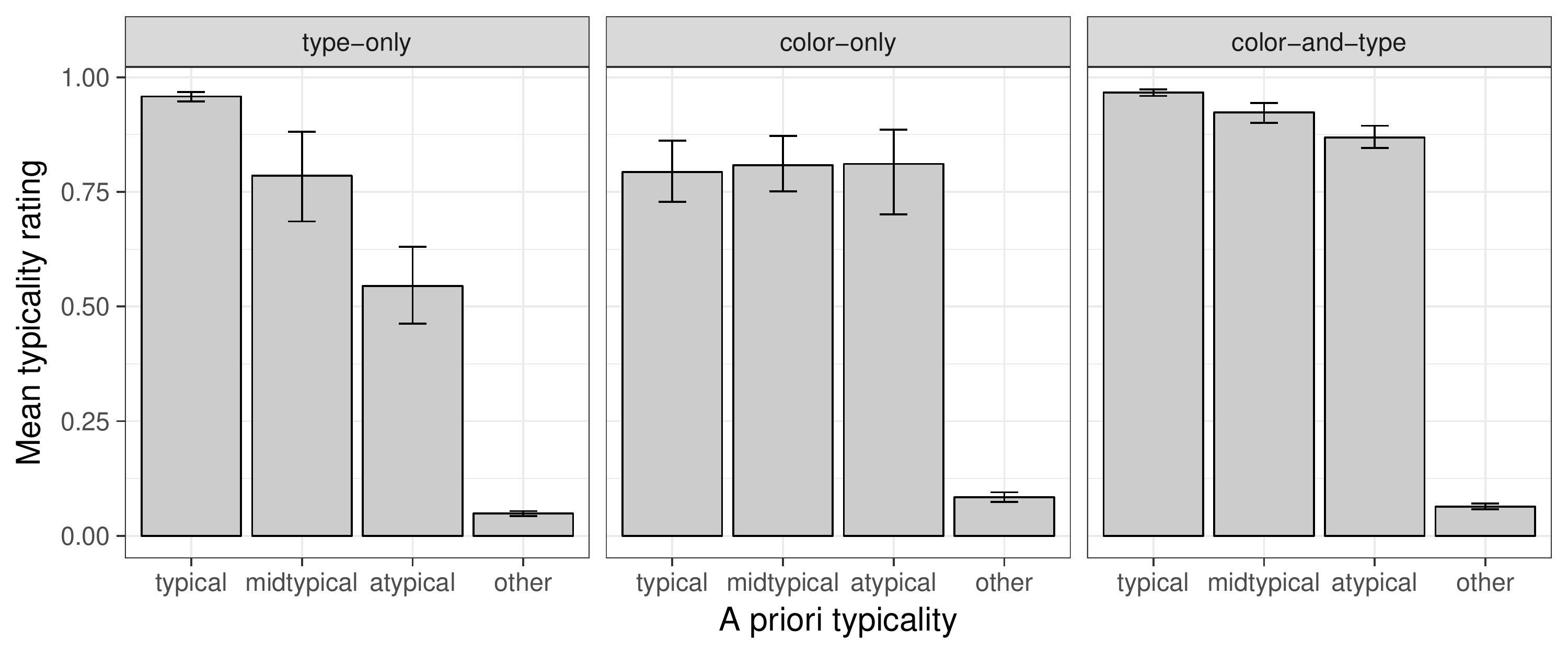}
	\caption{Mean typicality ratings for the three norming studies (type-only, color-only, color-and-type). The results are categorized according to the objects' a priori typicality as determined by the experimenters (yellow banana = typical, brown banana = midtypical, blue banana = atypical). The category \emph{other} comprises all utterance-object combinations where a Boolean semantics would return false (e.g. a pepper). Error bars indicate bootstrapped 95\% confidence intervals.}
	\label{fig:exp2colortypicalitymeans}
\end{figure}

Slider values were coded as falling between 0 (`very atypical') and 1 (`very typical'). 
For each utterance-object combination, we computed mean typicality ratings. 
As an example, the means for the banana items and associated color patches are shown in \tableref{tab:bananatypicalities}. 
The values exhibit the same gradient as those hypothesized for the purpose of the example in \tableref{tab:colorobjectfidelities}. 
The means for all items are visualized in \figref{fig:exp2colortypicalitymeans}.
Mean typicality values for utterance-object pairs obtained in the norming studies are used in the analyses and visualizations in the following.

\subsection{Behavioral results}
\label{sec:exp2resultsdisc}

\begin{figure}
\centering
	\begin{subfigure}{.85\textwidth}
		\centering
		\includegraphics[width=.8\textwidth]{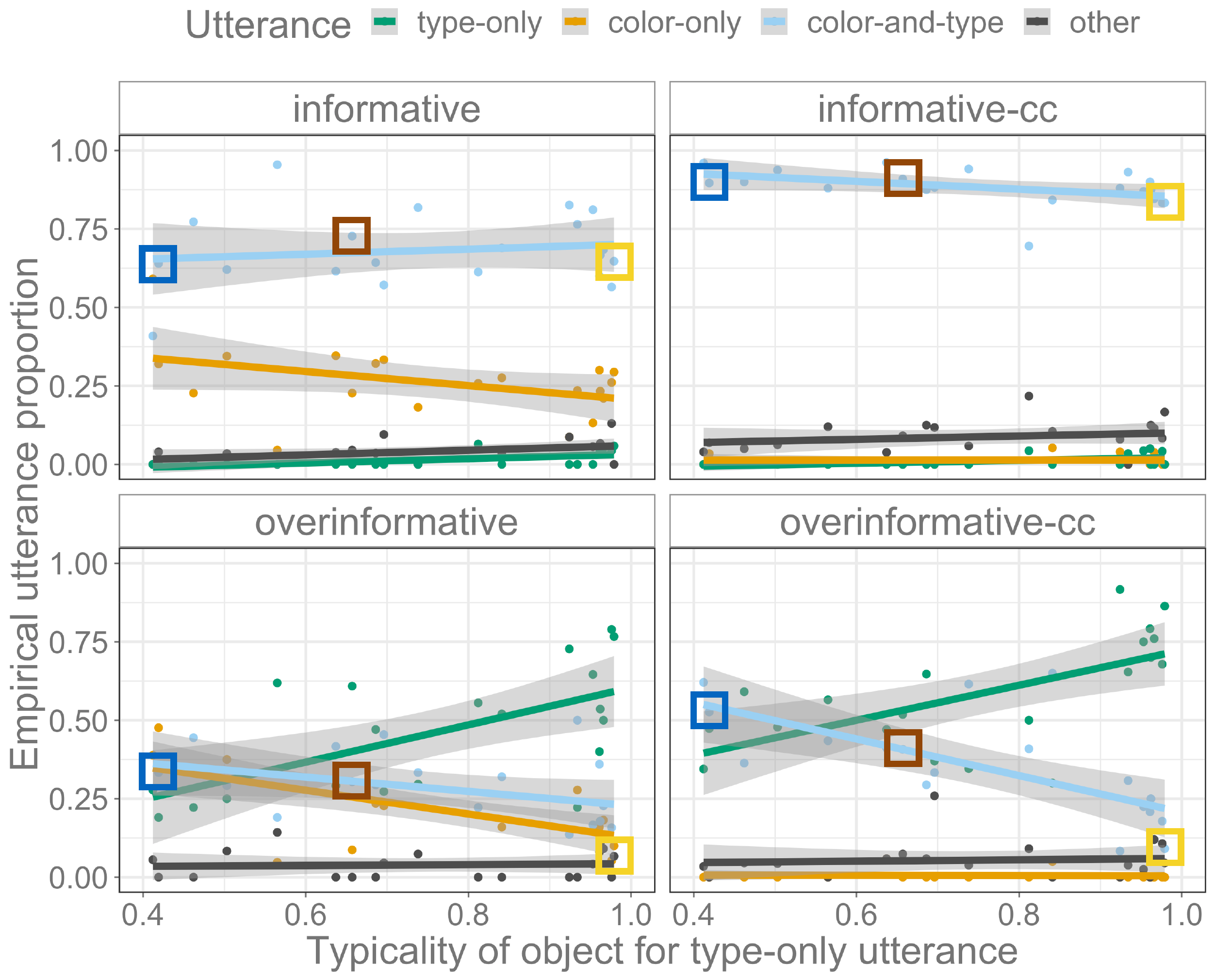}
		\caption{Empirical utterance proportions}
		\label{fig:exp2empirical}
	\end{subfigure}
	
	\begin{subfigure}{.85\textwidth}
		\centering
		\includegraphics[width=.77\textwidth]{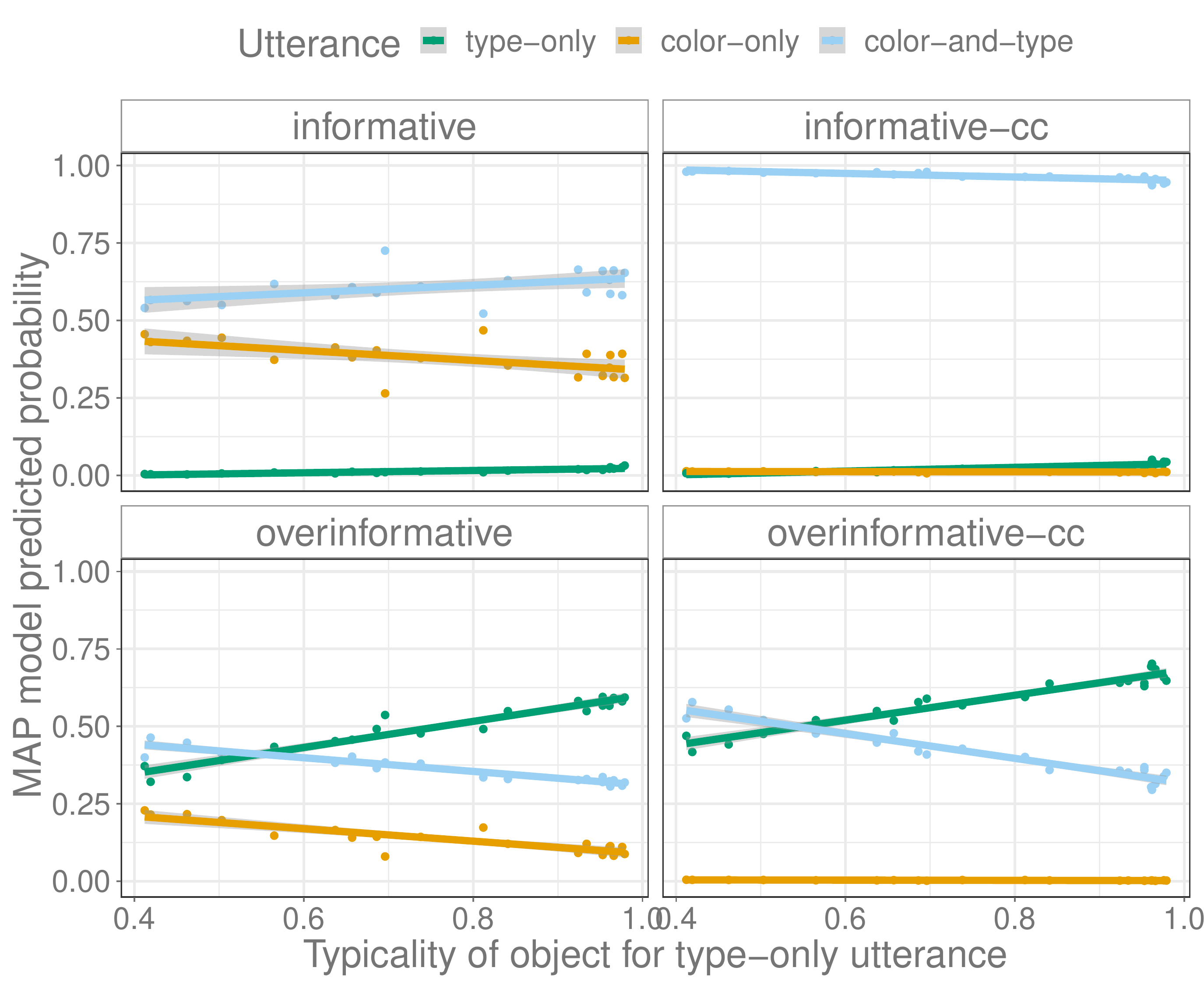}
		\caption{MAP model predicted utterance probabilities}
		\label{fig:exp2model}
	\end{subfigure}
\caption{(a) Empirical utterance proportions in Exp.~2 and (b) MAP model predicted utterance probabilities for each target as a function of mean object typicality for the type-only utterance (e.g., \emph{banana}). Color indicates utterance type: type-only (\emph{banana}), color-only (\emph{yellow}), color-and-type (\emph{yellow banana}), and other (\emph{funky carrot}). Facets indicate conditions (\emph{informative} vs.~\emph{overinformative}, color competitor present (\emph{cc}) or absent). Modified utterance data points for the banana items are circled in the banana's respective color in (a).}
\label{fig:exp2results}
\end{figure}

Proportions of type-only (\emph{banana}), color-and-type (\emph{yellow banana}), color-only (\emph{yellow}), and other (\emph{funky carrot}) utterances are shown in \figref{fig:exp2empirical} as a function of the described item's mean type-only (\emph{banana}) typicality. Visually inspecting just the explicitly marked \emph{yellow banana}, \emph{brown banana}, and \emph{blue banana} cases suggests a large typicality effect in the overinformative conditions as well as a smaller typicality effect in the informative conditions, such that color is less likely to be produced with increasing typicality of the object. 

The following questions are of interest. First, do we replicate the previously documented typicality effect on redundant color mention (as suggested by the visual inspection of the banana item)? Second, does typicality affect color mention even when color is informative (i.e., technically necessary for establishing unique reference)? Third, are speakers sensitive to the presence of color competitors in their use of color or are typicality effects invariant to the distractor items?

To address these questions we conducted a mixed effects logistic regression predicting color use from fixed effects of typicality, informativeness, and color competitor presence. We used the typicality norms obtained in the \emph{type}/object typicality elicitation study reported above (see \figref{fig:obj}) as the continuous typicality predictor. The informativeness condition was coded as a binary variable (color informative vs.~color overinformative trial) as was color competitor presence (absent vs.~present). All predictors were centered before entering the analysis. The model included by-speaker and by-item random intercepts, which was the most sophisticated random effects structure that allowed the model to converge.

We found a main effect of typicality, such that the more typical an object was for the type-only utterance, the lower the log odds of color mention ($\beta$ = -4.17, $SE$ = 0.45, $p <$ .0001), replicating previously documented typicality effects. Stepwise model comparison revealed that including interaction terms was not justified by the data, suggesting that speakers produce more typical colors less often even when the color is in principle necessary for establishing reference (i.e., in the informative conditions). This is notable: speakers sometimes call a yellow banana simply a \emph{banana} even when other bananas are present, presumably because they can rely on listeners drawing the inference that they must have meant the most typical banana. In contrast, blue bananas' color is always mentioned in the informative conditions.

There was also a main effect of informativeness, such that color mention was less likely when it was overinformative than when it was informative ($\beta$ = -5.56, $SE$ = 0.33, $p <$ .0001). Finally, there was a main effect of color competitor presence, such that color mention was more likely when a color competitor was absent  ($\beta$ = 0.71, $SE$ = 0.16, $p <$ .0001). This suggests that speakers are indeed sensitive to the contextual utility of color -- color typicality alone does not capture the full set of facts about color mention, as we already saw in \sectionref{sec:rsaevaluationbasicscene}.

\subsection{Model evaluation}
\label{sec:colorypicalitymodel}

We evaluated the cs-RSA model on the obtained production data from Exp.~2. In particular, we were interested in using model comparison to address the following issues:
First, can RSA using elicited typicality as the semantic values account for quantitative details of the production data?
Second, are typicality values sufficient, or is there additional utility in including a noise offset determined by the type of modifier, as was used in the previous section?
Third, does utterance cost explain any of the observed production behavior. 

While the architecture of the model remained the same as that of the model presented in \sectionref{sec:modifiedmodel}, we briefly review the minor necessary changes, some of which we already mentioned at the beginning of this section. These changes concerned the semantic values and the cost function.\footnote{See \tableref{tab:modeldiffs} for an overview of the models reported in the paper.} 

\subsubsection{Lexicon}

Previously, we considered only three utterance alternatives: \emph{color}, \emph{size}, and \emph{color-size}, collapsing over the precise values these took on.
Here, we no longer collapse over these values, including in the lexicon each possible color adjective, type noun, and combination of the two. 
This substantially increased the size of the lexicon to 37 unique utterances. For each combination of utterance $u$ and object $o$ that occurred in the experiment, we included a separate semantic value $x_{u,o}$, elicited in the norming experiments described in \sectionref{sec:typicalitynormingcolor} (rather than inferred as done for Exp.~1, to avoid overfitting).
For any given context, we assumed the utterance alternatives that correspond to the individually present features and their combinations. For example, for the context in \figref{fig:condOverinfcc}, the set of utterance alternatives was \emph{yellow, green, pear, banana, avocado, yellow pear, yellow banana}, and \emph{green avocado}. 

\subsubsection{Semantics}

We compared several choices of semantics for the model.
In the full \emph{fixed plus empirical semantics} version, we introduced a parameter $\beta_{\textrm{fixed}}$ interpolating between the empirically elicited typicality values ($\beta_{\textrm{fixed}}=0$) and the inferred type-level values as employed in \sectionref{sec:rsaevaluationbasicscene} ($\beta_{\textrm{fixed}}=1$).
The type-level values again consisted of one value for color terms and another for type terms, which are multiplied when the terms are composed in an utterance. 
In a lesioned \emph{empirical semantics} version, we set $\beta_{\textrm{fixed}}=0$ and only used the empirical values.
Conversly, in a lesioned \emph{fixed semantics} version, we set $\beta_{\textrm{fixed}}=1$ and only used the inferred type-level values. 
This allowed us to perform a nested model comparison, since the latter models are special cases of the first.

\subsubsection{Cost function}

For the purpose of evaluating the model in \sectionref{sec:rsaevaluationbasicscene} we inferred two constant costs (one for color and one for size), and found in the Bayesian Data Analysis that the role of cost in explaining the data was minimal at best. Here, we compared two different versions of utterance cost. In the \emph{fixed cost} model we treated cost the same way as in the previous section and included only a color and type level cost, inferred from the data. We then compared this model to an \emph{empirical cost model}, in which we included a more complex cost function. 
Specifically, we defined utterance cost $c(u)$ as follows:
\begin{equation} \label{eq:exp2cost}
c(u) = \beta_F\cdot p(u) + \beta_L\cdot l(u)
\end{equation}

Here, $p(u)$ is negative log utterance frequency, as estimated from the Google Books corpus (years 1950 to 2008); $l(u)$ is the mean empirical length of the utterance in characters in the production data (e.g., sometimes \emph{yellow} was abbreviated as \emph{yel}, leading to an $l(u)$ smaller than 6); $\beta_F$ is a weight on frequency; and $\beta_L$ is a weight on length. Both $p(u)$ and $l(u)$ were normalized to fall into the interval $[0,1]$.\footnote{Note that we changed the sign on frequency, which means that values closer to 1 in the normalized space reflect greater cost on both the length and the frequency dimension.} The empirical cost function thus prefers short and frequent utterances (e.g., \emph{blue}) over long and infrequent ones (\emph{turquoise-ish bananaesque thing}).
We compared both of these models to a simpler baseline in which utterances were assumed to have no cost.

\subsubsection{Model comparison}

To evaluate the effect of these choices of semantics and cost, we conducted a full Bayesian model comparison.
Specifically, we computed the Bayes Factor for each comparison, a measure quantifying the support for one model over another in terms of the relative likelihood they each assign to the observed data.
As opposed to classical likelihood ratios, which only use the maximum likelihood estimate, the likelihoods in the Bayes Factor integrate over all parameters, thus automatically correcting for the flexibility due to extra parameters (the ``Bayesian Occam's Razor''). 
Because it was intractable to analytically compute these integrals for our recursive model, we used Annealed Importance Sampling (AIS), a Monte Carlo algorithm commonly used to approximate these quantities.
To ensure high-quality estimates, we took the mean over 100 independent samples for each model, with each chain running for 30,000 steps.
The marginal log likelihoods for each model are shown in \tableref{tab:exp2-modelcomparison}. 
The best performing model used \emph{fixed plus empirical} semantics and did not include a cost term. 
Despite the greater number of parameters associated with adding the fixed semantics to the empirical semantics, the \emph{fixed plus empirical} semantics models were preferred across the board compared to their empirical-only ($BF = 3.7 \times 10^{48}$ for fixed costs, $BF = 2.1 \times 10^{60}$ for empirical costs, and $BF = 1.4 \times 10^{71}$ for no cost) and fixed-only counterparts ($BF=6.5 \times 10^{14}$ for fixed costs, $BF= 1.0 \times 10^{19}$ for empirical costs, and $BF = 1.06 \times 10^{15}$ for no cost).
In comparison, additional cost-related parameters were not justified, with $BF = 5.7 \times 10^{21}$ for no cost compared to fixed cost and $BF = 2.1 \times 10^{27}$ for compared to empirical cost.

\begin{table}
\caption{Marginal log likelihood for each model. Best model is in bold. Parentheses indicate number of free parameters.}
\centering
\begin{tabular}{l l c c l }
\toprule
& & \multicolumn{3}{c}{Semantics}\\
& & \emph{empirical} & \emph{fixed} & \emph{fixed plus empirical}\\
\midrule
\multirow{3}{*}{Cost} & \emph{empirical} & -1390.3 (4) & -1295.1 (6) & -1251.4 (7) \\
                                        & \emph{fixed} & -1350.3 (4) &  -1272.6 (6) & -1238.5 (7) \\
                                         & \emph{none} & -1352.2 (2) &  -1223.0 (4) & \textbf{-1188.4 (5)} \\ 
\bottomrule
\end{tabular}
\label{tab:exp2-modelcomparison}
\end{table}

\begin{figure}
\centering
\includegraphics[width=\textwidth]{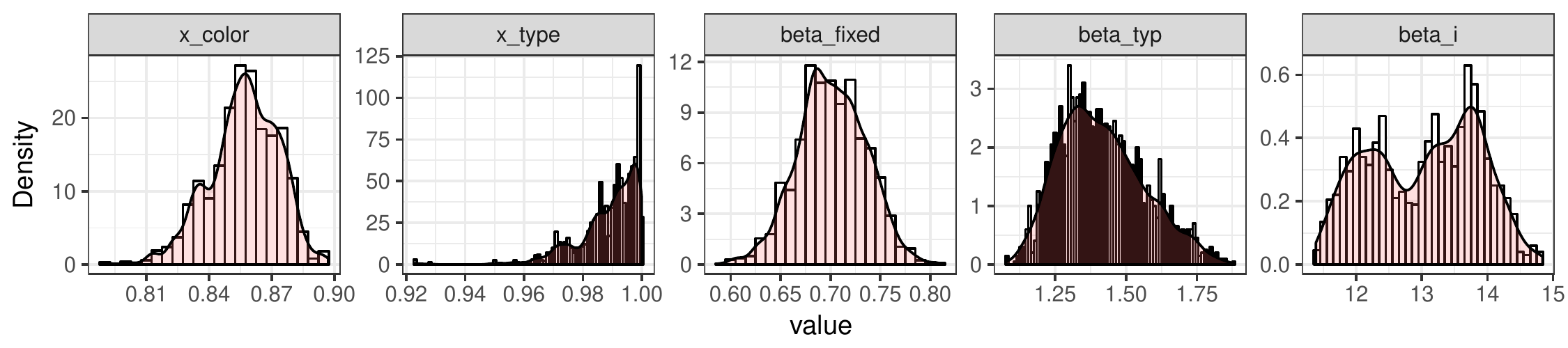}
\caption{For best model, posterior model parameter distributions for  interpolation weight on fixed vs.~empirical semantics $\beta_{\textrm{fixed}}$, informativity weight $\beta_i$, typicality weight $\beta_t$, and utterance type level semantic values for color $x_{\textrm{color}}$ and type $x_{\textrm{type}}$. Maximum a posteriori (MAP)  $\beta_{\textrm{fixed}}$ = 0.69, 95\% highest density interval (HDI) =  [0.64,0.77]; MAP $\beta_i$ = 13.74, HDI = [11.58,14.37]; MAP $\beta_t$ = 1.34, HDI = [1.19,1.75];  MAP  $x_{\textrm{color}}$ = 0.86, HDI = [0.82,0.89]; MAP $x_{\textrm{type}}$ = 0.998, HDI = [0.97,1.00].}
\label{fig:typparamposteriors}
\end{figure}

The correlation between empirical utterance proportions and the best model's MAP predictions at the by-item level was $r=.94$. Predictions for the best-performing model are visualized alongside empirical proportions in \figref{fig:exp2model}. The model successfully reproduces the empirically observed typicality effects in all four experimental conditions, with a reasonably good quantitative agreement. 
The interpolation weight between the fixed and empirical semantic values $\beta_{\textrm{fixed}}$ (\figref{fig:typparamposteriors}) is in the intermediate range: this provides evidence that a noisy truth-conditional semantics as employed in Exp.~1 is justified, but that taking into account graded category membership or typicality in an utterance's final semantic value is also necessary. 

There is one major, and interesting, divergence from the empirical data in conditions without color competitors. Here, \emph{color-and-type} utterances are systematically somewhat underpredicted in the informative condition, and systematically somewhat overpredicted in the overinformative condition. The reverse is true for \emph{color-only} utterances. It is worth looking at the posterior over parameters, shown in \figref{fig:typparamposteriors}, to understand the pattern. In particular, the utterance type level semantic value of type is inferred to be systematically higher than that of color, capturing that type utterances are less noisy than color utterances.\footnote{Interestingly, the inferred semantic value for color is very similar in absolute terms to that in Exp.~1.} An increase in \emph{color-only} mentions in the overinformative condition could be achieved by reducing the semantic value for type. However, that would lead to a further and undesirable increase in \emph{color-only} mentions in the informative condition as well. That is, the two conditions are in a tug-of-war with each other.

\subsection{Discussion}

In Experiment 2, we demonstrated that cs-RSA predicts color typicality effects in the production of referring expressions with only minimal extensions to support finer-grained prediction of item-by-item data (see \tableref{tab:modeldiffs} in the General Discussion for a more extensive comparison of the model details across sections).
This suggests that the dynamics at work in the choice of color vs.~size and in the choice of color as a function of the object's color typicality are very similar: speakers choose utterances by considering the fine-grained differences in information about the intended referent communicated by the ultimately chosen utterance compared to its competitor utterances. For noisier utterances (e.g., \emph{banana} as applied to a blue banana), including the `overinformative' color modifier is useful because it provides information. For less noisy utterances (e.g., \emph{banana} as applied to a yellow banana), including the color modifier is useless because the unmodified utterance is already highly informative with respect to the speaker's intention. These dynamics can sometimes even result in the color modifier being left out altogether, even when there is another---very atypical---object of the same type present, simply because the literal listener is expected to prefer the typical referent strongly enough.

Model comparison demonstrated the need for assuming a semantics that interpolates between a noisy truth-conditional semantics as employed in Exp.~1 and empirically elicited typicality values. 
This may reflect semantic knowledge that goes beyond graded category membership, additional effects of compositionality, or perhaps simply differences between our empirical typicality measure and the ``semantic fit'' expected by RSA models.
Perhaps surprisingly, we replicated the result from Exp.~1 that utterance cost does not add any predictive power, even when quantified via a more sophisticated cost function that takes into account an utterance's length and frequency.
In the next section, we move beyond the choice of modifier and ask whether cs-RSA provides a good account of  referring expression production more generally.

\section[]{Experiment 3: taxonomic level in unmodified referring expressions}
\label{sec:nominal}

In this section we investigate whether cs-RSA accounts for referring expression production beyond the choice of modifier. 
In particular, we focus on speakers' choice of taxonomic level of reference in nominal referring expressions. 
A particular object can be referred to at its subordinate (\emph{dalmatian}), basic (\emph{dog}), or superordinate (\emph{animal}) level, among other choices.
This choice of reference level is interestingly different from that of adding modifiers in that there is no additional word-level cost associated with being more specific -- the choice is between different one-word utterances, not between utterances differing in word count. 
Still, we hypothesized that similar factors may contribute: an expression's contextual informativeness, its cognitive cost \cite<short and frequent terms are preferred over long and infrequent ones,>{griffin1998,jescheniak1994}, and its typicality \cite<an utterance is more likely to be used if the object is a good instance of it,> {Jolicoeur1984}.

In order to evaluate cs-RSA for nominal choice, we proceeded as in \sectionref{sec:colortypicality}: we collected production data within the same reference game setting, but varied the contextual informativeness of utterances by varying whether distractors shared the same basic or superordinate category with the target (see \figref{fig:dogcontexts}). We also elicited typicality ratings for object-utterance combinations, which entered the model as the semantic values via the lexicon. We then conducted Bayesian data analysis, as in previous sections, for model comparison.
Our key insight is that compared to a traditional Boolean semantics where class labels (e.g. \emph{dog}) are strictly true or false of objects, a continuous semantics incorporating knowledge of typicality more successfully predicts preferences for taxonomic level of reference.
That is, cs-RSA accurately predicts, from Gricean principles, that speakers will increase their preference for subordinate terms (e.g. \emph{penguin}) when an object is a more atypical example of the basic-level (\emph{bird}) and prefer basic-level terms (e.g. \emph{bird}) when those objects are more atypical examples of the super-ordinate level (\emph{animal}).

\subsection{Method}

\paragraph{Participants}

We recruited 58 pairs of participants (116 participants total, the same participants as in Exp.~1) over Amazon's Mechanical Turk who were each paid \$1.75 for their participation. 

\paragraph{Procedure and materials}

The procedure was identical to that of Exp.~1.\footnote{A separate earlier data set was reported at the annual meeting of the Cognitive Science Society \cite{GrafEtAl2016}, and serves as a close replication of the reported study.} Participants proceeded through 72 trials. Of these, half were critical trials of interest and half were filler trials (the critical trials from Exp.~1). On critical trials, we varied the level of reference that was sufficient to mention for uniquely establishing reference.

Stimuli were selected from nine distinct domains, each corresponding to distinct basic level categories such as \emph{dog}.  For each domain, we selected four subcategories to form our target set (e.g. \emph{dalmatian}, \emph{pug}, \emph{German Shepherd} and \emph{husky}). See \tableref{tab:reflevelstimuli} in \appref{app:taxonomicstimuli} for a full list of domains and their associated target items. Each domain also contained an additional item which belonged to the same basic level category as the target (e.g., \emph{greyhound}) and items which belonged to the same supercategory but not the same basic level (e.g., \emph{elephant} or \emph{squirrel}). The latter items were used as distractors.

\begin{figure}
	\begin{subfigure}{.5\textwidth}
		\centering
		\includegraphics[width=.8\textwidth]{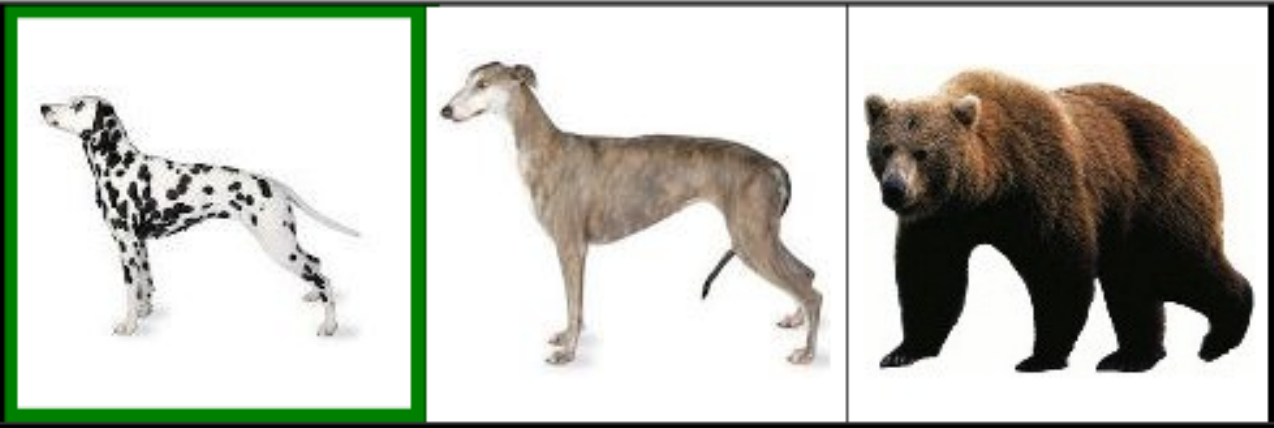}
		\caption{subordinate necessary}
		\label{fig:item12}
	\end{subfigure}
	\begin{subfigure}{.5\textwidth}
		\centering
		\includegraphics[width=.8\textwidth]{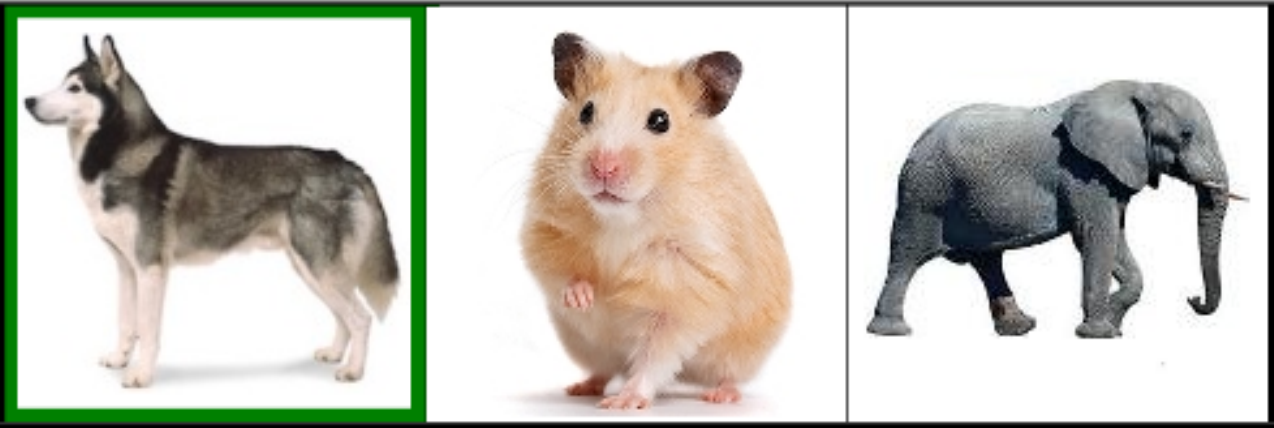}
		\centering
		\caption{basic sufficient (two superordinate distractors)}
		\label{fig:item22}
	\end{subfigure}
	\begin{subfigure}{.5\textwidth}
		\centering
		\includegraphics[width=.8\textwidth]{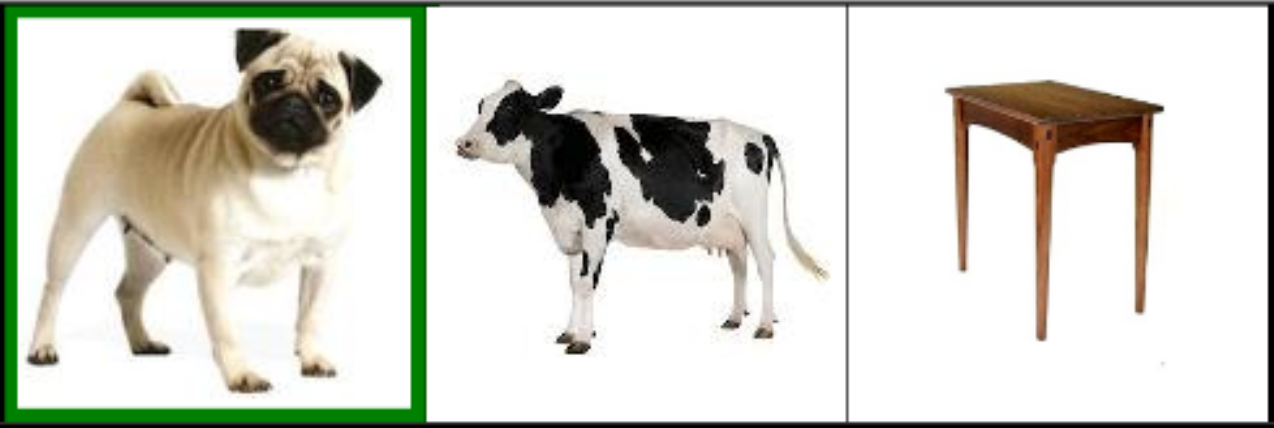}
		\caption{basic sufficient (one superordinate distractor)}
		\label{fig:item23}
	\end{subfigure}
	\begin{subfigure}{.5\textwidth}
		\centering
		\includegraphics[width=.8\textwidth]{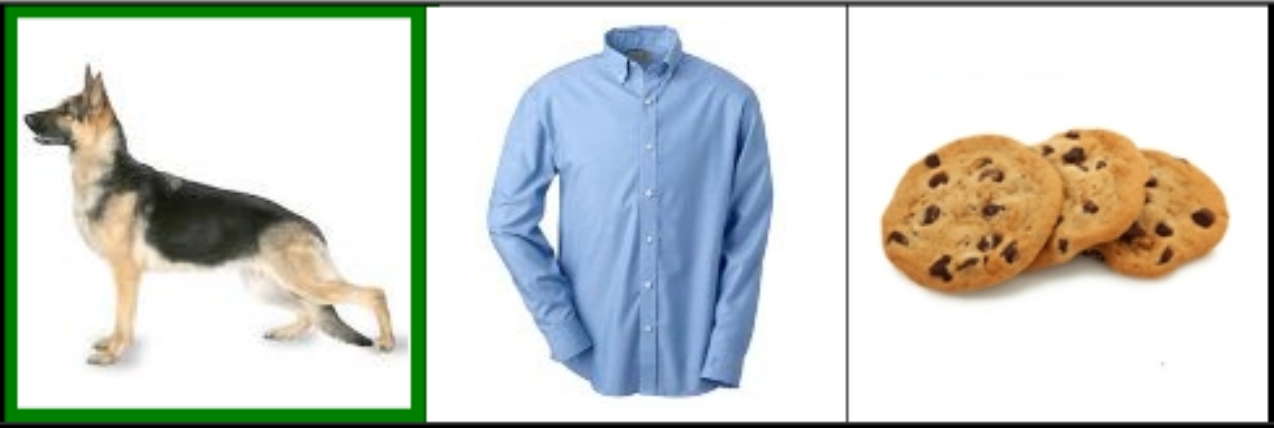}
		\centering
		\caption{superordinate sufficient}
		\label{fig:item33}
	\end{subfigure}
	\caption{Example contexts in which different levels of reference are necessary for establishing unique reference to the target marked with a thick border: (a) subordinate necessary (\emph{dalmatian}); (b, c) basic sufficient (\emph{dog}) and subordinate possible (\emph{husky}, \emph{pug}); (d) superordinate sufficient (\emph{animal}) and basic or subordinate possible (\emph{dog}, \emph{German Shepherd}).}
	\label{fig:dogcontexts}
\end{figure}

Each trial consisted of a display of three images, one of which was designated as the target object. Each pair of participants saw each target exactly once, for a total of 36 trials. These target items were randomly assigned distractor items which were selected from three different context conditions, corresponding to different communicative pressures (see \figref{fig:dogcontexts}). The \emph{subordinate necessary} contexts contained one distractor of the same basic category and one distractor of the same superordinate category (e.g., target: \emph{dalmatian}, distractors: \emph{greyhound} (also a dog) and \emph{squirrel} (also an animal)). The \emph{basic sufficient} contexts contained either two distractors of the same superordinate category but different basic category as the target (e.g., target: \emph{husky}, distractors: \emph{hamster} and \emph{elephant}) or one distractor of the same superordinate category and one unrelated item (e.g., target: \emph{pug}, distractors: \emph{cow} and \emph{table}). The \emph{superordinate sufficient} contexts contained two unrelated items (e.g., target: \emph{German Shepherd}, distractors: \emph{shirt} and \emph{cookie}). 

This context manipulation served as a manipulation of utterance informativeness: any target could be referred to at the subordinate (\emph{dalmatian}), basic (\emph{dog}) or superordinate (\emph{animal}) level. However, the level of reference necessary for uniquely referring differed across contexts.

\paragraph{Data pre-processing and exclusion}

We collected data from 2193 critical trials. Each referring expression was classified as containing the  target's correct \emph{sub}(ordinate, e.g., \emph{dalmatian}), \emph{basic} (e.g., \emph{dog}), or \emph{super}(ordinate, e.g., \emph{animal}) level term, or excluded if classification was not possible. See \appref{sec:preprocessing} for details on exclusion criteria and the pre-processing procedure.  After exclusions and pre-processing, 1872 cases entered the analysis.

\paragraph{Typicality norming}
\label{sec:typicalitynormingnominal}

In order to test for typicality effects on the production data and to evaluate cs-RSA's performance, we again collected empirical typicality values for each utterance/object pair. See \appref{app:typicalitynorms2} for details.

\subsection{Behavioral results and discussion}

\begin{figure}
\centering
\includegraphics[width=\textwidth]{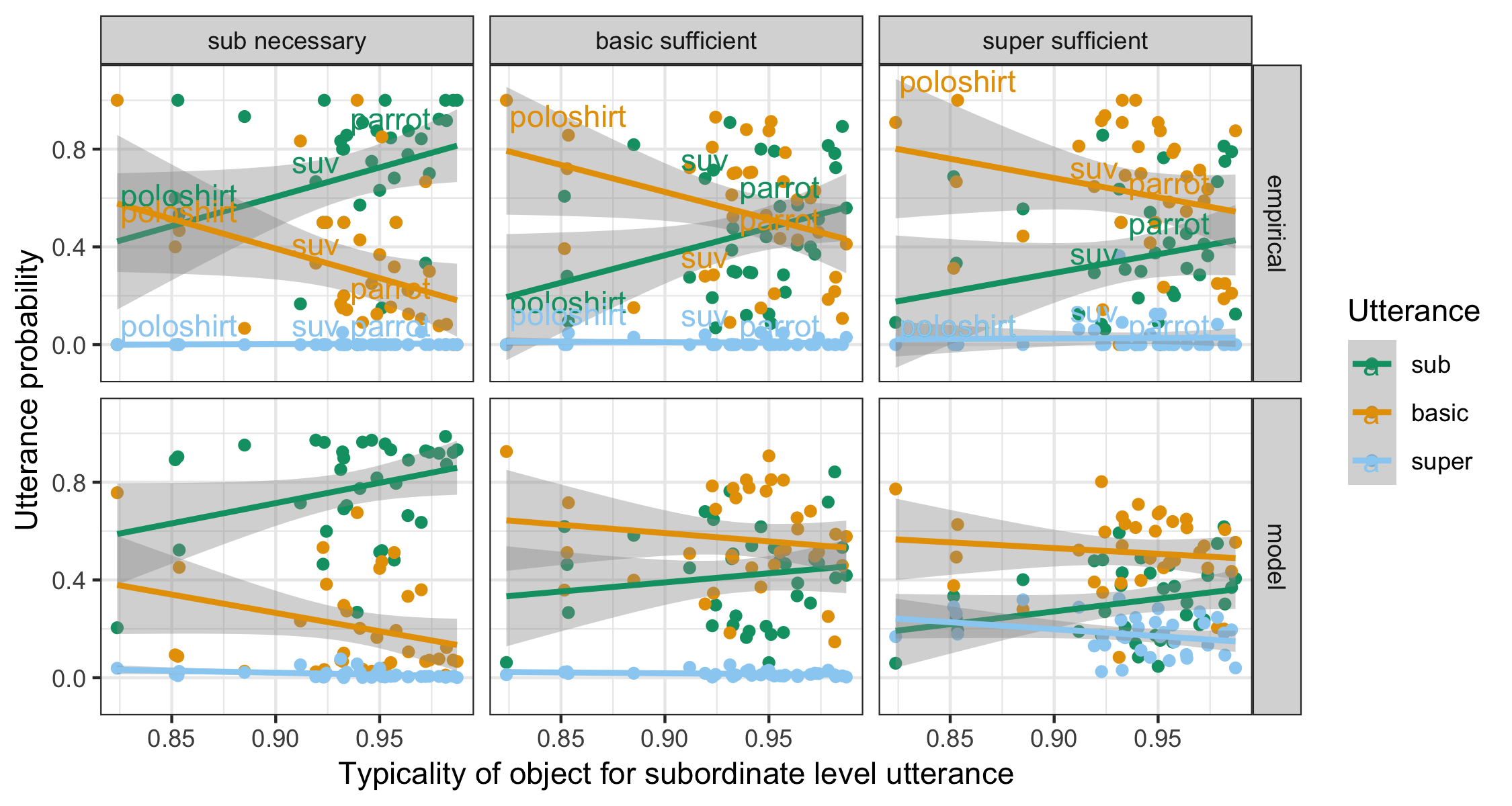}
\caption{Top: utterance proportions for each target item across different informativeness conditions as a function of the object's subordinate level typicality. Example target items \emph{polo shirt} (basic: \emph{shirt}, super: \emph{clothes}),  \emph{SUV} (basic: \emph{car}, super: \emph{vehicle}), and \emph{parrot} (basic: \emph{bird}, super: \emph{animal}) that were characteristic of relatively low to relatively high sub typicality items are labeled explicitly. Bottom: MAP model predicted utterance probabilities.}
\label{fig:exp3results}
\end{figure}

Proportions of sub, basic, and super level utterances are shown in \figref{fig:exp3results}. Overall, super level mentions are highly dispreferred ($< 2\%$), so we focus in this section only on predictors of sub over basic level mentions. The clearest pattern of note is that sub level mentions are only preferred in the most constrained context that necessitates the sub level mention for unique reference (e.g., target: dalmatian, distractor: greyhound; see \figref{fig:item12}). Nevertheless, even in these contexts there is a non-negligible proportion of basic level mentions (28\%). This includes cases of using just the basic level term (6\%, e.g., \emph{dog} for the German Shepherd when one of the distractors was a greyhound, an atypical dog, akin to the unmodified cases in the informative conditions discussed in \sectionref{sec:exp2resultsdisc}) as well as basic level terms with additional modifying material (22\%). In the remaining contexts, where the sub and basic level are equally informative, there is a clear preference for the basic level. Finally, mitigating this context effect, sub level mentions increased with increasing typicality of the object as an instance of the sub level utterance.

What explains these preferences? In order to test for effects of informativeness, length, frequency, and typicality on nominal choice we conducted a mixed effects logistic regression predicting sub over basic level mention from centered predictors for the factors of interest. Because the maximal model with by-speaker and by-item slopes for all fixed effects did not converge, we simplified the random effects structure, including only by-speaker and by-item random intercepts.

\emph{Frequency} was coded as the difference between the sub and the basic level's log frequency, as extracted from the Google Books Ngram English corpus ranging from 1960 to 2008. Speakers prefer more frequent words over less frequent ones \cite{oldfield1965response}.

\emph{Length} was coded as the ratio of the subordinate to the basic level term's length in characters. Even among one-word utterances, speakers prefer shorter ones over longer ones \cite{degenfrankejaeger2013, rohde2012}. We used the mean number of characters in the utterances participants produced. For example, the minivan, when referred to at the subcategory level, was sometimes called ``minivan'' and sometimes ``van'' leading to a mean empirical length of 5.71. This is the value that was used, rather than 7, the length of ``minivan''. 

\emph{Typicality} was coded as the ratio of the target's sub to basic level label typicality.\footnote{Typicalities were elicited in a separate norming study that was identical in procedure to that of Exp.~1a. See \appref{app:typicalitynorms2} for details about the study.} That is, the higher the ratio, the more typical the object was for the sub level label compared to the basic level; or in other words, a higher ratio indicates that the object was relatively atypical for the basic label compared to the sub label. For instance, the panda was relatively atypical for its basic level ``bear'' (mean rating 0.75) compared to the sub level term ``panda bear'' (mean rating 0.98), which resulted in a relatively \emph{high} typicality ratio. We predicted that subordinate terms may be preferred when an object is a particularly good instance of that term or a particularly bad instance of the basic level term, compared to the other objects in the context. 

\emph{Informativeness} condition was coded as a three-level factor: \emph{sub necessary}, \emph{basic sufficient}, and \emph{super sufficient}, where \emph{basic sufficient (two superordinate distractors)} and \emph{basic sufficient (one superordinate distractor)} were collapsed into \emph{basic sufficient}. Condition was Helmert-coded: two contrasts over the three condition levels were included in the model, comparing each level against the mean of the remaining levels (in order: \emph{sub necessary}, \emph{basic sufficient}, \emph{super sufficient}). This allowed us to determine whether the probabilities of type mention  for neighboring conditions were significantly different from each other, as suggested by \figref{fig:exp3results}.

The log odds of mentioning the sub level term were greater in the \emph{sub necessary} condition than in either of the other two conditions ($\beta = 2.11$, $SE = .17$, $p < .0001$), and greater in the \emph{basic sufficient} condition than in the \emph{super sufficient} condition ($\beta = .60$, $SE = .15$, $p < .0001$), suggesting that the contextual informativeness of the sub level mention has a gradient effect on utterance choice.\footnote{Importantly, model comparison between the reported model and one that subsumes basic and super under the same factor level revealed that the three-level condition variable is justified ($\chi ^2 (1) = 12.82$, $p < .0004$), suggesting that participants do not simply revert to the basic level when no basic-level distractor is in context.} There was also a main effect of typicality, such that the sub level term was preferred for objects that were more typical for the sub level compared to the basic level  description ($\beta = 4.82$, $SE = 1.35$, $p < .001$). In addition, there was a main effect of length, such that as the length of the sub level term increased compared to the basic level term (``chihuahua''/``dog'' vs.~``pug''/``dog''), the sub level term was dispreferred (``chihuahua'' is dispreferred compared to ``pug'', $\beta = -.95$, $SE = .27$, $p < .001$). The main effect of frequency did not reach significance ($\beta = .08$, $SE = .11$, $p < .45$).

Unsurprisingly, there was also significant by-participant and by-domain variation in sub level term mention. 
For instance, mentioning the sub over the basic level term was preferred more in some domains (e.g. in the ``candy'' domain) than in others. Likewise, some domains had a greater preference for basic level terms (e.g. the ``shirt'' domain). Using the super term also ranged from hardly being observable (e.g., \emph{plant} in the ``flower'' domain) to being used more frequently (e.g., \emph{furniture} in the ``table'' domain and \emph{vehicle} in the ``car'' domain). 
We thus replicated the well-documented preference to refer to objects at the basic level, which is partly modulated by contextual informativeness and partly a result of the basic level term's cognitive cost and typicality compared to its sub level competitor, mirroring the results from Exp.~2. 

Perhaps surprisingly, we did not observe an effect of frequency on sub level term mention. This is likely due to the modality of the experiment: the current study was a written production study, while most studies that have identified frequency as a factor governing production choices are spoken production studies. It may be that the cognitive cost of typing longer words may be disproportionately higher than that of producing longer words in speech, thus obscuring a potential effect of frequency. Support for this hypothesis comes from studies comparing written and spoken language, which has found that spoken descriptions are likely to be longer than written descriptions and, in English, seem to have a lower propositional information density than written descriptions \cite{VanMiltenburg2018}.\footnote{In order to address convergence issues with \verb+lmer+ when specifying the full random effects structure -- i.e., by-speaker and by-item random intercepts and slopes for all fixed effects -- we also ran a Bayesian binomial mixed effects model with weakly informative priors using the \verb+brms+ package \cite{brms} that included the same fixed effects structure as the lmer model and the full random effects structure. The results were qualitatively identical, yielding  evidence for main effects of context (sub vs basic sufficient: posterior mean $\beta$ = 2.44, $95\%$ CI = $[$1.87,3.06$]$, $p(\beta > 0)$ = 1; basic vs super sufficient: posterior mean $\beta$ = 0.70, $95\%$ CI = $[$0.32,1.09$]$, $p(\beta > 0)$ = 1), typicality (posterior mean $\beta$ = 9.96, $95\%$ CI = $[$3.55,17.51$]$, $p(\beta > 0)$ = 1), and length (posterior mean $\beta$ = -1.12, $95\%$ CI = $[$-2.00,-0.31$]$, $p(\beta < 0)$ = 1).}

\subsection{Model evaluation}
\label{sec:reflevelmodel}

We evaluated cs-RSA on the production data from Exp.~3. The architecture of the model is identical to that of the model presented in \sectionref{sec:colorypicalitymodel}. The only difference is the set of alternative utterances.\footnote{See \tableref{tab:modeldiffs} for an overview of the models reported in the paper.} Whereas the models from the previous sections treated all individual features and feature combinations present in the display as utterance alternatives, for computational efficiency we now consider only the three different levels of reference to the target as alternatives, i.e. subordinate (\emph{dalmatian}), basic (\emph{dog}), and superordinate (\emph{animal}). So, even when a German Shepherd is present as a distractor, \emph{German Shepherd} is \emph{not} considered an alternative utterance for the dalmatian target. This has minimal effects on model predictions as long as \emph{German Shepherd} has low semantic fit to the dalmatian target. 

In Experiment 2, we tested which of three different semantics was most justified -- the empirically elicited typicality semantics, a fixed semantics with type-level semantic values, and one that combined both. Here, the relevant comparison is between a fixed Boolean noun semantics (i.e. 1 if the object belongs to the given class label, and 0 otherwise) and the empirically elicited typicality semantics.
Again, we introduced a parameter that interpolated between these semantic values. 
Additionally, we evaluated which cost function was best supported by the data: the one defined in \eqref{eq:exp2cost} (a linear weighted combination of an utterance's length and its frequency) or a simpler baseline in which utterances were assumed to have no cost.

\begin{figure}
\centering
\includegraphics[width=\textwidth]{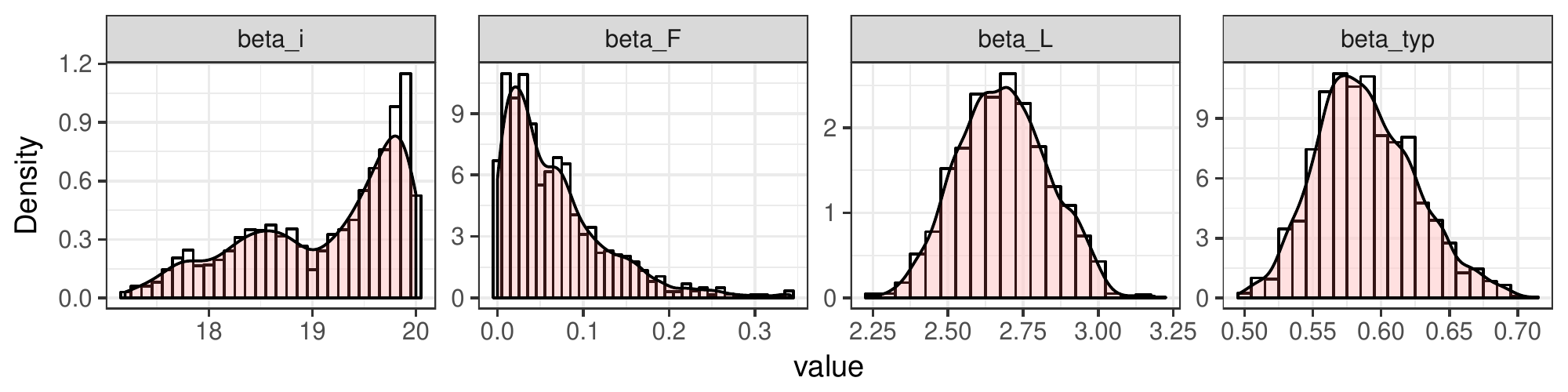}
\caption{Posterior model parameter distributions for interpolation weight on fixed vs.~empirical semantics $\beta_{\textrm{fixed}}$, informativity weight $\beta_i$, typicality weight $\beta_t $, frequency cost weight $\beta_{F}$, and length cost weight $\beta_{L}$. Maximum a posteriori (MAP) $\beta_{\textrm{fixed}}$ = 0.004, 95\% highest density interval (HDI) = [0.000,0.03];  $\beta_i$ = 19.8, HDI = [17.71,20.0]; MAP $\beta_t $ = 0.57, HDI = [0.53,0.67]; MAP $\beta_{F}$ = 0.0.02, HDI = [0.00,0.19]; MAP $\beta_{L}$ = 2.69, HDI = [2.42,2.99].}
\label{fig:nomparamposteriors}
\end{figure}

We employed the same procedure as in the previous section to compute the Bayes Factor for the comparison between the two cost models, and to compute the posteriors over parameters. Priors were again  $\beta_i  \sim \mathcal{U}(0,20)$,  $\beta_{F} \sim \mathcal{U}(0,5)$, $\beta_{L} \sim \mathcal{U}(0,5)$, $\beta_t  \sim \mathcal{U}(0,5)$.
Despite the greater number of parameters associated with adding the cost function, the  model that includes non-zero costs was preferred compared to its no-cost counterpart ($BF = 2.8 \times 10^{77}$). 
Posteriors over parameters are shown in \figref{fig:nomparamposteriors}. 
First, we observe that the semantic interpolation value was highly skewed toward 0, strongly indicating that empirical typicality values strongly improve model performance over a Boolean baseline.
Second, the weight on frequency is close to zero. That is, in line with the results from the mixed effects regression, it is an utterance's length, but not its frequency, that affects the probability with which it is produced in this paradigm.\footnote{As discussed in previous sections, the lack of importance of a word's frequency may well be attributable to the written modality within which participants generated referring expressions.} 

MAP model  predictions are shown alongside empirical utterance proportions in \figref{fig:exp3results}. 
The correlation between empirical utterance proportions and the model's MAP predictions at the level of targets, utterances, and conditions was $r = .86$. Further collapsing across targets yields a correlation of $r = .95$.
The model captures the qualitative patterns well, though it somewhat overpredicts subordinate level  and underpredicts basic level choices. It also accounts for the strong preference against super level mentions. The reason for this is that the semantics for each utterance (eg., \emph{dalmatian}, \emph{dog}, \emph{animal}) is taken from the empirically elicited typicality values for each utterance-object pair. As can be seen in the left panel of \figref{fig:typicalityboxplots},  the target images used in this experiment were generally rated as less typical instances of the superordinate level term than of the basic or subordinate level term. This difference is enough to lead to a general bias against using the superordinate level term, especially when coupled with the fact that superordinate terms tend to be costlier than basic level terms.

\section{General Discussion}
\label{sec:gd}

In this paper we have provided a unified account of referring expression choice that solves a long-recognized puzzle for rational theories of language use: why do speakers'  referring expressions often and systematically exhibit seeming overinformativeness? We have shown here that by allowing contextual utterance informativeness to be computed with respect to a continuous (or noisy) rather than a Boolean semantics, utterances that seem overinformative can in fact be sufficiently informative. This happens when what seems like the prima facie sufficiently informative utterance is in fact noisy and may lead a literal listener astray; adding redundancy ensures successful communication. This simple modification to the Rational Speech Act approach allowed us to capture: the basic well-documented asymmetry for speakers to be more likely to redundantly use color adjectives than size adjectives; the interaction between sufficient dimension and scene variation in the probability of redundancy; and typicality effects in both color modifier choice and noun choice. 

We have thus shown that with one key innovation -- a continuous semantics  -- one can retain the assumption that speakers rationally trade off informativeness and cost of utterances in language production. Rather than being wastefully overinformative, adding redundant modifiers or referring at a lower taxonomic level than strictly necessary \emph{is} in fact appropriately informative.
This innovation thus not only provides a unified explanation for a number of key patterns within the overinformative referring expression literature that have thus far eluded a unified explanation; it also extends to the domain of nominal choice. And in contrast to previously proposed computational models, it is straightforwardly extendable to any instance of definite referring expressions of the sort we have examined here. 

\subsection{Comparison of model components across experiments}

In order to address the possible concern that the different models employed are too different from one another to be comparable, we begin by providing an overview of the parts of the model that remained the same or differed across experiments. While the core architecture with relaxed semantics remained constant throughout the paper, some peripheral components were adjusted to accommodate the aims of the different experiments. 
These different choices are fully consistent with one another, and many of them were justified against alternatives via model comparison.
We have provided an overview of the best-fitting RSA models for each of the three reported production datasets in \tableref{tab:modeldiffs}. 

Most prominently, Exps.~2 and 3 aimed to predict patterns of reference via typicality at the \emph{object-level}; in those cases the model thus required semantic values for each utterance-object pair in the lexicon.
While these values could have in principle been inferred from the data, as we inferred the two type-level values in Exp.~1, it would have introduced a large number of additional parameters (see \emph{size of lexicon}). 
Instead, we addressed this problem by empirically eliciting these values in an independent task and introducing a single free concentration parameter $\beta_t$ that modulated their strength.
In the case of Exp. 2, we found that the best-fitting model smoothly integrated these empirical values with type-level values used in Exp.~1.

The need to make object-level predictions also drove decisions about what to use as the cost function and the set of alternative utterances.
For instance, in Exp.~3 we could have inferred the cost of each noun but this again would have introduced a large number of free parameters and risked overfitting. 
Instead we used the empirically estimated \emph{length} and \emph{frequency} of each word. 
For Exp.~2, we tested models both using fixed costs for each modifier as in Exp.~1 and empirical length and frequency costs as in Exp.~3, but our model comparison showed that neither sufficiently improved the model's predictions.

\begin{table}
\caption{Overview of the best-performing models used for the three different production datasets Color/size (Exp.~1), Color typicality (Exp.~2), and Nominal choice (Exp.~3). Parameter names:  $x_{\textrm{color}}$: semantic value of color; $x_{\textrm{size}}$: semantic value of size; $\beta_{c(\textrm{color})}$: cost of color;  $\beta_{c(\textrm{size})}$: cost of size; $\beta_i$: weight on informativity; $\beta_F$: weight on cost (as estimated by utterance frequency); $\beta_L$: weight on cost (as estimated by utterance length); $\beta_t$: weight on elicited typicality values; $\beta_{\textrm{fixed}}$: interpolation weight between fixed type-level and empirical semantic values} 
\begin{tabular}{p{2.5cm} p{4.2cm} p{4.2cm} p{4.2cm} }
\toprule
& Color/size & Color typicality & Nominal choice \\
\midrule
\makecell{Semantic \\ values} & at type-level (inferred) & \makecell{at type-level (inferred) \\ + object-level (elicited)} & at object-level (elicited) \\
\midrule
\makecell{Size of \\lexicon $\mathcal{L}(u,o)$}  & 8 (all combinations of \emph{size} and \emph{color}) & 814 (1 for each utterance-object pair) & 51 (1 for each utterance-object pair)\\
\midrule
\makecell{Set of \\alternatives} & 8 contextually available feature combinations (size, color) & 8 or 9 contextually available feature combinations (type, color) & 3 target alternatives (level of reference: \emph{sub, basic, super}) \\
\midrule
Cost & \makecell{type-level \\ (color and size)}& none necessary & \makecell{empirical \\ (length and frequency)} \\
\midrule
\makecell{Free\\ parameters} & 
	\makecell{$x_{\textrm{color}}$, $x_{\textrm{size}}$, \\ $\beta_{c(\textrm{color})}$,  $\beta_{c(\textrm{size})}$, $\beta_i$} & 
	\makecell{$x_{\textrm{color}}$, $x_{\textrm{type}}$, \\ $\beta_i$, $\beta_t$, $\beta_{\textrm{fixed}}$} & 
	$\beta_F$, $\beta_L$, $\beta_i$, $\beta_t$ \\
\bottomrule
\end{tabular}
\label{tab:modeldiffs}
\end{table}

Finally, the set of alternative utterances differed slightly across the three experiments for computational reasons.
Because Exp.~1 collapsed over the particular levels of size and color, it was practical to consider all utterances in the lexicon for every target. 
In Exp.~2 and Exp.~3, however, the space of possible utterances was large enough that this exhaustive approach became impractical. 
We noticed that the probability of using some utterances (e.g. `table' to refer to a Dalmatian)  was low enough that we could prune the utterance space to only those that could plausibly apply to the objects in context without substantially altering the model's behavior.
Future work must address how predictions may change as more complex referring expressions outside the scope of this paper enter the set of alternatives (e.g. the option of combining adjectives with nominal expressions, as in \emph{the cute, spotted dog}).
In the following we discuss a number of intriguing questions that this work raises and avenues for future research it suggests.

\subsection{Comparison with PRO}

While a detailed comparison of cs-RSA with PRO \cite{VanGompel2019}, the hitherto most state-of-the-art computational model of human production of modified referring expressions, is outside the scope of this paper, we include some comparative remarks here. PRO has the advantage of being computationally more efficient than cs-RSA, partly because it aims to be an algorithmic-level model, which may be of importance for applications. PRO may further have the advantage of having fewer parameters, though this is a bit harder to evaluate in general: while PRO as applied to the choice of color and size in principle involves 2 parameters, $s$ (size preference) and $e$ (overspecification eagerness), in the 2019 paper the maximum likelihood parameter values are estimated on each of the experimental conditions separately, effectively resulting in 6 parameters. In the extension to 3 properties, this results in 14 parameters, and this number increases further as more properties and conditions are added. If the parameter values had been estimated on all conditions jointly, which is what we did in the evaluation of cs-RSA instead of separately, then indeed PRO would have fewer effective parameters than cs-RSA, though it is unclear how this would affect the data fit. 

One advantage of the cs-RSA approach is greater generality. For instance, it is not immediately clear how PRO should be extended to contexts that vary in the number and nature of distractors, where empirical overmodification proportions  change, but the PRO predictions would not. We see this as one of the great strengths of cs-RSA: scene variation effects fall out of the model directly. Finally, we have proposed here a way to account for typicality effects. PRO may be able to accommodate typicality effects if its preference parameters can be made to be sensitive to typicality. In general, a systematic comparison and possible combination of these models is an important next step.

\subsection{`Overinformativeness'}

This work challenges the traditional notion of overinformativeness as it is commonly employed in the linguistic and psychological literature. The reason that redundant referring expressions are interesting for psycholinguists to study is that they seem to constitute a clear violation of rational theories of language production. For example, Grice's Quantity-2 maxim, which asks of speakers to ``not make [their] contribution more informative than is required'' \cite{grice1975}, appears violated by any redundant referring expression -- if one feature uniquely distinguishes the target object from the rest and a second one does not, mentioning the second does not contribute any information that is not already communicated by the first. Hence, the second is considered `overinformative', a referring expression that contains it  `overspecified.'

This conception of (over-)informativeness assumes that all modifiers are born equal -- i.e., that there are no a priori differences in the utility of mentioning different properties of an object. Under this conception of modifiers, there are hard lines between modifiers that are and aren't informative in a context. However, what we have shown here is that under a continuous semantics, a modifier that would be regarded as overinformative under the traditional conception may in fact communicate  information about the referent. The more visual variation there is in the scene, and the less noisy the redundant modifier is compared to the modifier that selects the  dimension that uniquely singles out the target, the more information the redundant modifier adds about the referent, and the more likely it therefore is to be mentioned. This work thus challenges the traditional notion of utterance overinformativeness by providing an alternative that captures the quantitative variation observed in speakers' production in a principled way while still assuming that speakers are aiming to be informative, and is compatible with other efficiency-based accounts of `overinformative' referring expressions \cite<e.g.,>{sedivy2003a,rubiofernandez2016}.

But this raises a question: what counts as a truly overinformative utterance under RSA with a continuous semantics? Cs-RSA shifts the standard for overinformativeness and turns it into a graded notion: the less expected the use of a redundant modifier is contextually,  the more the use of that modifier should be considered overinformative. For example, consider again \figref{fig:exp1results}: the less scene variation there is, the more truly overinformative the use of the redundant modifier is. Referring to \emph{the big purple stapler} when there are only purple staplers in the scene should be considered overinformative. If there is one red stapler, the utterance should be judged less overinformative, and the more non-purple staplers there are, the less overinformative the utterance should be judged. We leave a systematic test of this prediction for our stimuli for future research, though we point to some qualitative examples where it has been borne out previously in the next subsection.

\subsection{Comprehension}

While the account proposed in this paper is an account of the \emph{production} of referring expressions, it can be extended straightforwardly to \emph{comprehension}. RSA models typically assume that listeners  interpret utterances by reasoning about their model of the speaker. In this paper we have provided precisely such a model of the speaker. In what way should the predicted speaker probabilities enter into comprehension? There are two interpretations of this question: first, what is the ultimate interpretation that listeners who reason about speakers characterized by the model provided in this paper arrive at, i.e.~what are the predictions for referent choice? And second, how do the production probabilities enter into online processing of prima facie overinformative utterances? The first question has a clear answer. For the second question we offer a more speculative answer.

\subsubsection{Choice of referent}

Most RSA reference models, unlike the one reported in this paper, have focused on comprehension \cite{frank2012, degenfrankejaeger2013, QingFranke2015, FrankeDegen2016}. The formula that characterizes pragmatic listeners' referent choices is:

\begin{equation}
P_{L_1}(o | u) \propto P_{S_1}(u | o) \cdot P(o)
\end{equation}

That is, the pragmatic listener interprets utterance $u$ (e.g., \emph{the big purple stapler}) via Bayesian inference, taking into account both the speaker probability of producing \emph{the big purple stapler} and its alternatives, given a particular object $o$ the speaker had in mind, as well as the listener's prior beliefs about which object the speaker is likely to intend to refer to in the context. For the situations considered in this paper, in which the utterance is  semantically compatible with only one of the referents in the context, this always predicts that the listener should choose the target. And indeed, in Exps.~1-3 the error rate on the listeners' end was always below 1\%. From a referent choice point of view, then, these contexts are not very interesting. They are much more interesting from an online processing point of view, which we discuss next.

\subsubsection{Online processing}

The question that has typically been asked about the online processing of redundant utterances is this: do redundant utterances, compared to their minimally specified alternatives, help or hinder comprehenders in choosing the intended referent? `Help' and `hinder' are typically translated into `speed up' and `slow down', respectively. What does the RSA model presented here have to say about this? 

In sentence processing, the current wisdom is that the processing effort spent on linguistic material is related to how surprising it is \cite{Hale2001,Levy2008}. In particular, an utterance's log reading time is linear in its surprisal \cite{Smith2013}, where surprisal is defined as $-\log p(u)$. In these studies, surprisal is usually estimated from linguistic corpora. Consequently, an utterance of \emph{the big purple stapler} receives a particular probability estimate independent of the non-linguistic context it occurred in. Here we provide a speaker model from which we can derive estimates of \emph{pragmatic surprisal} directly for a particular context. We can thus speculate on a linking hypothesis: the more expected a redundant utterance is under the pragmatic continuous semantics speaker model, the faster it should be to process compared to its minimally specified alternative, all else being equal. We have shown that redundant expressions are more likely than minimal expressions when the sufficient dimension is relatively noisy and scene variation is relatively high. Under our speculative linking hypothesis, the redundant expression should be easier to process in these sorts of contexts  than in contexts where the redundant expression is relatively less likely. 

Is there evidence that listeners do behave in accordance with this prediction? Indeed, the literature reports evidence that in situations where the redundant modifier does provide some information about the referent, listeners are faster to respond and select the intended referent when they observe a redundant referring expression than when they observe a minimal one \cite{Arts2011,  Paraboni2007}. However, there is also evidence that redundancy sometimes incurs a processing cost: both \citeA{Engelhardt2011} and \citeA{Davies2013} (Exp.~2) found that listeners were slower to identify the target referent in response to redundant compared to minimal utterances. It is useful to examine the stimuli they used. In the Engelhardt et al study, there was only one distractor that varied in type, i.e., type was sufficient for establishing reference. This distractor varied either in size or in color. Thus, scene variation was very low and redundant expressions therefore likely surprising. Interestingly, the incurred cost was greater for redundant size than for redundant color modifiers, in line with the RSA predictions that color should be generally more likely to be used redundantly than size. In the Davies et al.~study, the `overinformative' conditions contained displays of four objects which differed in type. Stimuli were selected via a production pre-test: only those objects that in isolation were not referred to with a modifier were selected for the study. That is, stimuli were selected precisely on the basis that redundant modifier use would be unlikely.

While the online processing of redundant referring expressions is yet to be systematically explored under the cs-RSA account, this cursory overview of the patterns reported in the existing literature suggests that pragmatic surprisal may be a plausible linking function from model predictions to processing times. Excitingly, it has the potential for unifying the equivocal processing time evidence by providing a model of utterance probabilities that can be computed from the features of the objects in the context.

\subsection{Continuous semantics}
\label{sec:contsemantics}

The crucial component of the model that allows for capturing `overinformativeness' effects is the continuous semantics.  In this section, we consider the nature of these continuous semantic values. Readers already convinced of the utility of a continuous semantics are invited to skip to the next section.

For the purpose of Exp.~1 (modifier choice), a semantic value was assigned to modifier \emph{type}. The semantics of modifiers was underlyingly truth-conditional and the semantic value captured the probability that a modifier's truth conditions would accidentally be inverted. This model included only two semantic values, one for size and one for color, which we inferred from the data. For the datasets from Exps.~2 and 3, we then extended the continuous semantics to apply at the level of utterance-object  combinations (e.g., \emph{banana} vs.~\emph{blue banana} as applied to the blue banana item, \emph{dalmatian} vs.~\emph{dog} as applied to the dalmatian item) to account for typicality effects in modifier and nominal choice. In this instantiation of the model, the semantic value differed for every utterance-object combination and captured how good of an instance of an utterance an object was. These values were elicited experimentally to avoid over-fitting, and for the dataset from Exp.~2 we found further that a combination of a relaxed (noisy) truth-conditional semantics and the empirically elicited continuous semantics best accounted for the obtained production data. 

What we have said nothing about thus far is what determines these semantic values; in particular, which aspects of language users' experience -- perceptual, conceptual, communicative, linguistic -- they represent. We will offer some speculative remarks and directions for future research here. 

First, semantic values may  represent the difficulty associated with verifying whether the property denoted by the utterance holds of the object. This difficulty may be perceptual -- for example, it may be relatively easier to visually determine of an object whether it is red than whether it it is big (at least in our stimuli). Similarly, at the object-utterance level, it may be easier to determine of a yellow banana than of a blue banana whether it exhibits banana-hood,  consequently yielding a lower semantic value for a blue banana than for a yellow banana as an instance of \emph{banana}. Further, the value may be context-invariant or context-dependent. If it is context-invariant, the semantic value inferred for color vs.~size, for instance, should not vary by making size differences more salient and color differences less salient. If, instead, it is context-dependent, increasing the salience of size differences and decreasing the salience of color differences should result, e.g.,  in  color modifiers being more noisy, with concomitant effects on production, i.e., redundant color modifiers should become less likely. This is indeed what \citeA{Viethen2017} found. Similarly, \citeA{VanGompel2014} found that the asymmetry in redundant use of color vs.~with size disappeared when participants were shown displays with very noticeable size contrasts and barely noticeable color contrasts.

Another possibility is that semantic values represent aspects of agents' prior beliefs (world knowledge) about the correlations between features of objects. For example, conditioning on an object being a banana, experience dictates that the probability of it being yellow is much greater than of it being blue. This predicts the relative ordering of the typicality values we elicited empirically, i.e., the blue banana received a lower semantic value than the yellow banana as an instance of \emph{banana}.  

Another possibility is that the semantic values capture the past probability of communicative success in using a particular expression. For example, the semantic value of \emph{banana} as applied to a yellow banana may be high because in the past, referring to yellow bananas simply as \emph{banana} was on average successful. Conversely,  the semantic value of \emph{banana} as applied to a blue banana may be low because in the past, referring to blue bananas simply as \emph{banana} was on average unsuccessful (or the speaker may have uncertainty about its communicative success because they have never encountered blue bananas before). Similarly, the noise difference between color and size modifiers may be due to the inherent relativity of size modifiers compared to color modifiers -- while color modifiers vary somewhat in meaning across domains (consider, e.g., the difference in redness between \emph{red hair} and \emph{red wine}), the interpretation of size modifiers is highly dependent on a comparison class (consider, e.g., the difference between a \emph{big phone} and a \emph{big building}). In negotiating what counts as \emph{red}, then, speakers are likely to agree more often than in negotiating what counts as \emph{big}.  That is, size adjectives are more subjective than color adjectives. If semantic values encode adjective subjectivity, speakers should be even more likely to redundantly use adjectives that are more objective than color. In a study showing that adjective subjectivity is almost perfectly correlated with an adjective's average distance from the noun, \citeA{scontras2017} collected subjectivity ratings for many different adjectives and found that material adjectives like \emph{wooden} and \emph{plastic} are rated to be even more objective than color adjectives. Thus, under the hypothesis that semantic values represent adjective subjectivity, material adjectives should be even more likely to be used redundantly than color adjectives. This is not the case. For instance, \citeA{sedivy2003a} reports that material adjectives are used redundantly about as often as size adjectives. Hence, while the hypothesis that semantic values capture the past probability of communicative success in using a particular expression has yet to be systematically investigated, subjectivity alone seems not to be the determining factor.

Finally, it is also possible that semantic values are simply an irreducible part of the lexical entry of each utterance-object pair. This seems unlikely because it would require a separate semantic value for each utterance and object token, and most potentially encounterable object tokens in the world have not been encountered, making it impossible to store utterance-token-level values. However, it is possible that, reminiscent of prototype theory, semantic values are stored at the level of utterances and object \emph{types}. This view of semantic values  suggests that they should not be updated in response to further exposure of objects. For example, if semantic values were a fixed component of the lexical entry \emph{banana}, then even being exposed to a large number of blue bananas should not change the value. This seems unlikely but merits further investigation.

The various possibilities for the interpretation of the continuous semantic values included in the model are neither independent nor incompatible with each other. Disentangling these possibilities presents an exciting avenue for future research.

What is highlighted by the above discussion is that we have been using the term `semantics' at a fairly high level, to refer to conventional aspects of meaning that are relatively stable across contexts -- in RSA these are the representations on which the literal listener performs computations. These real valued representations could be primitives arising in lexical representations and threading through composition, which would constitute a fundamentally different basic semantics than has often been assumed. Alternatively, the necessary real values could arise by adding the right kind of use or world-knowledge related noise to a standard Boolean truth-conditional semantics. Most minimally, intensional parameters of a standard semantics could be set stochastically.
Whatever their source, these continuous values provide the right basis for capturing the production choices explored in this paper. When to assume a relaxed semantics, and what the implications of such a relaxation are for other semantic and pragmatic phenomena, are questions for future research.

\subsection{Audience design}

One question which has plagued the literature on language production is that of whether, and to what extent, speakers actually tailor their utterances to their audience \cite{Clark1982, horton1996, Brown-schmidt2014}. This is also known as the issue of \emph{audience design}. With regards to redundant referring expressions, the question is whether speakers produce redundant expressions because it is helpful to them (i.e., due to internal production pressures) or  because it is helpful to their interlocutor. For instance, \citeA{Walker1993} shows that redundancy is more likely when processing resources are limited. On the other hand, there is evidence that redundant  utterances are frequently used in response to signs of
listener non-comprehension, when responding to listener questions, or when speaking to strangers \cite{Baker2008}. 
Audience design has played an especially large role in explanations of typicality effects. For example, \citeA{Huettig2011} found that listeners, after hearing a noun with a diagnostic color (e.g., \emph{frog}), are more likely to fixate objects of that diagnostic color (green), indicating that typical object features like color are rapidly activated and aid visual search. Similarly, \citeA{Arts2011} showed that overspecified expressions result in faster referent identification.  Nevertheless, such benefits might simply be a happy coincidence and speakers might not, in fact, be deliberately designing their utterances for their addressees. 

RSA seems to make a claim about this issue: speakers are trying to be informative with respect to a literal listener. That is, it would seem that speakers produce referring expressions that are tailored to their listeners. However, this is misleading. The ontological status of the literal listener is as a ``dummy component'' that allows the pragmatic recursion to get off the ground. Actual listeners are, in line with previous work and briefly discussed above, more likely fall into the class of pragmatic $L_1$ listeners; listeners who reason about the speaker's intended meaning via Bayesian inference \cite{frank2012, goodmanstuhlmueller2013}.\footnote{But see \citeA{FrankeDegen2016} for an evaluation of the distribution of listener and speaker types in Quantity inferences.}
Because RSA is a computational-level theory \cite{marr1982} of language use, it does not claim that the mechanism of language production requires that speakers actively consult an  internal model of a listener every time they choose an utterance, just that the distribution of utterances they produce reflect informativity with respect to such a model. It is possible that this distribution is cached or computed using some other algorithm that doesn't explicitly involve a listener component. 
Thus, the RSA model as formulated here remains agnostic about whether speakers' (over-) informativeness should be considered geared towards listeners' needs or simply a production-internal process. Instead, the claim is that redundancy emerges as a property of the communicative situation as a whole.

\subsection{Other factors that affect redundancy}

RSA with a continuous semantics as presented in this paper straightforwardly accounts for effects of typicality, cost, and scene variation on redundancy in referring expressions. However, other factors have been identified as contributing to redundancy. For example, \citeA{rubiofernandez2016} showed that colors are mentioned more often redundantly for clothes than for geometrical shapes. Her explanation is that knowing an object's color is generally more useful for clothing than it is for shapes. It is plausible that agents' knowledge of \emph{goals} may be relevant here. For example, knowing the color of clothing is relevant to the goal of deciding what to wear or buy. In contrast, knowing the color of geometrical shapes is rarely relevant to any everyday goal agents might have. While the RSA model as implemented here does not accommodate an agent's goals, it can be extended to do so via projection functions, as has been done for capturing figurative language use \cite<e.g.,>{kao2014} or question-answer behavior \cite{Hawkins2015}. This should be explored further in future research.

One factor that has been repeatedly discussed in the literature and that we have not taken up here is the \emph{incrementality} of language production, both at the conceptual level of content or property selection and at the level of linguistic realization. For instance, according to \citeA{Pechmann1989}, incrementality is to blame for redundancy: speakers retrieve and subsequently produce words as soon as they can. Because color modifiers are easier to retrieve than size modifiers, speakers produce them regardless of whether or not they are redundant. The problem with this account is that it predicts that color adjectives should occur before size adjectives, thereby inverting the well-documented adjective ordering preferences for English \cite{bloomfield1933,sproat1991,scontras2017}. Pechmann does observe some, but not many, instances of this. An interesting test case for the incrementality hypothesis are cases where adjective ordering preferences are weak. \citeA{fukumura2018} reports one such case in which speakers prefer to order more discriminative and more available properties before less discriminative and less available ones, highlighting incrementality as an important factor affecting the choice of referring expression. On the other hand, it is unclear how incrementality -- whether in linguistic realization or in content selection -- could account for the systematic increase in color redundancy with increasing scene variation and decreasing color typicality, unless one makes the auxiliary assumption that the more contextually discriminative or salient color is, the more available (i.e., easily retrievable) the modifier is. Indeed, \citeA{Clark2004} emphasize the importance of \emph{salience against the common ground} in speakers' decisions about which of an object's properties to include in a referring expression. 

There are other ways incrementality could play a role in modifier choice. For example, mentioning the color adjective may buy the speaker time when the noun is hard to retrieve. This predicts that in languages with post-nominal adjectives, where this delay strategy cannot be used for noun planning, redundant color mention should be less frequent; indeed, this is what \citeA{rubiofernandez2016} shows for Spanish.\footnote{Rubio-Fern\'{a}ndez herself argues for an efficiency-based account of the Spanish/English asymmetry in color overmodification: ``color adjectives are a more efficient cue in pre-nominal position than in post-nominal position because in the latter case the hearer's visual search is initially guided by the noun'' \cite[p.~9]{rubiofernandez2016}. Cs-RSA as presented here does not predict positional asymmetries in modifier use. However, see \citeA{cohn2018} for an incremental version of RSA that captures the asymmetry reported by \citeA{rubiofernandez2016} as the result of incremental rather than global reasoning about utterance informativeness.} In sum, the incrementality of language production clearly affects the choice of referring expression; the ways in which considerations of incrementality should be incorporated in RSA are to be explored further \cite<see>[for a step in this direction]{cohn2018}.

\subsection{Extensions to other language production phenomena}

In this paper we focused on providing a computationally explicit account of definite modified and nominal referring expressions in reference games, focusing on the use of prenominal size and color adjectives as well as on the taxonomic level of noun reference. The cs-RSA model can be straightforwardly extended to different  nominal domains and different properties. For instance, the literature has also explored `overinformative' referring expressions that include material (\emph{wooden, plastic}), other dimensional (\emph{long, short}), and other physical (\emph{spotted, striped}) adjectives. 

However, beyond the relatively limited linguistic forms we have explored here, future research should also investigate the very intriguing potential for this approach to be extended to any language production phenomenon that involves a choice in which aspects of an event or entity to mention (content selection) and how to realize that content linguistically, including in the domain of reference (pronouns, names, definite descriptions with post-nominal modification) and event descriptions. For example, in investigations of optional instrument mentions, \citeA{brown1987} showed that atypical instruments are more likely to be mentioned than typical ones -- if a stabbing occurred with an icepick, speakers prefer \emph{The man was stabbed with an ice pick} rather than \emph{The man was stabbed}. If instead a stabbing occurred with a knife, \emph{The man was stabbed} is preferred over \emph{The man was stabbed with a knife}). This is very much parallel to the case of atypical color mention. 

Similarly, the approach outlined here might be extended to the case of non-restrictive modifiers. In these cases, the modifier is not used to distinguish a target referent from possible competitors. Instead, the speaker may intend to communicate an aspect of an already contextually established referent, as in \emph{Sit by the newly painted table}, where the speaker is warning the listener not to put their elbows on the table \cite{dale1995}; or \emph{Forrest looks at the massive crowd}, where the speaker is commenting on the extraordinary size of the crowd \cite{Hahn2018}. In these cases, the table and the crowd are contextually given referents; the modifiers \emph{newly painted} and \emph{massive} are used to highlight informative aspects of these referents. The approach proposed in this paper could be extended to these cases by allowing the speaker to be informative with respect to goals other than getting the listener to infer the intended referent. For instance, the speaker may want to be informative with respect to the goal of highlighting task-relevant properties of contextually given referents, as in the newly painted table case. 

More generally, the approach should extend to any  phenomenon that affords a choice between a more or less specific utterance. Whenever the more specific utterance adds relevant information compared to the less specific one, it should be produced. This is related to surprisal based theories of production like Uniform Information Density \cite<UID,>{jaeger2006, levy2007, frank2008, jaeger2010}, where  speakers have been found to be more likely to omit linguistic signal if the underlying meaning or syntactic structure is highly predictable to the listener. Importantly, UID diverges from our account in that it is an account of the choice between meaning-equivalent alternative utterances and includes no pragmatic reasoning component. 

\subsection{Conclusion}
\label{sec:conclusion}

In conclusion, we have provided an account of redundant referring expressions that challenges the traditional notion of `overinformativeness', unifies multiple language production literatures, and has the potential for many further extensions. We take this work to provide evidence that, rather than being wastefully overinformative, speakers are usefully redundant.

\appendix

\section{Effects of semantic value on utterance probabilities}
\label{app:modelexploration}

\begin{figure}
\centering
\includegraphics[width=.82\textwidth]{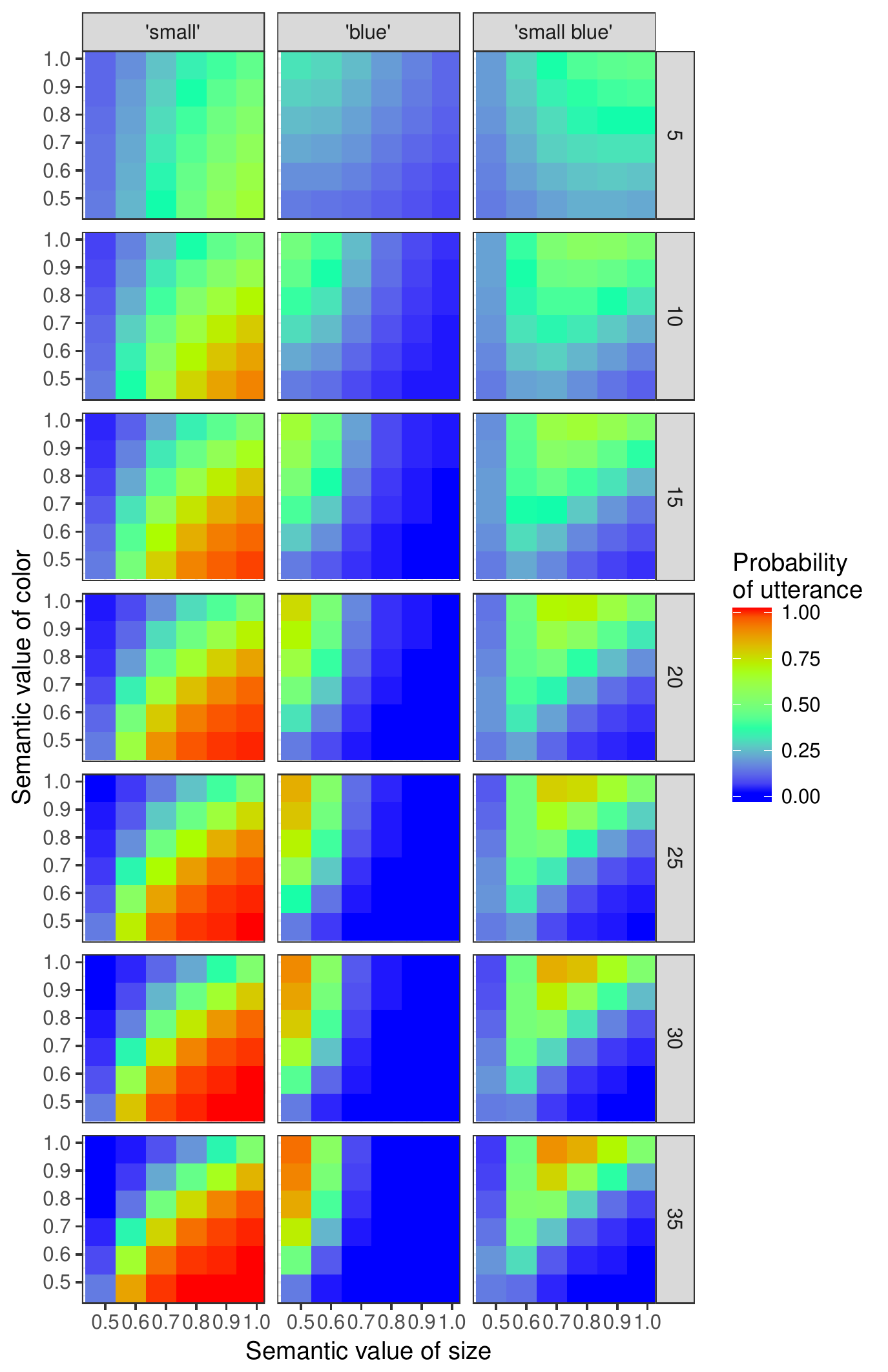}
\caption{Probability of producing sufficient \emph{small pin}, insufficient \emph{blue pin}, and redundant \emph{small blue pin} in contexts as depicted in \figref{fig:sizesufficient}, as a function of semantic value of color and size utterances and varying $\beta_i$ row-wise (for $ \beta_c = 0$).}
\label{fig:fullexploration}
\end{figure}

Here we visualize the effect of different adjective types' semantic value on the probability of producing the insufficient color-only utterance (\emph{blue pin}), the sufficient size-only utterance (\emph{small pin}), or the redundant color-and-size utterance  (\emph{small blue pin}) to refer to the target in context \figref{fig:sizesufficient} under varying $\beta_i$ values, in \figref{fig:fullexploration}. This constitutes a generalization of \figref{fig:basicasymmetry}, which is duplicated in row 6 ($\beta_i = 30$).

\section{Pre-experiment quiz}
\label{app:numdistractors}

Before continuing to the main experiment, each participant was required to correctly respond ``True'' or ``False'' to the following statements. Correct answers are given in parentheses after the statement.

\begin{itemize}
	\item The speaker can click on an object. (False)
	\item The listener wants to click on the object that the speaker is
  telling them about. (True)
  \item  The target is the object which has the red circle around it. (False)
  \item Only the speaker can send messages. (False)
  \item There are a total of 72 rounds. (True)
  \item The locations of the three objects are the same for the speaker and the listener. (False)
\end{itemize}

\section{Exp.~1 items}
\label{app:itemtypes}

\tableref{tab:exp1items} lists the 36 object types from Exp.~1 and the colors they appeared in:

\begin{table}[h]
\caption{Exp.~1 items and the colors they appeared in.}
\centering
\begin{tabular}{l l l l}
\toprule
Object & Colors & Object & Colors \\
\midrule
avocado & black, green & balloon & pink, yellow \\
belt & black, brown & bike & purple, red\\
billiard ball & orange, purple & binder & blue, green \\
book & black, blue & bracelet & green, purple \\
bucket & pink, red & butterfly & blue, purple\\
candle & blue, red & cap & blue, orange \\
chair & green, red & coat hanger & orange, purple \\
comb & black, blue & cushion & blue, orange\\
flower & purple, red & frame & green, pink \\
golf ball & blue, pink & guitar & blue, green\\
hair dryer & pink, purple & jacket & brown, green\\
napkin & orange, yellow & ornament & blue, purple\\
pepper & green, red & phone & pink, white\\
rock & green, purple & rug & blue, purple \\
shoe & white, yellow & stapler & purple, red\\
thumb tack & blue, red & tea cup & pink, white \\
toothbrush & blue, red & turtle & black, brown \\
wedding cake & pink, white & yarn & purple, red\\
\bottomrule
\end{tabular}
\label{tab:exp1items}
\end{table}

\section{Data pre-processing and exclusions}
\label{sec:preprocessing}

\subsection{Exp.~1}

Median completion time was 10 minutes. One participant was excluded because their native language was not English (but they participated in the listener role, so their exclusion was of no consequence to the analysis). 83\% of participants thought their partner was human. Participants were not excluded if they didn't believe their partner was human.

We collected data from 2177 critical trials. Because we did not restrict participants' utterances in any way, they produced many different kinds of referring expressions. Testing the model's predictions required, for each trial, classifying the produced utterance as an instance of a \emph{color}-only mention, a \emph{size}-only mention, or a \emph{color-and-size} mention (or excluding the trial if no classification was possible). To this end we conducted the following semi-automatic data pre-processing. 

An R script first automatically checked whether the speaker's utterance contained a precoded color (i.e. \emph{black, blue, brown, gold, green, orange, pink, purple, red, silver, violet, white, yellow}) or size (i.e. \emph{big, bigger, biggest, huge, large, larger, largest, little, small, smaller, smallest, tiny}) term. In this way, 95.7 \% of cases were classified as mentioning size and/or color. However, this did not capture that sometimes, participants produced meaning-equivalent modifications of color/size terms for instance by adding suffixes (e.g., \emph{bluish}), using abbreviations (e.g., \emph{lg} for \emph{large} or \emph{purp} for \emph{purple}), or using non-precoded color labels (e.g., \emph{lime} or \emph{lavender}). Expressions containing a typo (e.g., \emph{pruple} instead of \emph{purple}) could also not be classified automatically. In the next step, one of the authors (CG) therefore manually checked the automatic coding to include these kinds of modifications in the analysis. This covered another 1.9\% of trials. Most of the time, participants converged on a convention of producing only the target's size and/or color, e.g., \emph{purple} or \emph{big blue}, but not a determiner (e.g., \emph{the}) or the noun corresponding to the object's type  (e.g., \emph{comb}). Determiners were omitted in 88.6 \% of cases and nouns were omitted in 71.6 \% of cases. We did not analyze this any further.

There were 50 cases (2.3\%) in which the speaker made reference to the distinguishing dimension in an abstract way, e.g.~\emph{different color}, \emph{unique one}, \emph{ripest}, \emph{very girly}, or \emph{guitar closest to viewer}. While interesting as utterance choices,\footnote{Certain participants seemed to have deliberately used this as a strategy even though simply mentioning the distinguishing property would have been shorter in most cases. In all, only 12 participants produced these kinds of utterances: one 18 times, one 8 times, one 6 times, two 3 times, one 2 times, and the remaining six only once each.} these cases were excluded from the analysis. There were 3 cases that were nonsensical, e.g. \emph{bigger off a shade}, which were also excluded. In 6 cases only the insufficient dimension was mentioned -- these were excluded from the analysis reported in the next section, where we are only interested in minimal or redundant utterances, not underinformative ones, but were included in the Bayesian data analysis reported in \sectionref{sec:modifiermodeleval}. Finally, we excluded six trials where the speaker did not produce any utterances, and 33 trials on which the listener selected the wrong referent, leading to the elimination of 1.5\% of trials. After the exclusion, 2076 cases classified as one of \emph{color}, \emph{size}, or \emph{color-and-size} entered the analysis.

\subsection{Exp.~2}

Median completion time was 7 minutes. All participants self-reported English as their native language. 93\% of participants thought their partner was human. Participants were not excluded if they believed their partner was human.

Two participant-pairs were excluded because they did not finish the experiment and therefore could not receive payment. Trials on which the speaker did not produce any utterances were also excluded, resulting in the exclusion of two additional participant-pairs.
Finally, there were 10 speakers who consistently used roundabout descriptions instead of direct referring expressions (e.g., \emph{monkeys love\dots} to refer to banana).\footnote{It is unclear whether these participants misunderstood task instructions, or were simply being playful. This is interestingly different from Exp.~1, where participants did not produce such descriptions, presumably because the object type was identical and therefore there were no differences between objects except for color and size. We speculate that on average the functional differences between objects that differ in type are greater than between those that differ in color or size, with the former consequently being more inspiring for the generation of creative referring expressions.} These pairs were also excluded, since such indirect expressions do not inform our questions about modifier production.

We analyzed data from 1974 trials. Just as in Exp.~1, participants communicated freely, which led to a vast amount of different referring expressions. To test the model's predictions, the utterance produced for each trial was to be classified as belonging to one of the following categories: \textit{type-only} (``banana''), \textit{color-and-type} (``yellow banana''), and \textit{color-only} (``yellow'') utterances. Referring expressions that included superordinate categories (``yellow fruit''), descriptions (``has green stem''), color-circumscriptions (``funky carrot''), and negations (``yellow but not banana'') were regarded as \textit{other} and excluded. To this end we conducted the following semi-automatic data pre-processing.

The referring expressions were analyzed similarly to Exp.~1. First, 32 trials (1.6\%) were excluded because the listener selected the wrong referent. 109 trials (5.6\%) were excluded because the referring expressions included one of the exceptional cases described above (e.g., using negations). 
An R script then automatically checked the remaining 1833 utterances for whether they contained a precoded color term (i.e. \emph{green, purple, white, black, brown, yellow, orange, blue, pink, red, grey}) or type (i.e. \emph{apple, banana, carrot, tomato, pear, pepper, avocado}). This way, 96.5\% of the remaining cases were classified as mentioning type and/or color. 

However, this did not capture that sometimes, participants produced meaning-equivalent modifications of color/type terms for instance by adding suffixes (e.g., \emph{pinkish}), using abbreviations (e.g., \emph{yel} for \emph{yellow}), or using non-precoded color and type labels (e.g., \emph{lavender} or \emph{jalapeno}). In addition, expressions that contained a typo (e.g., \emph{blakc} instead of \emph{black}) could also not be classified automatically. One of the authors (EK) therefore manually hand-coded these cases.
There were 6 cases (0.3\%) that could not be categorized and were excluded.
Overall, 1827 utterances were classified as one of \emph{color}, \emph{type}, or \emph{color-and-type} entered the analysis.

\subsection{Exp.~3}

We collected 2193 referring expressions. To determine the level of reference for each trial, we followed the following procedure. First, speakers' and listeners' messages were parsed automatically; the referring expression used by the speaker was extracted for each trial and checked for whether it contained the current target's correct \emph{sub}(ordinate), \emph{basic}, or \emph{super}(ordinate) level term using a simple grep search. In this way, 71.4\% of trials were labelled as mentioning a pre-coded level of reference. In the next step, remaining utterances were checked manually by one of the authors (CG) to determine whether they contained a correct level of reference term which was not detected by the grep search due to typos or grammatical modification of the expression. In this way, meaning-equivalent alternatives such as \emph{doggie} for \emph{dog}, or reduced forms such as \emph{gummi}, \emph{gummies} and \emph{bears} for \emph{gummy bears} were counted as containing the corresponding level of reference term. This covered another 15.0\% of trials. 41 trials on which the listener selected the wrong referent were excluded, leading to the elimination of 2.1\% of trials. Six trials were excluded because the speaker did not produce any utterances. Additionally, a total of 12.5\% of correct trials were excluded because the utterance consisted only of an attribute of the superclass (\emph{the living thing} for \emph{animal}), of the basic level (\emph{can fly} for \emph{bird}), of the subcategory (\emph{barks} for \emph{dog}) or of the particular instance (\emph{the thing facing left}) rather than a category noun. These kinds of attributes were also mentioned in addition to the noun on trials which were included in the analysis for 8.9\% of sub level terms, 18.9\% of basic level terms, and 60.9\% of super level terms. On 1.2\% of trials two different levels of reference were mentioned; in this case the more specific level of reference was counted as being mentioned in this trial. After all exclusion and pre-processing, 1872 cases classified as one of \emph{sub}, \emph{basic}, or \emph{super} entered into the analysis.

\section{Typicality effects in Exp.~1}
\label{sec:exp1typicality}

To assess whether we replicate the color typicality effects previously reported in the literature \cite{sedivy2003a, Westerbeek2015, rubiofernandez2016}, we elicited color typicality norms for each of the items in Exp.~1 and then included typicality as an additional predictor of redundant adjective use in the regression analysis reported in \sectionref{sec:modelempiricalresults}. 

\subsection{Methods}

\subsubsection{Participants}

We recruited 60 participants over Amazon's Mechanical Turk who were each paid \$0.25 for their participation.

\subsubsection{Procedure and materials}

\begin{figure}
\centering
\includegraphics[width=.5\textwidth]{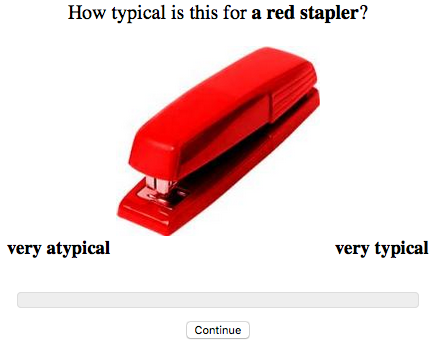}
\caption{A modified example trial from the typicality elicitation experiment.}
\label{fig:modifiertypstimulus}
\end{figure}

On each trial, participants saw one of the big versions of the items used in Exp.~1 and were asked to answer the question ``How typical is this for an \emph{X}?'' on a continuous slider with endpoints labeled ``very atypical'' to ``very typical.'' \emph{X} was a referring expression consisting of either only the correct noun (e.g., \emph{stapler}) or the noun modified by the correct color (e.g., \emph{red stapler}). \figref{fig:modifiertypstimulus} shows an example of a modified trial.

Each participant saw each of the 36 objects once. An object was randomly displayed in one of the two colors it occurred with in Exp.~1 and was randomly displayed with either the correct modified utterance or the correct unmodified utterance, in order to obtain roughly equal numbers of object-utterance combinations.

Importantly, we only elicited typicality norms for unmodified utterances and utterances with color modifiers, but not utterances with size modifiers. This was because it is impossible to obtain size typicality norms for objects presented in isolation, due to the inherently relational nature of size adjectives. Consequently, we only test for the effect of typicality on \emph{size-sufficient} trials, i.e.~when color is redundant. 

\subsection{Results and discussion}

\begin{figure}
\centering
\includegraphics[width=.9\textwidth]{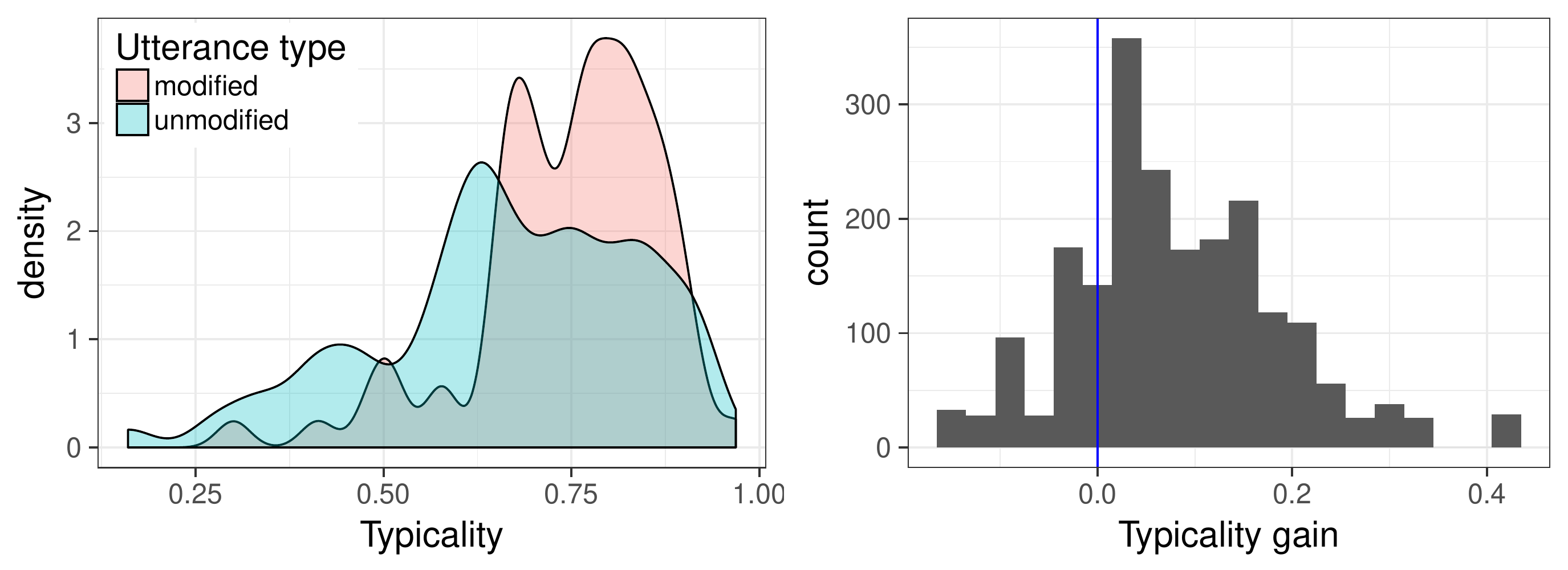}
\caption{Typicality densities for modified and unmodified utterances (left) and histogram of typicality gains (differences between modified and unmodified typicalities, right).}
\label{fig:typicalitydists}
\end{figure}

We coded the slider endpoints as 0 (``very atypical'') and 1 (``very typical''), essentially treating each response as a typicality value between 0 and 1. For each combination of object, color, and utterance (modified/unmodified), we computed that item's mean. Mean typicalities were generally lower for unmodified than for modified utterances: mean typicality for unmodified utterances was .67 (sd=.17, mode=.76) and for modified utterances .75 (sd=.12, mode=.81). This can also be seen on the left in \figref{fig:typicalitydists}. Note that, as expected given how the stimuli were constructed, typicality was generally skewed towards the high end, even for unmodified utterances. This means that there was not much variation in  the difference in typicality between modified and unmodified utterances. We will refer to this difference as \emph{typicality gain}, reflecting the overall gain in typicality via color modification over the unmodified baseline. As can be seen on the right in \figref{fig:typicalitydists}, in most cases typicality gain was close to zero.

This makes the typicality analysis difficult: if typicality gain is close to zero for most cases (and, taking into account confidence intervals, effectively zero), it is hard to evaluate the effect of typicality on redundant adjective use. In order to maximize power, we therefore conducted the analysis only on those items for which for at least one color the confidence intervals for the modified and unmodified utterances did not overlap. There were only four such cases: \emph{(pink) golfball}, \emph{(pink) wedding cake}, \emph{(green) chair}, and \emph{(red) stapler}, for a total of 231 data points.

\begin{figure}
\centering
\includegraphics[width=.9\textwidth]{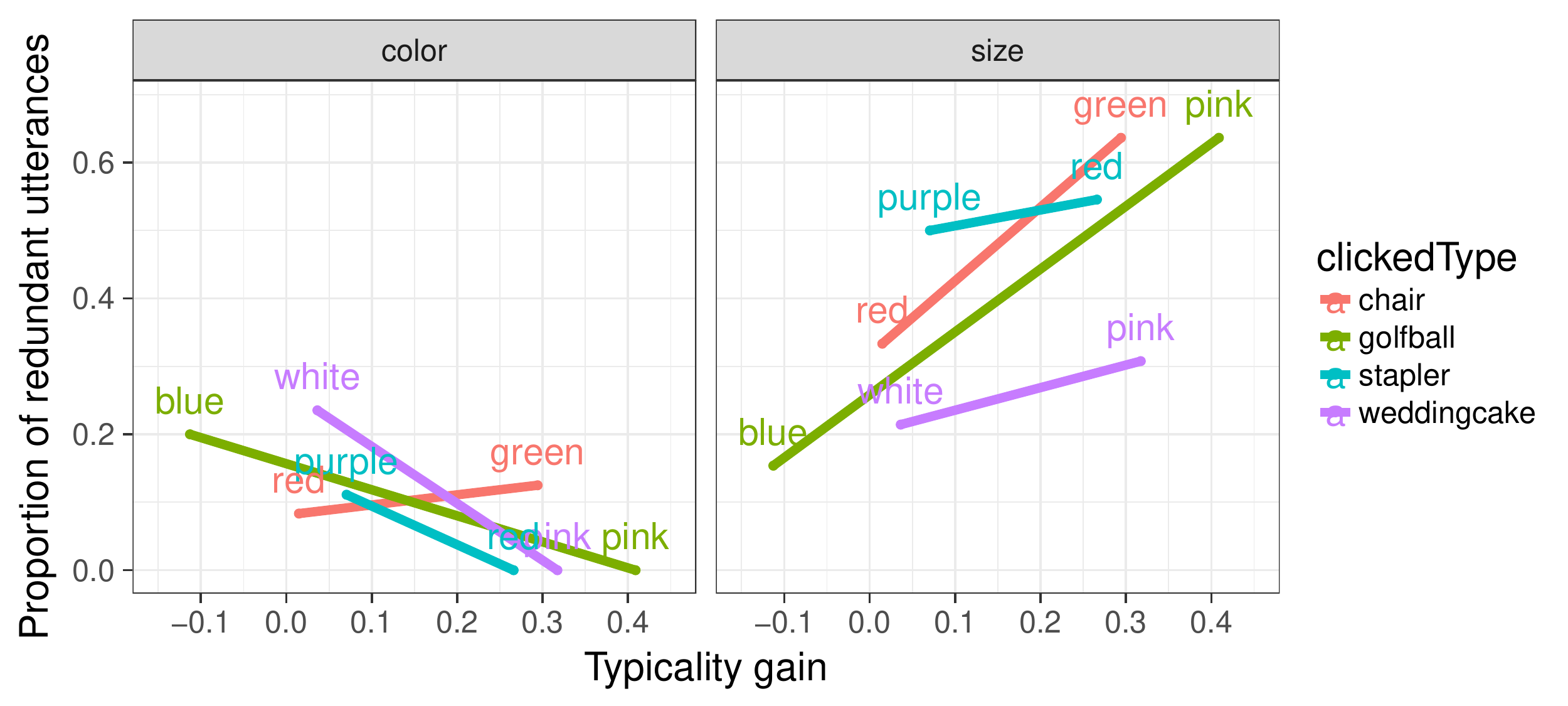}
\caption{Utterance probability for four items as a function of difference in typicality between modified and unmodified utterance (x-axis) and sufficient dimension (columns). }
\label{fig:maxtypicalitydiff}
\end{figure}

Predictions differ for size-sufficient and color-sufficient trials. Given the typicality effects reported in the literature and the predictions of cs-RSA, we expect greater redundant color use on size-sufficient trials with \emph{increasing} typicality gain. The predictions for redundant size use on color-sufficient trials are unclear from the previous literature. Cs-RSA, however,  predicts greater redundant size use with \emph{decreasing} typicality gain: small color typicality gains reflect the relatively low out-of-context utility of color. In these cases, it may be useful to redundantly use a size modifier even if that modifier is noisy. If borne out, these predictions should surface in an interaction between sufficient property and typicality gain. Visual inspection of the empirical proportions of redundant adjective use in \figref{fig:maxtypicalitydiff} suggests that this pattern is indeed borne out.

In order to investigate the effect of typicality gain on redundant adjective use, we conducted a mixed effects logistic regression analysis predicting redundant over minimal adjective use from fixed effects of scene variation, sufficient dimension, the interaction of scene variation and sufficient property, and the interaction of typicality gain and sufficient property. This is the same model as reported in \sectionref{sec:modelempiricalresults}, with the only difference that the interaction between sufficient property and typicality gain was added. All predictors were centered before entering the analysis. The model contained the most sophisticated random effects structure that allowed it to converge: by-participant and by-item (where item was a color-object combination) random intercepts. 

The model summary is shown in \tableref{tab:colortypicalityresults}. We replicate the effects of sufficient property and scene variation observed earlier on this smaller dataset. Crucially, we observe a significant interaction between sufficient property and typicality gain.\footnote{Conducting the same analysis on the entire dataset (i.e., using all of the noisy typicality estimates, replicated the scene variation and sufficient property effects. The interaction of typicality gain and sufficient property went in the same direction numerically, but failed to reach significance ($\beta = 1.52$, $SE = 1.45$, $p < .29$).} Simple effects analysis reveals that this interaction is due to a positive effect of typicality gain on redundant adjective use in the size-sufficient condition ($\beta = 4.47$, $SE = 1.65$, $p < .007$) but a negative effect of typicality gain on redundant adjective use in the color-sufficient condition  ($\beta = -5.77$, $SE = 2.49$, $p < .03$). 

\begin{table}[!tbp]
\caption{Model coefficients, standard errors, and p-values. Significant p-values are bolded.}
\begin{center}
\begin{tabular}{lrrl}
\toprule
\multicolumn{1}{l}{}&\multicolumn{1}{c}{Coef $\beta$}&\multicolumn{1}{c}{SE($\beta$)}&\multicolumn{1}{c}{$p$}\tabularnewline
\midrule
Intercept&$-1.85$&$0.34$&\textbf{\textless .0001}\tabularnewline
Scene variation&$ 4.29$&$1.16$&\textbf{\textless .001}\tabularnewline
Sufficient property&$ 2.72$&$0.60$&\textbf{\textless .0001}\tabularnewline
Scene variation : Sufficient property&$ 0.88$&$2.12$&\textless 0.68\tabularnewline
Sufficient property : Typicality gain&$ 9.43$&$2.68$&\textbf{\textless .001}\tabularnewline
\bottomrule
\end{tabular}\end{center}
\label{tab:colortypicalityresults}
\end{table}

 An important point is of note: the typicality elicitation procedure we employed here is somewhat different from that employed by \citeA{Westerbeek2015}, who asked their participants ``How typical is this color for this object?'' We did this for conceptual reasons: the values that go into the semantics of the RSA model are most easily conceptualized as the typicality of an object as an instance of an utterance. While the typicality of a feature for an object type no doubt plays into how good of an instance of the utterance the object is, deriving our typicalities from the  statistical properties of the subjective distributions of features over objects is beyond the scope of this paper. However, in a separate experiment we did ask participants the Westerbeek question. The correlation between mean typicality ratings from the Westerbeek version and the unmodified ``How typical is this for \emph{X}'' version was .75. The correlation between the Westerbeek version and the modified version was .64. The correlation between the Westerbeek version and typicality gain was -.52.

For comparison, including typicality means obtained via the Westerbeek question as a predictor instead of typicality gain on the four high-powered items replicated the significant interaction between typicality and sufficient property ($\beta = -6.77$, $SE = 1.88$, $p < .0003$). Simple effects analysis revealed that the interaction is again due to a difference in slope in the two sufficient property conditions: in the size-sufficient condition, color is less likely to be mentioned with increasing color typicality   ($\beta = -3.66$, $SE = 1.18$, $p < .002$), whereas in the color-sufficient condition, size is more likely to be mentioned with increasing color typicality ($\beta = 3.09$, $SE = 1.45$, $p < .04$).\footnote{Again, conducting this analysis on the entire dataset yielded only a marginal interaction of sufficient property and color typicality in the right direction ($\beta = -1.10$, $SE = .64$, $p < .09$).}

We thus overall find moderate evidence for typicality effects in our dataset. Typicality effects are strong for those items that clearly display typicality differences between the modified and unmodified utterance, but much weaker for the remaining items. That the evidence for typicality effects is relatively scarce is no surprise: the stimuli were specifically designed to minimize effects of typicality. However, the fact that both ways of quantifying typicality predicted redundant adjective use in the expected direction suggests that with more power or with stimuli that exhibit greater typicality variation, these effects may show up more clearly.

\section{Experiment 3 items}
\label{app:taxonomicstimuli}

\tableref{tab:exp3items} lists all items used in Exp.~3 and the mean empirical utterance lengths that participants produced to refer to them:

\begin{table}
\centering
\caption{List of domains and associated superordinate category, target stimuli, and mean length (standard deviation) in characters of actually produced subordinate level utterances in Exp.~3.}
	\label{tab:reflevelstimuli}
	\begin{tabular}{l l l l}
	\toprule
	Domain & Super & Targets & Mean sub length (sd)\\
	\midrule
	\multirow{4}{*}{bear} & \multirow{4}{*}{animal} & black bear & 9.9 (.14)\\
	& & polar bear & 8.8 (.35)\\
	& & panda bear & 5.5 (.2)\\
	& & grizzly bear & 9 (.98)\\
	\midrule
	\multirow{4}{*}{bird} & \multirow{4}{*}{animal} & eagle & 4.9 (.1)\\
	& 	& parrot & 6.1 (.13)\\
	& & pigeon & 5.9 (.22)\\
	& 	& hummingbird & 10.1 (.5)\\
	\midrule
	\multirow{4}{*}{candy} & \multirow{4}{*}{snack} & MnMs & 4.4 (.49)\\
		& & skittles & 6.9 (.43)\\
		& & gummy bears & 8.5 (.47)\\
		& & jelly beans & 9.3 (.44)\\
	\midrule
	\multirow{4}{*}{car} & \multirow{4}{*}{vehicle} & SUV & 3 (0)\\
		& & minivan & 5.7 (.27)\\
		& & sports car & 9.8 (.23)\\
		& & convertible & 11.1 (.2)\\
	\midrule
	\multirow{4}{*}{dog} & \multirow{4}{*}{animal} & pug & 3 (.08)\\
		& & husky & 4.7 (.22)\\
		& & dalmatian & 8.8 (.18)\\
		& & German Shepherd & 13.1 (.82)\\
	\midrule
	\multirow{4}{*}{fish} & \multirow{4}{*}{animal} & catfish & 6.6 (.4)\\
		& & goldfish & 7.9 (.22)\\
		& & swordfish & 8 (.43)\\
		& & clownfish & 9.1 (.38)\\
	\midrule
	\multirow{4}{*}{flower} & \multirow{4}{*}{plant} & rose & 4 (0)\\
		& & tulip & 4.4 (.18)\\
		& & daisy & 5.9 (.55)\\
		& & sunflower & 9 (.11)\\
	\midrule
	\multirow{4}{*}{shirt} & \multirow{4}{*}{clothing} & T-shirt & 6.4 (.48)\\
		& & polo shirt & 6.7 (.79)\\
		& & dress shirt & 11 (0)\\
		& & Hawaii shirt & 12.6 (.46)\\
	\midrule
	\multirow{4}{*}{table} & \multirow{4}{*}{furniture} & picnic table & 9.7 (.58)\\
		& & dining table & 12 (0)\\
		& & coffee table & 9.1 (.95)\\
		& & bedside table & 8.3 (.68)\\				
	\bottomrule
	\end{tabular}
	\label{tab:exp3items}
\end{table}

\section{Typicality norms for Experiment 3}
\label{app:typicalitynorms2}

Analogous to the color typicality norms elicited for utterances in Exps.~1-2, we elicited typicality norms for utterances in Exp.~3. The elicited typicalities were used in the mixed effects analyses and Bayesian Data Analysis reported in \sectionref{sec:nominal}.

\subsubsection{Methods}

\paragraph{Participants}

We recruited 240 participants over Amazon's Mechanical Turk who were each paid \$0.50 for their participation.

\paragraph{Procedure and materials}

On each trial, participants saw one of the images used in Exp.~3 and were asked to answer the question ``How typical is this for an \emph{X}?'' on a continuous slider with endpoints labeled ``very atypical'' to ``very typical.'' \emph{X} was a nominal referring expression. We did not test all utterance-object combinations, which would have led to an explosion of conditions. Instead, we tested each target object with its three utterances (e.g., the dalamtian was paired with \emph{dalmatian}, \emph{dog}, and \emph{animal}; the pug was paired with \emph{pug}, \emph{dog}, and \emph{animal}, etc.). That yielded a total of 108 combinations -- four targets in nine domains with three utterances each. We further tested each distractor item that shared the target's superordinate category (\emph{dist-samesuper}, e.g., elephants share the superordinate category animal with dogs) on both the basic level and the superordinate level term (e.g., \emph{dog} for elephant and \emph{animal} for elephant), for a total of 469 combinations. Finally, we also tested each distractor of a different superordinate category than the target on the target's superordinate level term (\emph{dist-diffsuper}, e.g., \emph{animal} for rose). This yielded another 168 combinations. Overall, we obtained typicality norms for 745 object-utterance combinations. All other object-utterance combinations were assumed to have typicality 0. Each participant rated 45 items: 7 targets, 10 dist-diffsuper, and 28 dist-samesuper cases. These were randomly sampled from the overall pool of items in each category. 

\subsubsection{Results and discussion}

Each combination was rated at least 5 times and at most 27 times. We coded the slider endpoints as 0 (``very atypical'') and 1 (``very typical''). In order to evaluate the model, we used each object-utterance combination's typicality mean as input. 

\begin{figure}[h!]
\centering
\includegraphics[width=\textwidth]{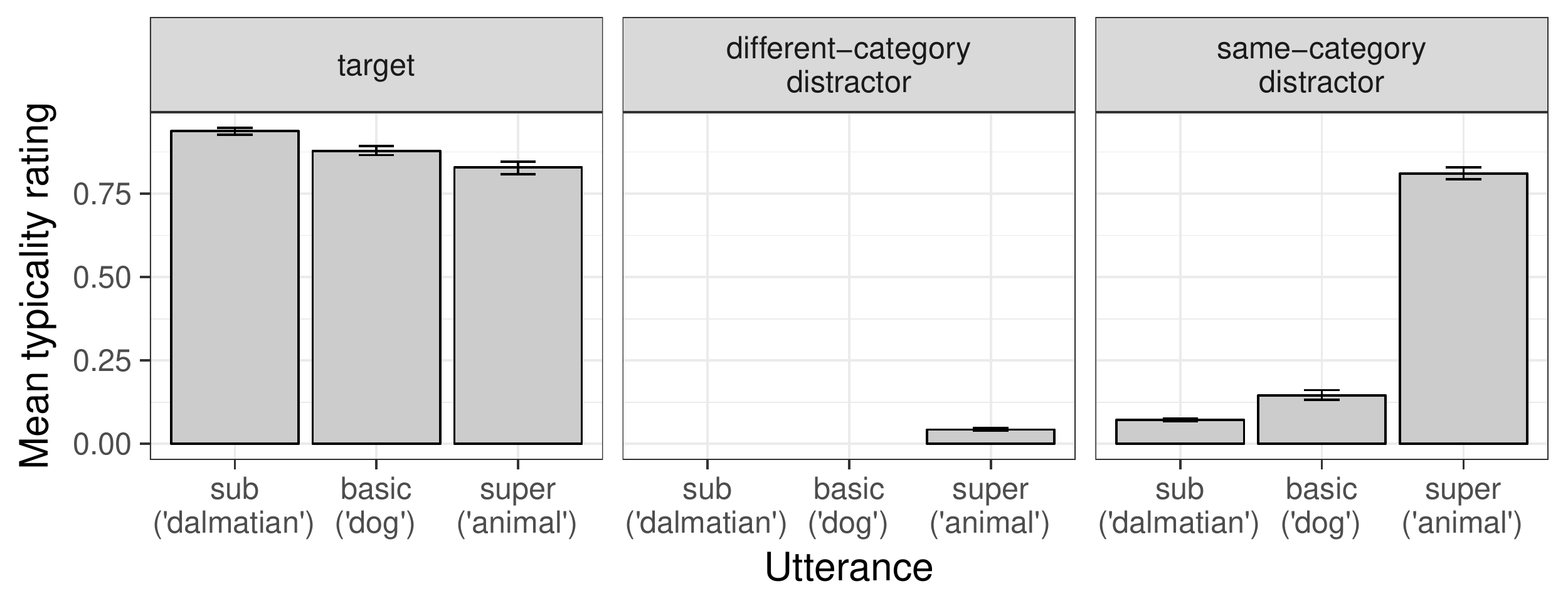}
\caption{Mean typicality ratings by utterance (target subordinate, basic, and superordinate level term) for targets (e.g., \emph{dalmatian}, left panel), distractors with a different superordinate category from the target (e.g., \emph{rose}, middle panel), and distractors with the same superordinate category as the target (e.g., \emph{elephant}, right panel). Error bars indicate bootstrapped 95\% confidence intervals.}
\label{fig:typicalityboxplots}
\end{figure}

Typicality ratings by item type (target, dist-samesuper, dist-diffsuper) and utterance type (sub, basic, super) are visualized in \figref{fig:typicalityboxplots}. As expected, typicality was close to 0 for distractor items with a different superordinate category as the target, and for subordinate/basic level terms used with distractors of the same superordinate category. However, even for these cases, there was some variation. 

For targets, typicality of the object for the utterance decreased with increasing reference level, mirroring the typicality ratings obtained for Exp.~1 -- a particular object is a better instance of the more specific term than of the more general term for that object.

\bibliographystyle{apacite}

\setlength{\bibleftmargin}{.125in}
\setlength{\bibindent}{-\bibleftmargin}

\bibliography{bibs}

\begin{thebibliography}{}

\bibitem [\protect \citeauthoryear {%
Arts%
, Maes%
, Noordman%
\BCBL {}\ \BBA {} Jansen%
}{%
Arts%
\ \protect \BOthers {.}}{%
{\protect \APACyear {2011}}%
}]{%
Arts2011}
\APACinsertmetastar {%
Arts2011}%
\begin{APACrefauthors}%
Arts, A.%
, Maes, A.%
, Noordman, L.%
\BCBL {}\ \BBA {} Jansen, C.%
\end{APACrefauthors}%
\unskip\
\newblock
\APACrefYearMonthDay{2011}{}{}.
\newblock
{\BBOQ}\APACrefatitle {{Overspecification facilitates object identification}}
  {{Overspecification facilitates object identification}}.{\BBCQ}
\newblock
\APACjournalVolNumPages{Journal of Pragmatics}{43}{1}{361--374}.
\newblock
\begin{APACrefDOI} \doi{10.1016/j.pragma.2010.07.013} \end{APACrefDOI}
\PrintBackRefs{\CurrentBib}

\bibitem [\protect \citeauthoryear {%
Baker%
, Gill%
\BCBL {}\ \BBA {} Cassell%
}{%
Baker%
\ \protect \BOthers {.}}{%
{\protect \APACyear {2008}}%
}]{%
Baker2008}
\APACinsertmetastar {%
Baker2008}%
\begin{APACrefauthors}%
Baker, R.%
, Gill, A\BPBI J.%
\BCBL {}\ \BBA {} Cassell, J.%
\end{APACrefauthors}%
\unskip\
\newblock
\APACrefYearMonthDay{2008}{}{}.
\newblock
{\BBOQ}\APACrefatitle {{Reactive redundancy and listener comprehension in
  direction-giving}} {{Reactive redundancy and listener comprehension in
  direction-giving}}.{\BBCQ}
\newblock
\BIn{} \APACrefbtitle {9th SIGdial Workshop on Discourse and Dialogue} {9th
  sigdial workshop on discourse and dialogue}\ (\BPGS\ 37--45).
\newblock
\begin{APACrefDOI} \doi{10.3115/1622064.1622071} \end{APACrefDOI}
\PrintBackRefs{\CurrentBib}

\bibitem [\protect \citeauthoryear {%
Bates%
, M{\"a}chler%
, Bolker%
\BCBL {}\ \BBA {} Walker%
}{%
Bates%
\ \protect \BOthers {.}}{%
{\protect \APACyear {2015}}%
}]{%
lme4}
\APACinsertmetastar {%
lme4}%
\begin{APACrefauthors}%
Bates, D.%
, M{\"a}chler, M.%
, Bolker, B.%
\BCBL {}\ \BBA {} Walker, S.%
\end{APACrefauthors}%
\unskip\
\newblock
\APACrefYearMonthDay{2015}{}{}.
\newblock
{\BBOQ}\APACrefatitle {Fitting Linear Mixed-Effects Models Using {lme4}}
  {Fitting linear mixed-effects models using {lme4}}.{\BBCQ}
\newblock
\APACjournalVolNumPages{Journal of Statistical Software}{67}{1}{1--48}.
\newblock
\begin{APACrefDOI} \doi{10.18637/jss.v067.i01} \end{APACrefDOI}
\PrintBackRefs{\CurrentBib}

\bibitem [\protect \citeauthoryear {%
Belke%
\ \BBA {} Meyer%
}{%
Belke%
\ \BBA {} Meyer%
}{%
{\protect \APACyear {2002}}%
}]{%
Belke2002}
\APACinsertmetastar {%
Belke2002}%
\begin{APACrefauthors}%
Belke, E.%
\BCBT {}\ \BBA {} Meyer, A\BPBI S.%
\end{APACrefauthors}%
\unskip\
\newblock
\APACrefYearMonthDay{2002}{}{}.
\newblock
{\BBOQ}\APACrefatitle {{Tracking the time course of multidimensional stimulus
  discrimination: Analyses of viewing patterns and processing times during
  “same”-“different“ decisions}} {{Tracking the time course of
  multidimensional stimulus discrimination: Analyses of viewing patterns and
  processing times during “same”-“different“ decisions}}.{\BBCQ}
\newblock
\APACjournalVolNumPages{European Journal of Cognitive
  Psychology}{14}{2}{237--266}.
\newblock
\begin{APACrefDOI} \doi{10.1080/09541440143000050} \end{APACrefDOI}
\PrintBackRefs{\CurrentBib}

\bibitem [\protect \citeauthoryear {%
Bergen%
, Levy%
\BCBL {}\ \BBA {} Goodman%
}{%
Bergen%
\ \protect \BOthers {.}}{%
{\protect \APACyear {2016}}%
}]{%
bergen2016}
\APACinsertmetastar {%
bergen2016}%
\begin{APACrefauthors}%
Bergen, L.%
, Levy, R.%
\BCBL {}\ \BBA {} Goodman, N.%
\end{APACrefauthors}%
\unskip\
\newblock
\APACrefYearMonthDay{2016}{}{}.
\newblock
{\BBOQ}\APACrefatitle {{Pragmatic reasoning through semantic inference}}
  {{Pragmatic reasoning through semantic inference}}.{\BBCQ}
\newblock
\APACjournalVolNumPages{Semantics and Pragmatics}{9}{1984}{1--46}.
\newblock
\begin{APACrefDOI} \doi{10.3765/sp.9.20} \end{APACrefDOI}
\PrintBackRefs{\CurrentBib}

\bibitem [\protect \citeauthoryear {%
Bernardy%
, Blanck%
, Chatzikyriakidis%
\BCBL {}\ \BBA {} Lappin%
}{%
Bernardy%
\ \protect \BOthers {.}}{%
{\protect \APACyear {2018}}%
}]{%
Bernardy2018}
\APACinsertmetastar {%
Bernardy2018}%
\begin{APACrefauthors}%
Bernardy, J\BHBI P.%
, Blanck, R.%
, Chatzikyriakidis, S.%
\BCBL {}\ \BBA {} Lappin, S.%
\end{APACrefauthors}%
\unskip\
\newblock
\APACrefYearMonthDay{2018}{}{}.
\newblock
{\BBOQ}\APACrefatitle {{A Compositional Bayesian Semantics for Natural
  Language}} {{A Compositional Bayesian Semantics for Natural
  Language}}.{\BBCQ}
\newblock
\BIn{} \APACrefbtitle {Proceedings of the First International Workshop on
  Language Cognition and Computational Models} {Proceedings of the first
  international workshop on language cognition and computational models}\
  (\BPGS\ 1--10).
\newblock
\APACaddressPublisher{Santa Fe, New Mexico}{}.
\PrintBackRefs{\CurrentBib}

\bibitem [\protect \citeauthoryear {%
Bloomfield%
}{%
Bloomfield%
}{%
{\protect \APACyear {1933}}%
}]{%
bloomfield1933}
\APACinsertmetastar {%
bloomfield1933}%
\begin{APACrefauthors}%
Bloomfield, L.%
\end{APACrefauthors}%
\unskip\
\newblock
\APACrefYear{1933}.
\newblock
\APACrefbtitle {{Language}} {{Language}}.
\newblock
\APACaddressPublisher{New York}{Holt}.
\PrintBackRefs{\CurrentBib}

\bibitem [\protect \citeauthoryear {%
Brennan%
\ \BBA {} Clark%
}{%
Brennan%
\ \BBA {} Clark%
}{%
{\protect \APACyear {1996}}%
}]{%
brennan1996}
\APACinsertmetastar {%
brennan1996}%
\begin{APACrefauthors}%
Brennan, S\BPBI E.%
\BCBT {}\ \BBA {} Clark, H\BPBI H.%
\end{APACrefauthors}%
\unskip\
\newblock
\APACrefYearMonthDay{1996}{nov}{}.
\newblock
{\BBOQ}\APACrefatitle {{Conceptual pacts and lexical choice in conversation.}}
  {{Conceptual pacts and lexical choice in conversation.}}{\BBCQ}
\newblock
\APACjournalVolNumPages{Journal of experimental psychology. Learning, memory,
  and cognition}{22}{6}{1482 -- 1493}.
\PrintBackRefs{\CurrentBib}

\bibitem [\protect \citeauthoryear {%
P.~Brown%
\ \BBA {} Dell%
}{%
P.~Brown%
\ \BBA {} Dell%
}{%
{\protect \APACyear {1987}}%
}]{%
brown1987}
\APACinsertmetastar {%
brown1987}%
\begin{APACrefauthors}%
Brown, P.%
\BCBT {}\ \BBA {} Dell, G.%
\end{APACrefauthors}%
\unskip\
\newblock
\APACrefYearMonthDay{1987}{}{}.
\newblock
{\BBOQ}\APACrefatitle {{Adapting Production to Comprehension : Mention of
  Instruments}} {{Adapting Production to Comprehension : Mention of
  Instruments}}.{\BBCQ}
\newblock
\APACjournalVolNumPages{Cognitive Psychology}{472}{}{441--472}.
\PrintBackRefs{\CurrentBib}

\bibitem [\protect \citeauthoryear {%
R.~Brown%
}{%
R.~Brown%
}{%
{\protect \APACyear {1958}}%
}]{%
brown1958words}
\APACinsertmetastar {%
brown1958words}%
\begin{APACrefauthors}%
Brown, R.%
\end{APACrefauthors}%
\unskip\
\newblock
\APACrefYear{1958}.
\newblock
\APACrefbtitle {Words and things.} {Words and things.}
\newblock
\APACaddressPublisher{}{Free Press}.
\PrintBackRefs{\CurrentBib}

\bibitem [\protect \citeauthoryear {%
Brown-Schmidt%
\ \BBA {} Heller%
}{%
Brown-Schmidt%
\ \BBA {} Heller%
}{%
{\protect \APACyear {2014}}%
}]{%
Brown-schmidt2014}
\APACinsertmetastar {%
Brown-schmidt2014}%
\begin{APACrefauthors}%
Brown-Schmidt, S.%
\BCBT {}\ \BBA {} Heller, D.%
\end{APACrefauthors}%
\unskip\
\newblock
\APACrefYearMonthDay{2014}{}{}.
\newblock
{\BBOQ}\APACrefatitle {{What language processing can tell us about perspective
  taking: A reply to Bezuidenhout (2013)}} {{What language processing can tell
  us about perspective taking: A reply to Bezuidenhout (2013)}}.{\BBCQ}
\newblock
\APACjournalVolNumPages{Journal of Pragmatics}{60}{}{279--284}.
\newblock
\begin{APACrefDOI} \doi{10.1016/j.pragma.2013.09.003} \end{APACrefDOI}
\PrintBackRefs{\CurrentBib}

\bibitem [\protect \citeauthoryear {%
B\"urkner%
}{%
B\"urkner%
}{%
{\protect \APACyear {2017}}%
}]{%
brms}
\APACinsertmetastar {%
brms}%
\begin{APACrefauthors}%
B\"urkner, P\BHBI C.%
\end{APACrefauthors}%
\unskip\
\newblock
\APACrefYearMonthDay{2017}{}{}.
\newblock
{\BBOQ}\APACrefatitle {{brms}: An {R} Package for Bayesian Multilevel Models
  Using {Stan}} {{brms}: An {R} package for bayesian multilevel models using
  {Stan}}.{\BBCQ}
\newblock
\APACjournalVolNumPages{Journal of Statistical Software}{80}{1}{1--28}.
\newblock
\begin{APACrefDOI} \doi{10.18637/jss.v080.i01} \end{APACrefDOI}
\PrintBackRefs{\CurrentBib}

\bibitem [\protect \citeauthoryear {%
Clark%
\ \BBA {} Bangerter%
}{%
Clark%
\ \BBA {} Bangerter%
}{%
{\protect \APACyear {2004}}%
}]{%
Clark2004}
\APACinsertmetastar {%
Clark2004}%
\begin{APACrefauthors}%
Clark, H\BPBI H.%
\BCBT {}\ \BBA {} Bangerter, A.%
\end{APACrefauthors}%
\unskip\
\newblock
\APACrefYearMonthDay{2004}{}{}.
\newblock
{\BBOQ}\APACrefatitle {{Changing Ideas about Reference}} {{Changing Ideas about
  Reference}}.{\BBCQ}
\newblock
\BIn{} I\BPBI A.~Noveck\ \BBA {} D.~Sperber\ (\BEDS), \APACrefbtitle
  {Experimental Pragmatics} {Experimental pragmatics}\ (\BPGS\ 25--49).
\newblock
\APACaddressPublisher{Basingstoke, UK}{Palgrave MacMillan}.
\newblock
\begin{APACrefDOI} \doi{10.1057/9780230524125_2} \end{APACrefDOI}
\PrintBackRefs{\CurrentBib}

\bibitem [\protect \citeauthoryear {%
Clark%
\ \BBA {} Murphy%
}{%
Clark%
\ \BBA {} Murphy%
}{%
{\protect \APACyear {1982}}%
}]{%
Clark1982}
\APACinsertmetastar {%
Clark1982}%
\begin{APACrefauthors}%
Clark, H\BPBI H.%
\BCBT {}\ \BBA {} Murphy, G\BPBI L.%
\end{APACrefauthors}%
\unskip\
\newblock
\APACrefYearMonthDay{1982}{}{}.
\newblock
{\BBOQ}\APACrefatitle {{Audience Design in Meaning and Reference}} {{Audience
  Design in Meaning and Reference}}.{\BBCQ}
\newblock
\APACjournalVolNumPages{Advances in Psychology}{9}{C}{287--299}.
\newblock
\begin{APACrefDOI} \doi{10.1016/S0166-4115(09)60059-5} \end{APACrefDOI}
\PrintBackRefs{\CurrentBib}

\bibitem [\protect \citeauthoryear {%
Cohen%
\ \BBA {} Murphy%
}{%
Cohen%
\ \BBA {} Murphy%
}{%
{\protect \APACyear {1984}}%
}]{%
cohen1984models}
\APACinsertmetastar {%
cohen1984models}%
\begin{APACrefauthors}%
Cohen, B.%
\BCBT {}\ \BBA {} Murphy, G\BPBI L.%
\end{APACrefauthors}%
\unskip\
\newblock
\APACrefYearMonthDay{1984}{}{}.
\newblock
{\BBOQ}\APACrefatitle {Models of concepts} {Models of concepts}.{\BBCQ}
\newblock
\APACjournalVolNumPages{Cognitive science}{8}{1}{27--58}.
\PrintBackRefs{\CurrentBib}

\bibitem [\protect \citeauthoryear {%
Cohn-Gordon%
, Goodman%
\BCBL {}\ \BBA {} Potts%
}{%
Cohn-Gordon%
\ \protect \BOthers {.}}{%
{\protect \APACyear {2018}}%
}]{%
cohn2018}
\APACinsertmetastar {%
cohn2018}%
\begin{APACrefauthors}%
Cohn-Gordon, R.%
, Goodman, N\BPBI D.%
\BCBL {}\ \BBA {} Potts, C.%
\end{APACrefauthors}%
\unskip\
\newblock
\APACrefYearMonthDay{2018}{}{}.
\newblock
{\BBOQ}\APACrefatitle {{An Incremental Iterated Response Model of Pragmatics}}
  {{An Incremental Iterated Response Model of Pragmatics}}.{\BBCQ}
\newblock

\PrintBackRefs{\CurrentBib}

\bibitem [\protect \citeauthoryear {%
Dale%
}{%
Dale%
}{%
{\protect \APACyear {1989}}%
}]{%
Dale1989}
\APACinsertmetastar {%
Dale1989}%
\begin{APACrefauthors}%
Dale, R.%
\end{APACrefauthors}%
\unskip\
\newblock
\APACrefYearMonthDay{1989}{}{}.
\newblock
{\BBOQ}\APACrefatitle {{Cooking up referring expressions}} {{Cooking up
  referring expressions}}.{\BBCQ}
\newblock
\APACjournalVolNumPages{Proceedings of the 27th Annual Meeting on Association
  for Computational Linguistics (ACL'89)}{}{}{68--75}.
\newblock
\begin{APACrefDOI} \doi{10.3115/981623.981632} \end{APACrefDOI}
\PrintBackRefs{\CurrentBib}

\bibitem [\protect \citeauthoryear {%
Dale%
\ \BBA {} Reiter%
}{%
Dale%
\ \BBA {} Reiter%
}{%
{\protect \APACyear {1995}}%
}]{%
dale1995}
\APACinsertmetastar {%
dale1995}%
\begin{APACrefauthors}%
Dale, R.%
\BCBT {}\ \BBA {} Reiter, E.%
\end{APACrefauthors}%
\unskip\
\newblock
\APACrefYearMonthDay{1995}{}{}.
\newblock
{\BBOQ}\APACrefatitle {{Computational Interpretations of the Gricean Maxims in
  the Generation of Referring Expressions}} {{Computational Interpretations of
  the Gricean Maxims in the Generation of Referring Expressions}}.{\BBCQ}
\newblock
\APACjournalVolNumPages{Cognitive Science}{19}{}{233 -- 263}.
\PrintBackRefs{\CurrentBib}

\bibitem [\protect \citeauthoryear {%
Davies%
\ \BBA {} Katsos%
}{%
Davies%
\ \BBA {} Katsos%
}{%
{\protect \APACyear {2013}}%
}]{%
Davies2013}
\APACinsertmetastar {%
Davies2013}%
\begin{APACrefauthors}%
Davies, C.%
\BCBT {}\ \BBA {} Katsos, N.%
\end{APACrefauthors}%
\unskip\
\newblock
\APACrefYearMonthDay{2013}{}{}.
\newblock
{\BBOQ}\APACrefatitle {{Are speakers and listeners only moderately Gricean? An
  empirical response to Engelhardt et al. (2006)}} {{Are speakers and listeners
  only moderately Gricean? An empirical response to Engelhardt et al.
  (2006)}}.{\BBCQ}
\newblock
\APACjournalVolNumPages{Journal of Pragmatics}{49}{1}{78--106}.
\newblock
\begin{APACrefDOI} \doi{10.1016/j.pragma.2013.01.004} \end{APACrefDOI}
\PrintBackRefs{\CurrentBib}

\bibitem [\protect \citeauthoryear {%
Degen%
, Franke%
\BCBL {}\ \BBA {} J\"{a}ger%
}{%
Degen%
\ \protect \BOthers {.}}{%
{\protect \APACyear {2013}}%
}]{%
degenfrankejaeger2013}
\APACinsertmetastar {%
degenfrankejaeger2013}%
\begin{APACrefauthors}%
Degen, J.%
, Franke, M.%
\BCBL {}\ \BBA {} J\"{a}ger, G.%
\end{APACrefauthors}%
\unskip\
\newblock
\APACrefYearMonthDay{2013}{}{}.
\newblock
{\BBOQ}\APACrefatitle {{Cost-Based Pragmatic Inference about Referential
  Expressions}} {{Cost-Based Pragmatic Inference about Referential
  Expressions}}.{\BBCQ}
\newblock
\BIn{} \APACrefbtitle {Proceedings of the 35th Annual Conference of the
  Cognitive Science Society.} {Proceedings of the 35th annual conference of the
  cognitive science society.}
\PrintBackRefs{\CurrentBib}

\bibitem [\protect \citeauthoryear {%
Devlin%
, Chang%
, Lee%
\BCBL {}\ \BBA {} Toutanova%
}{%
Devlin%
\ \protect \BOthers {.}}{%
{\protect \APACyear {2018}}%
}]{%
devlin2018bert}
\APACinsertmetastar {%
devlin2018bert}%
\begin{APACrefauthors}%
Devlin, J.%
, Chang, M\BHBI W.%
, Lee, K.%
\BCBL {}\ \BBA {} Toutanova, K.%
\end{APACrefauthors}%
\unskip\
\newblock
\APACrefYearMonthDay{2018}{}{}.
\newblock
{\BBOQ}\APACrefatitle {Bert: Pre-training of deep bidirectional transformers
  for language understanding} {Bert: Pre-training of deep bidirectional
  transformers for language understanding}.{\BBCQ}
\newblock
\APACjournalVolNumPages{arXiv preprint arXiv:1810.04805}{}{}{}.
\PrintBackRefs{\CurrentBib}

\bibitem [\protect \citeauthoryear {%
Emberson%
, Loncar%
, Mazzei%
, Treves%
\BCBL {}\ \BBA {} Goldberg%
}{%
Emberson%
\ \protect \BOthers {.}}{%
{\protect \APACyear {2019}}%
}]{%
emberson2019blowfish}
\APACinsertmetastar {%
emberson2019blowfish}%
\begin{APACrefauthors}%
Emberson, L\BPBI L.%
, Loncar, N.%
, Mazzei, C.%
, Treves, I.%
\BCBL {}\ \BBA {} Goldberg, A\BPBI E.%
\end{APACrefauthors}%
\unskip\
\newblock
\APACrefYearMonthDay{2019}{}{}.
\newblock
{\BBOQ}\APACrefatitle {The blowfish effect: children and adults use atypical
  exemplars to infer more narrow categories during word learning} {The blowfish
  effect: children and adults use atypical exemplars to infer more narrow
  categories during word learning}.{\BBCQ}
\newblock
\APACjournalVolNumPages{Journal of child language}{46}{5}{938--954}.
\PrintBackRefs{\CurrentBib}

\bibitem [\protect \citeauthoryear {%
Engelhardt%
, Bailey%
\BCBL {}\ \BBA {} Ferreira%
}{%
Engelhardt%
\ \protect \BOthers {.}}{%
{\protect \APACyear {2006}}%
}]{%
Engelhardt2006}
\APACinsertmetastar {%
Engelhardt2006}%
\begin{APACrefauthors}%
Engelhardt, P\BPBI E.%
, Bailey, K.%
\BCBL {}\ \BBA {} Ferreira, F.%
\end{APACrefauthors}%
\unskip\
\newblock
\APACrefYearMonthDay{2006}{}{}.
\newblock
{\BBOQ}\APACrefatitle {{Do speakers and listeners observe the Gricean Maxim of
  Quantity?}} {{Do speakers and listeners observe the Gricean Maxim of
  Quantity?}}{\BBCQ}
\newblock
\APACjournalVolNumPages{Journal of Memory and Language}{54}{4}{554--573}.
\newblock
\begin{APACrefDOI} \doi{10.1016/j.jml.2005.12.009} \end{APACrefDOI}
\PrintBackRefs{\CurrentBib}

\bibitem [\protect \citeauthoryear {%
Engelhardt%
, Demiral%
\BCBL {}\ \BBA {} Ferreira%
}{%
Engelhardt%
\ \protect \BOthers {.}}{%
{\protect \APACyear {2011}}%
}]{%
Engelhardt2011}
\APACinsertmetastar {%
Engelhardt2011}%
\begin{APACrefauthors}%
Engelhardt, P\BPBI E.%
, Demiral, S\BPBI B.%
\BCBL {}\ \BBA {} Ferreira, F.%
\end{APACrefauthors}%
\unskip\
\newblock
\APACrefYearMonthDay{2011}{}{}.
\newblock
{\BBOQ}\APACrefatitle {{Over-specified referring expressions impair
  comprehension: An ERP study}} {{Over-specified referring expressions impair
  comprehension: An ERP study}}.{\BBCQ}
\newblock
\APACjournalVolNumPages{Brain and Cognition}{77}{2}{304--314}.
\newblock
\begin{APACrefDOI} \doi{10.1016/j.bandc.2011.07.004} \end{APACrefDOI}
\PrintBackRefs{\CurrentBib}

\bibitem [\protect \citeauthoryear {%
A.~Frank%
\ \BBA {} Jaeger%
}{%
A.~Frank%
\ \BBA {} Jaeger%
}{%
{\protect \APACyear {2008}}%
}]{%
frank2008}
\APACinsertmetastar {%
frank2008}%
\begin{APACrefauthors}%
Frank, A.%
\BCBT {}\ \BBA {} Jaeger, T\BPBI F.%
\end{APACrefauthors}%
\unskip\
\newblock
\APACrefYearMonthDay{2008}{}{}.
\newblock
{\BBOQ}\APACrefatitle {{Speaking rationally: Uniform information density as an
  optimal strategy for language production}} {{Speaking rationally: Uniform
  information density as an optimal strategy for language production}}.{\BBCQ}
\newblock
\BIn{} \APACrefbtitle {The 30th Annual Meeting of the Cognitive Science
  Society.} {The 30th annual meeting of the cognitive science society.}
\PrintBackRefs{\CurrentBib}

\bibitem [\protect \citeauthoryear {%
M\BPBI C.~Frank%
\ \BBA {} Goodman%
}{%
M\BPBI C.~Frank%
\ \BBA {} Goodman%
}{%
{\protect \APACyear {2012}}%
}]{%
frank2012}
\APACinsertmetastar {%
frank2012}%
\begin{APACrefauthors}%
Frank, M\BPBI C.%
\BCBT {}\ \BBA {} Goodman, N\BPBI D.%
\end{APACrefauthors}%
\unskip\
\newblock
\APACrefYearMonthDay{2012}{}{}.
\newblock
{\BBOQ}\APACrefatitle {{Predicting pragmatic reasoning in language games}}
  {{Predicting pragmatic reasoning in language games}}.{\BBCQ}
\newblock
\APACjournalVolNumPages{Science}{336}{}{998}.
\PrintBackRefs{\CurrentBib}

\bibitem [\protect \citeauthoryear {%
Franke%
\ \BBA {} Degen%
}{%
Franke%
\ \BBA {} Degen%
}{%
{\protect \APACyear {2016}}%
}]{%
FrankeDegen2016}
\APACinsertmetastar {%
FrankeDegen2016}%
\begin{APACrefauthors}%
Franke, M.%
\BCBT {}\ \BBA {} Degen, J.%
\end{APACrefauthors}%
\unskip\
\newblock
\APACrefYearMonthDay{2016}{}{}.
\newblock
{\BBOQ}\APACrefatitle {{Reasoning in Reference Games : Individual- vs .
  Population-Level Probabilistic Modeling}} {{Reasoning in Reference Games :
  Individual- vs . Population-Level Probabilistic Modeling}}.{\BBCQ}
\newblock
\APACjournalVolNumPages{PLoS ONE}{11}{5}{1--25}.
\newblock
\begin{APACrefDOI} \doi{10.1371/journal.pone.0154854} \end{APACrefDOI}
\PrintBackRefs{\CurrentBib}

\bibitem [\protect \citeauthoryear {%
Franke%
\ \BBA {} J{\"{a}}ger%
}{%
Franke%
\ \BBA {} J{\"{a}}ger%
}{%
{\protect \APACyear {2016}}%
}]{%
FrankeJaeger2016}
\APACinsertmetastar {%
FrankeJaeger2016}%
\begin{APACrefauthors}%
Franke, M.%
\BCBT {}\ \BBA {} J{\"{a}}ger, G.%
\end{APACrefauthors}%
\unskip\
\newblock
\APACrefYearMonthDay{2016}{}{}.
\newblock
{\BBOQ}\APACrefatitle {{Probabilistic pragmatics, or why Bayes' rule is
  probably important for pragmatics}} {{Probabilistic pragmatics, or why Bayes'
  rule is probably important for pragmatics}}.{\BBCQ}
\newblock
\APACjournalVolNumPages{Zeitschrift fur Sprachwissenschaft}{35}{1}{3--44}.
\newblock
\begin{APACrefDOI} \doi{10.1515/zfs-2016-0002} \end{APACrefDOI}
\PrintBackRefs{\CurrentBib}

\bibitem [\protect \citeauthoryear {%
Fukumura%
}{%
Fukumura%
}{%
{\protect \APACyear {2018}}%
}]{%
fukumura2018}
\APACinsertmetastar {%
fukumura2018}%
\begin{APACrefauthors}%
Fukumura, K.%
\end{APACrefauthors}%
\unskip\
\newblock
\APACrefYearMonthDay{2018}{}{}.
\newblock
{\BBOQ}\APACrefatitle {{Ordering adjectives in referential communication}}
  {{Ordering adjectives in referential communication}}.{\BBCQ}
\newblock
\APACjournalVolNumPages{Journal of Memory and Language}{101}{March}{37--50}.
\newblock
\begin{APACrefURL} \url{https://doi.org/10.1016/j.jml.2018.03.003}
  \end{APACrefURL}
\newblock
\begin{APACrefDOI} \doi{10.1016/j.jml.2018.03.003} \end{APACrefDOI}
\PrintBackRefs{\CurrentBib}

\bibitem [\protect \citeauthoryear {%
Gatt%
, Krahmer%
, van Deemter%
\BCBL {}\ \BBA {} van Gompel%
}{%
Gatt%
\ \protect \BOthers {.}}{%
{\protect \APACyear {2014}}%
}]{%
Gatt2014}
\APACinsertmetastar {%
Gatt2014}%
\begin{APACrefauthors}%
Gatt, A.%
, Krahmer, E.%
, van Deemter, K.%
\BCBL {}\ \BBA {} van Gompel, R\BPBI P.%
\end{APACrefauthors}%
\unskip\
\newblock
\APACrefYearMonthDay{2014}{}{}.
\newblock
{\BBOQ}\APACrefatitle {{Models and empirical data for the production of
  referring expressions}} {{Models and empirical data for the production of
  referring expressions}}.{\BBCQ}
\newblock
\APACjournalVolNumPages{Language, Cognition and Neuroscience}{29}{8}{899--911}.
\PrintBackRefs{\CurrentBib}

\bibitem [\protect \citeauthoryear {%
Gatt%
, Krahmer%
, Van~Deemter%
\BCBL {}\ \BBA {} van Gompel%
}{%
Gatt%
\ \protect \BOthers {.}}{%
{\protect \APACyear {2017}}%
}]{%
gatt2017}
\APACinsertmetastar {%
gatt2017}%
\begin{APACrefauthors}%
Gatt, A.%
, Krahmer, E.%
, Van~Deemter, K.%
\BCBL {}\ \BBA {} van Gompel, R\BPBI P.%
\end{APACrefauthors}%
\unskip\
\newblock
\APACrefYearMonthDay{2017}{}{}.
\newblock
{\BBOQ}\APACrefatitle {Reference production as search: The impact of domain
  size on the production of distinguishing descriptions} {Reference production
  as search: The impact of domain size on the production of distinguishing
  descriptions}.{\BBCQ}
\newblock
\APACjournalVolNumPages{Cognitive science}{41}{}{1457--1492}.
\PrintBackRefs{\CurrentBib}

\bibitem [\protect \citeauthoryear {%
Gatt%
, van Gompel%
, Krahmer%
\BCBL {}\ \BBA {} van Deemter%
}{%
Gatt%
\ \protect \BOthers {.}}{%
{\protect \APACyear {2011}}%
}]{%
gatt2011}
\APACinsertmetastar {%
gatt2011}%
\begin{APACrefauthors}%
Gatt, A.%
, van Gompel, R\BPBI P\BPBI G.%
, Krahmer, E.%
\BCBL {}\ \BBA {} van Deemter, K.%
\end{APACrefauthors}%
\unskip\
\newblock
\APACrefYearMonthDay{2011}{}{}.
\newblock
{\BBOQ}\APACrefatitle {{Non-deterministic attribute selection in reference
  production}} {{Non-deterministic attribute selection in reference
  production}}.{\BBCQ}
\newblock
\BIn{} \APACrefbtitle {Proceedings of the workshop on production of referring
  expressions: Bridging the gap between empirical, computational and
  psycholinguistic approaches to reference (pre-cogsci’11).} {Proceedings of
  the workshop on production of referring expressions: Bridging the gap between
  empirical, computational and psycholinguistic approaches to reference
  (pre-cogsci’11).}
\newblock
\APACaddressPublisher{Boston}{}.
\PrintBackRefs{\CurrentBib}

\bibitem [\protect \citeauthoryear {%
Goodman%
\ \BBA {} Frank%
}{%
Goodman%
\ \BBA {} Frank%
}{%
{\protect \APACyear {2016}}%
}]{%
goodman2016}
\APACinsertmetastar {%
goodman2016}%
\begin{APACrefauthors}%
Goodman, N\BPBI D.%
\BCBT {}\ \BBA {} Frank, M\BPBI C.%
\end{APACrefauthors}%
\unskip\
\newblock
\APACrefYearMonthDay{2016}{}{}.
\newblock
{\BBOQ}\APACrefatitle {{Pragmatic Language Interpretation as Probabilistic
  Inference}} {{Pragmatic Language Interpretation as Probabilistic
  Inference}}.{\BBCQ}
\newblock
\APACjournalVolNumPages{Trends in Cognitive Sciences}{20}{11}{818--829}.
\newblock
\begin{APACrefDOI} \doi{10.1016/j.tics.2016.08.005} \end{APACrefDOI}
\PrintBackRefs{\CurrentBib}

\bibitem [\protect \citeauthoryear {%
Goodman%
\ \BBA {} Stuhlm\"{u}ller%
}{%
Goodman%
\ \BBA {} Stuhlm\"{u}ller%
}{%
{\protect \APACyear {2013}}%
}]{%
goodmanstuhlmueller2013}
\APACinsertmetastar {%
goodmanstuhlmueller2013}%
\begin{APACrefauthors}%
Goodman, N\BPBI D.%
\BCBT {}\ \BBA {} Stuhlm\"{u}ller, A.%
\end{APACrefauthors}%
\unskip\
\newblock
\APACrefYearMonthDay{2013}{}{}.
\newblock
{\BBOQ}\APACrefatitle {{Knowledge and implicature: modeling language
  understanding as social cognition.}} {{Knowledge and implicature: modeling
  language understanding as social cognition.}}{\BBCQ}
\newblock
\APACjournalVolNumPages{Topics in Cognitive Science}{5}{1}{173--84}.
\PrintBackRefs{\CurrentBib}

\bibitem [\protect \citeauthoryear {%
Goodman%
\ \BBA {} Stuhlm{\"u}ller%
}{%
Goodman%
\ \BBA {} Stuhlm{\"u}ller%
}{%
{\protect \APACyear {electronic}}%
}]{%
GoodmanStuhlmuller14_DIPPL}
\APACinsertmetastar {%
GoodmanStuhlmuller14_DIPPL}%
\begin{APACrefauthors}%
Goodman, N\BPBI D.%
\BCBT {}\ \BBA {} Stuhlm{\"u}ller, A.%
\end{APACrefauthors}%
\unskip\
\newblock
\APACrefYearMonthDay{electronic}{}{}.
\newblock
\APACrefbtitle {The Design and Implementation of Probabilistic Programming
  Languages.} {The design and implementation of probabilistic programming
  languages.}
\newblock
\begin{APACrefURL} [{2015/1/16}]\url{http://dippl.org} \end{APACrefURL}
\PrintBackRefs{\CurrentBib}

\bibitem [\protect \citeauthoryear {%
Graf%
, Degen%
, Hawkins%
\BCBL {}\ \BBA {} Goodman%
}{%
Graf%
\ \protect \BOthers {.}}{%
{\protect \APACyear {2016}}%
}]{%
GrafEtAl2016}
\APACinsertmetastar {%
GrafEtAl2016}%
\begin{APACrefauthors}%
Graf, C.%
, Degen, J.%
, Hawkins, R\BPBI X\BPBI D.%
\BCBL {}\ \BBA {} Goodman, N\BPBI D.%
\end{APACrefauthors}%
\unskip\
\newblock
\APACrefYearMonthDay{2016}{}{}.
\newblock
{\BBOQ}\APACrefatitle {{Animal, dog, or dalmatian? Level of abstraction in
  nominal referring expressions}} {{Animal, dog, or dalmatian? Level of
  abstraction in nominal referring expressions}}.{\BBCQ}
\newblock
\BIn{} A.~Papafragou, D.~Grodner, D.~Mirman\BCBL {}\ \BBA {} J.~Trueswell\
  (\BEDS), \APACrefbtitle {Proceedings of the 38th Annual Conference of the
  Cognitive Science Society} {Proceedings of the 38th annual conference of the
  cognitive science society}\ (\BPGS\ 2261--2266).
\newblock
\APACaddressPublisher{Austin, TX}{Cognitive Science Society}.
\PrintBackRefs{\CurrentBib}

\bibitem [\protect \citeauthoryear {%
Grice%
}{%
Grice%
}{%
{\protect \APACyear {1975}}%
}]{%
grice1975}
\APACinsertmetastar {%
grice1975}%
\begin{APACrefauthors}%
Grice, H\BPBI P.%
\end{APACrefauthors}%
\unskip\
\newblock
\APACrefYearMonthDay{1975}{}{}.
\newblock
{\BBOQ}\APACrefatitle {{Logic and Conversation}} {{Logic and
  Conversation}}.{\BBCQ}
\newblock
\APACjournalVolNumPages{Syntax and Semantics}{3}{}{41--58}.
\PrintBackRefs{\CurrentBib}

\bibitem [\protect \citeauthoryear {%
Griffin%
\ \BBA {} Bock%
}{%
Griffin%
\ \BBA {} Bock%
}{%
{\protect \APACyear {1998}}%
}]{%
griffin1998}
\APACinsertmetastar {%
griffin1998}%
\begin{APACrefauthors}%
Griffin, Z\BPBI M.%
\BCBT {}\ \BBA {} Bock, K.%
\end{APACrefauthors}%
\unskip\
\newblock
\APACrefYearMonthDay{1998}{}{}.
\newblock
{\BBOQ}\APACrefatitle {{Constraint, word frequency, and levels of processing in
  spoken word production}} {{Constraint, word frequency, and levels of
  processing in spoken word production}}.{\BBCQ}
\newblock
\APACjournalVolNumPages{Journal of Memory and Language}{38}{38}{313--338}.
\PrintBackRefs{\CurrentBib}

\bibitem [\protect \citeauthoryear {%
Hahn%
, Degen%
, Goodman%
, Jurafsky%
\BCBL {}\ \BBA {} Futrell%
}{%
Hahn%
\ \protect \BOthers {.}}{%
{\protect \APACyear {2018}}%
}]{%
Hahn2018}
\APACinsertmetastar {%
Hahn2018}%
\begin{APACrefauthors}%
Hahn, M.%
, Degen, J.%
, Goodman, N.%
, Jurafsky, D.%
\BCBL {}\ \BBA {} Futrell, R.%
\end{APACrefauthors}%
\unskip\
\newblock
\APACrefYearMonthDay{2018}{}{}.
\newblock
{\BBOQ}\APACrefatitle {{An Information-Theoretic Explanation of Adjective
  Ordering Preferences}} {{An Information-Theoretic Explanation of Adjective
  Ordering Preferences}}.{\BBCQ}
\newblock
\BIn{} \APACrefbtitle {Proceedings of the 40th Annual Conference of the
  Cognitive Science Society.} {Proceedings of the 40th annual conference of the
  cognitive science society.}
\PrintBackRefs{\CurrentBib}

\bibitem [\protect \citeauthoryear {%
Hale%
}{%
Hale%
}{%
{\protect \APACyear {2001}}%
}]{%
Hale2001}
\APACinsertmetastar {%
Hale2001}%
\begin{APACrefauthors}%
Hale, J.%
\end{APACrefauthors}%
\unskip\
\newblock
\APACrefYearMonthDay{2001}{}{}.
\newblock
{\BBOQ}\APACrefatitle {{A Probabilistic Earley Parser as a Psycholinguistic
  Model}} {{A Probabilistic Earley Parser as a Psycholinguistic Model}}.{\BBCQ}
\newblock
\BIn{} \APACrefbtitle {Proceedings of the Second Meeting of the North American
  Chapter of the Asssociation for Computational Linguistics} {Proceedings of
  the second meeting of the north american chapter of the asssociation for
  computational linguistics}\ (\BPGS\ 1--8).
\newblock
\APACaddressPublisher{}{Association for Computational Linguistics}.
\PrintBackRefs{\CurrentBib}

\bibitem [\protect \citeauthoryear {%
Hawkins%
}{%
Hawkins%
}{%
{\protect \APACyear {2015}}%
}]{%
Hawkins15_RealTimeWebExperiments}
\APACinsertmetastar {%
Hawkins15_RealTimeWebExperiments}%
\begin{APACrefauthors}%
Hawkins, R\BPBI X\BPBI D.%
\end{APACrefauthors}%
\unskip\
\newblock
\APACrefYearMonthDay{2015}{}{}.
\newblock
{\BBOQ}\APACrefatitle {Conducting real-time multiplayer experiments on the web}
  {Conducting real-time multiplayer experiments on the web}.{\BBCQ}
\newblock
\APACjournalVolNumPages{Behavior Research Methods}{47}{4}{966-976}.
\PrintBackRefs{\CurrentBib}

\bibitem [\protect \citeauthoryear {%
Hawkins%
, Stuhlm\"uller%
, Degen%
\BCBL {}\ \BBA {} Goodman%
}{%
Hawkins%
\ \protect \BOthers {.}}{%
{\protect \APACyear {2015}}%
}]{%
Hawkins2015}
\APACinsertmetastar {%
Hawkins2015}%
\begin{APACrefauthors}%
Hawkins, R\BPBI X\BPBI D.%
, Stuhlm\"uller, A.%
, Degen, J.%
\BCBL {}\ \BBA {} Goodman, N\BPBI D.%
\end{APACrefauthors}%
\unskip\
\newblock
\APACrefYearMonthDay{2015}{}{}.
\newblock
{\BBOQ}\APACrefatitle {{Why do you ask ? Good questions provoke informative
  answers .}} {{Why do you ask ? Good questions provoke informative answers
  .}}{\BBCQ}
\newblock
\BIn{} \APACrefbtitle {Proceedings of the 37th Annual Conference of the
  Cognitive Science Society.} {Proceedings of the 37th annual conference of the
  cognitive science society.}
\PrintBackRefs{\CurrentBib}

\bibitem [\protect \citeauthoryear {%
Herrmann%
\ \BBA {} Deutsch%
}{%
Herrmann%
\ \BBA {} Deutsch%
}{%
{\protect \APACyear {1976}}%
}]{%
herrmann1976}
\APACinsertmetastar {%
herrmann1976}%
\begin{APACrefauthors}%
Herrmann, T.%
\BCBT {}\ \BBA {} Deutsch, W.%
\end{APACrefauthors}%
\unskip\
\newblock
\APACrefYear{1976}.
\newblock
\APACrefbtitle {{Psychologie der Objektbenennung}} {{Psychologie der
  Objektbenennung}}.
\newblock
\APACaddressPublisher{}{Huber}.
\PrintBackRefs{\CurrentBib}

\bibitem [\protect \citeauthoryear {%
Hoffmann%
\ \BBA {} Ziessler%
}{%
Hoffmann%
\ \BBA {} Ziessler%
}{%
{\protect \APACyear {1983}}%
}]{%
hoffmann1983objektidentifikation}
\APACinsertmetastar {%
hoffmann1983objektidentifikation}%
\begin{APACrefauthors}%
Hoffmann, J.%
\BCBT {}\ \BBA {} Ziessler, C.%
\end{APACrefauthors}%
\unskip\
\newblock
\APACrefYearMonthDay{1983}{}{}.
\newblock
{\BBOQ}\APACrefatitle {Objektidentifikation in k{\"u}nstlichen
  Begriffshierarchien.} {Objektidentifikation in k{\"u}nstlichen
  begriffshierarchien.}{\BBCQ}
\newblock
\APACjournalVolNumPages{Zeitschrift f{\"u}r Psychologie mit Zeitschrift f{\"u}r
  angewandte Psychologie}{}{}{}.
\PrintBackRefs{\CurrentBib}

\bibitem [\protect \citeauthoryear {%
Horton%
\ \BBA {} Keysar%
}{%
Horton%
\ \BBA {} Keysar%
}{%
{\protect \APACyear {1996}}%
}]{%
horton1996}
\APACinsertmetastar {%
horton1996}%
\begin{APACrefauthors}%
Horton, W.%
\BCBT {}\ \BBA {} Keysar, B.%
\end{APACrefauthors}%
\unskip\
\newblock
\APACrefYearMonthDay{1996}{}{}.
\newblock
{\BBOQ}\APACrefatitle {{When do speakers take into account common ground?}}
  {{When do speakers take into account common ground?}}{\BBCQ}
\newblock
\APACjournalVolNumPages{Cognition}{59}{}{91--117}.
\PrintBackRefs{\CurrentBib}

\bibitem [\protect \citeauthoryear {%
Huettig%
\ \BBA {} Altmann%
}{%
Huettig%
\ \BBA {} Altmann%
}{%
{\protect \APACyear {2011}}%
}]{%
Huettig2011}
\APACinsertmetastar {%
Huettig2011}%
\begin{APACrefauthors}%
Huettig, F.%
\BCBT {}\ \BBA {} Altmann, G\BPBI T\BPBI M.%
\end{APACrefauthors}%
\unskip\
\newblock
\APACrefYearMonthDay{2011}{}{}.
\newblock
{\BBOQ}\APACrefatitle {{Looking at anything that is green when hearing "frog":
  how object surface colour and stored object colour knowledge influence
  language-mediated overt attention.}} {{Looking at anything that is green when
  hearing "frog": how object surface colour and stored object colour knowledge
  influence language-mediated overt attention.}}{\BBCQ}
\newblock
\APACjournalVolNumPages{Quarterly journal of experimental psychology
  (2006)}{64}{1}{122--145}.
\newblock
\begin{APACrefDOI} \doi{10.1080/17470218.2010.481474} \end{APACrefDOI}
\PrintBackRefs{\CurrentBib}

\bibitem [\protect \citeauthoryear {%
Jaeger%
}{%
Jaeger%
}{%
{\protect \APACyear {2006}}%
}]{%
jaeger2006}
\APACinsertmetastar {%
jaeger2006}%
\begin{APACrefauthors}%
Jaeger, T\BPBI F.%
\end{APACrefauthors}%
\unskip\
\newblock
\APACrefYear{2006}.
\unskip\
\newblock
\APACrefbtitle {{Redundancy and Reduction in Spontaneous Speech}} {{Redundancy
  and Reduction in Spontaneous Speech}}\ \APACtypeAddressSchool {\BUPhD}{}{}.
\unskip\
\newblock
\APACaddressSchool {}{Stanford University}.
\PrintBackRefs{\CurrentBib}

\bibitem [\protect \citeauthoryear {%
Jaeger%
}{%
Jaeger%
}{%
{\protect \APACyear {2010}}%
}]{%
jaeger2010}
\APACinsertmetastar {%
jaeger2010}%
\begin{APACrefauthors}%
Jaeger, T\BPBI F.%
\end{APACrefauthors}%
\unskip\
\newblock
\APACrefYearMonthDay{2010}{}{}.
\newblock
{\BBOQ}\APACrefatitle {{Redundancy and reduction: speakers manage syntactic
  information density}} {{Redundancy and reduction: speakers manage syntactic
  information density}}.{\BBCQ}
\newblock
\APACjournalVolNumPages{Cognitive Psychology}{61}{1}{23--62}.
\PrintBackRefs{\CurrentBib}

\bibitem [\protect \citeauthoryear {%
Jescheniak%
\ \BBA {} Levelt%
}{%
Jescheniak%
\ \BBA {} Levelt%
}{%
{\protect \APACyear {1994}}%
}]{%
jescheniak1994}
\APACinsertmetastar {%
jescheniak1994}%
\begin{APACrefauthors}%
Jescheniak, J\BPBI D.%
\BCBT {}\ \BBA {} Levelt, W\BPBI J\BPBI M.%
\end{APACrefauthors}%
\unskip\
\newblock
\APACrefYearMonthDay{1994}{}{}.
\newblock
{\BBOQ}\APACrefatitle {{Word frequency effects in speech production: Retrieval
  of syntactic information and of phonological form.}} {{Word frequency effects
  in speech production: Retrieval of syntactic information and of phonological
  form.}}{\BBCQ}
\newblock
\APACjournalVolNumPages{Journal of Experimental Psychology: Learning, Memory,
  and Cognition}{20}{4}{824--843}.
\newblock
\begin{APACrefDOI} \doi{10.1037//0278-7393.20.4.824} \end{APACrefDOI}
\PrintBackRefs{\CurrentBib}

\bibitem [\protect \citeauthoryear {%
Johnson%
\ \BBA {} Mervis%
}{%
Johnson%
\ \BBA {} Mervis%
}{%
{\protect \APACyear {1997}}%
}]{%
Johnson1997}
\APACinsertmetastar {%
Johnson1997}%
\begin{APACrefauthors}%
Johnson, K\BPBI E.%
\BCBT {}\ \BBA {} Mervis, C\BPBI B.%
\end{APACrefauthors}%
\unskip\
\newblock
\APACrefYearMonthDay{1997}{}{}.
\newblock
{\BBOQ}\APACrefatitle {{Effects of varying levels of expertise on the basic
  level of categorization}} {{Effects of varying levels of expertise on the
  basic level of categorization}}.{\BBCQ}
\newblock
\APACjournalVolNumPages{Journal of Experimental Psychology:
  General}{126}{3}{248--277}.
\newblock
\begin{APACrefDOI} \doi{10.1037/0096-3445.126.3.248} \end{APACrefDOI}
\PrintBackRefs{\CurrentBib}

\bibitem [\protect \citeauthoryear {%
Jolicoeur%
, Gluck%
\BCBL {}\ \BBA {} Kosslyn%
}{%
Jolicoeur%
\ \protect \BOthers {.}}{%
{\protect \APACyear {1984}}%
}]{%
Jolicoeur1984}
\APACinsertmetastar {%
Jolicoeur1984}%
\begin{APACrefauthors}%
Jolicoeur, P.%
, Gluck, M\BPBI A.%
\BCBL {}\ \BBA {} Kosslyn, S\BPBI M.%
\end{APACrefauthors}%
\unskip\
\newblock
\APACrefYearMonthDay{1984}{}{}.
\newblock
{\BBOQ}\APACrefatitle {{Pictures and names: Making the connection}} {{Pictures
  and names: Making the connection}}.{\BBCQ}
\newblock
\APACjournalVolNumPages{Cognitive Psychology}{16}{2}{243--275}.
\PrintBackRefs{\CurrentBib}

\bibitem [\protect \citeauthoryear {%
Kamp%
\ \BBA {} Partee%
}{%
Kamp%
\ \BBA {} Partee%
}{%
{\protect \APACyear {1995}}%
}]{%
kamp1995}
\APACinsertmetastar {%
kamp1995}%
\begin{APACrefauthors}%
Kamp, H.%
\BCBT {}\ \BBA {} Partee, B.%
\end{APACrefauthors}%
\unskip\
\newblock
\APACrefYearMonthDay{1995}{}{}.
\newblock
{\BBOQ}\APACrefatitle {{Prototype theory and compositionality.}} {{Prototype
  theory and compositionality.}}{\BBCQ}
\newblock
\APACjournalVolNumPages{Cognition}{57}{2}{129--91}.
\PrintBackRefs{\CurrentBib}

\bibitem [\protect \citeauthoryear {%
Kao%
, Wu%
, Bergen%
\BCBL {}\ \BBA {} Goodman%
}{%
Kao%
\ \protect \BOthers {.}}{%
{\protect \APACyear {2014}}%
}]{%
kao2014}
\APACinsertmetastar {%
kao2014}%
\begin{APACrefauthors}%
Kao, J.%
, Wu, J.%
, Bergen, L.%
\BCBL {}\ \BBA {} Goodman, N\BPBI D.%
\end{APACrefauthors}%
\unskip\
\newblock
\APACrefYearMonthDay{2014}{}{}.
\newblock
{\BBOQ}\APACrefatitle {{Nonliteral understanding of number words}} {{Nonliteral
  understanding of number words}}.{\BBCQ}
\newblock
\APACjournalVolNumPages{Proceedings of the National Academy of Sciences of the
  United States of America}{111}{33}{12002--12007}.
\PrintBackRefs{\CurrentBib}

\bibitem [\protect \citeauthoryear {%
Kennedy%
\ \BBA {} McNally%
}{%
Kennedy%
\ \BBA {} McNally%
}{%
{\protect \APACyear {2005}}%
}]{%
kennedy2005}
\APACinsertmetastar {%
kennedy2005}%
\begin{APACrefauthors}%
Kennedy, C.%
\BCBT {}\ \BBA {} McNally, L.%
\end{APACrefauthors}%
\unskip\
\newblock
\APACrefYearMonthDay{2005}{}{}.
\newblock
{\BBOQ}\APACrefatitle {{Scale Structure, Degree Modification, and the Semantics
  of Gradable Predicates}} {{Scale Structure, Degree Modification, and the
  Semantics of Gradable Predicates}}.{\BBCQ}
\newblock
\APACjournalVolNumPages{Language}{81}{2}{345--381}.
\newblock
\begin{APACrefDOI} \doi{10.1353/lan.2005.0071} \end{APACrefDOI}
\PrintBackRefs{\CurrentBib}

\bibitem [\protect \citeauthoryear {%
Kennedy%
\ \BBA {} Mcnally%
}{%
Kennedy%
\ \BBA {} Mcnally%
}{%
{\protect \APACyear {2010}}%
}]{%
Kennedy2010}
\APACinsertmetastar {%
Kennedy2010}%
\begin{APACrefauthors}%
Kennedy, C.%
\BCBT {}\ \BBA {} Mcnally, L.%
\end{APACrefauthors}%
\unskip\
\newblock
\APACrefYearMonthDay{2010}{}{}.
\newblock
{\BBOQ}\APACrefatitle {{Color, context, and compositionality}} {{Color,
  context, and compositionality}}.{\BBCQ}
\newblock
\APACjournalVolNumPages{Synthese}{174}{1}{79--98}.
\PrintBackRefs{\CurrentBib}

\bibitem [\protect \citeauthoryear {%
Koolen%
, Gatt%
, Goudbeek%
\BCBL {}\ \BBA {} Krahmer%
}{%
Koolen%
\ \protect \BOthers {.}}{%
{\protect \APACyear {2011}}%
}]{%
Koolen2011}
\APACinsertmetastar {%
Koolen2011}%
\begin{APACrefauthors}%
Koolen, R.%
, Gatt, A.%
, Goudbeek, M.%
\BCBL {}\ \BBA {} Krahmer, E.%
\end{APACrefauthors}%
\unskip\
\newblock
\APACrefYearMonthDay{2011}{}{}.
\newblock
{\BBOQ}\APACrefatitle {{Factors causing overspecification in definite
  descriptions}} {{Factors causing overspecification in definite
  descriptions}}.{\BBCQ}
\newblock
\APACjournalVolNumPages{Journal of Pragmatics}{43}{13}{3231--3250}.
\PrintBackRefs{\CurrentBib}

\bibitem [\protect \citeauthoryear {%
Koolen%
, Goudbeek%
\BCBL {}\ \BBA {} Krahmer%
}{%
Koolen%
\ \protect \BOthers {.}}{%
{\protect \APACyear {2013}}%
}]{%
Koolen2013}
\APACinsertmetastar {%
Koolen2013}%
\begin{APACrefauthors}%
Koolen, R.%
, Goudbeek, M.%
\BCBL {}\ \BBA {} Krahmer, E.%
\end{APACrefauthors}%
\unskip\
\newblock
\APACrefYearMonthDay{2013}{}{}.
\newblock
{\BBOQ}\APACrefatitle {{The effect of scene variation on the redundant use of
  color in definite reference}} {{The effect of scene variation on the
  redundant use of color in definite reference}}.{\BBCQ}
\newblock
\APACjournalVolNumPages{Cognitive Science}{37}{2}{395--411}.
\newblock
\begin{APACrefDOI} \doi{10.1111/cogs.12019} \end{APACrefDOI}
\PrintBackRefs{\CurrentBib}

\bibitem [\protect \citeauthoryear {%
Krahmer%
, van Erk%
\BCBL {}\ \BBA {} Verleg%
}{%
Krahmer%
\ \protect \BOthers {.}}{%
{\protect \APACyear {2003}}%
}]{%
Krahmer2003}
\APACinsertmetastar {%
Krahmer2003}%
\begin{APACrefauthors}%
Krahmer, E.%
, van Erk, S.%
\BCBL {}\ \BBA {} Verleg, A.%
\end{APACrefauthors}%
\unskip\
\newblock
\APACrefYearMonthDay{2003}{}{}.
\newblock
{\BBOQ}\APACrefatitle {{Graph-based generation of referring expressions}}
  {{Graph-based generation of referring expressions}}.{\BBCQ}
\newblock
\APACjournalVolNumPages{Computational Linguistics}{29}{1}{53--72}.
\PrintBackRefs{\CurrentBib}

\bibitem [\protect \citeauthoryear {%
Levinson%
}{%
Levinson%
}{%
{\protect \APACyear {1983}}%
}]{%
levinson1983pragmatics}
\APACinsertmetastar {%
levinson1983pragmatics}%
\begin{APACrefauthors}%
Levinson, S\BPBI C.%
\end{APACrefauthors}%
\unskip\
\newblock
\APACrefYear{1983}.
\newblock
\APACrefbtitle {Pragmatics (Cambridge textbooks in linguistics)} {Pragmatics
  (cambridge textbooks in linguistics)}.
\newblock
\APACaddressPublisher{}{Cambridge University Press}.
\PrintBackRefs{\CurrentBib}

\bibitem [\protect \citeauthoryear {%
Levy%
}{%
Levy%
}{%
{\protect \APACyear {2008}}%
}]{%
Levy2008}
\APACinsertmetastar {%
Levy2008}%
\begin{APACrefauthors}%
Levy, R.%
\end{APACrefauthors}%
\unskip\
\newblock
\APACrefYearMonthDay{2008}{}{}.
\newblock
{\BBOQ}\APACrefatitle {{Expectation-based syntactic comprehension.}}
  {{Expectation-based syntactic comprehension.}}{\BBCQ}
\newblock
\APACjournalVolNumPages{Cognition}{106}{3}{1126--77}.
\newblock
\begin{APACrefDOI} \doi{10.1016/j.cognition.2007.05.006} \end{APACrefDOI}
\PrintBackRefs{\CurrentBib}

\bibitem [\protect \citeauthoryear {%
Levy%
\ \BBA {} Jaeger%
}{%
Levy%
\ \BBA {} Jaeger%
}{%
{\protect \APACyear {2007}}%
}]{%
levy2007}
\APACinsertmetastar {%
levy2007}%
\begin{APACrefauthors}%
Levy, R.%
\BCBT {}\ \BBA {} Jaeger, T\BPBI F.%
\end{APACrefauthors}%
\unskip\
\newblock
\APACrefYearMonthDay{2007}{}{}.
\newblock
{\BBOQ}\APACrefatitle {{Speakers optimize information density through syntactic
  reduction}} {{Speakers optimize information density through syntactic
  reduction}}.{\BBCQ}
\newblock
\BIn{} B.~Schl{\"{o}}kopf, J.~Platt\BCBL {}\ \BBA {} T.~Hoffman\ (\BEDS),
  \APACrefbtitle {Advances in Neural Information Processing Systems} {Advances
  in neural information processing systems}\ (\BVOL~19, \BPGS\ 849--856).
\newblock
\APACaddressPublisher{Cambridge, MA}{MIT Press}.
\PrintBackRefs{\CurrentBib}

\bibitem [\protect \citeauthoryear {%
Luce%
}{%
Luce%
}{%
{\protect \APACyear {1959}}%
}]{%
luce1959}
\APACinsertmetastar {%
luce1959}%
\begin{APACrefauthors}%
Luce, R\BPBI D.%
\end{APACrefauthors}%
\unskip\
\newblock
\APACrefYear{1959}.
\newblock
\APACrefbtitle {{Individual Choice Behavior: A Theoretical Analysis}}
  {{Individual Choice Behavior: A Theoretical Analysis}}.
\newblock
\APACaddressPublisher{New York}{Wiley}.
\PrintBackRefs{\CurrentBib}

\bibitem [\protect \citeauthoryear {%
Maes%
, Arts%
\BCBL {}\ \BBA {} Noordman%
}{%
Maes%
\ \protect \BOthers {.}}{%
{\protect \APACyear {2004}}%
}]{%
Maes2004}
\APACinsertmetastar {%
Maes2004}%
\begin{APACrefauthors}%
Maes, A.%
, Arts, A.%
\BCBL {}\ \BBA {} Noordman, L.%
\end{APACrefauthors}%
\unskip\
\newblock
\APACrefYearMonthDay{2004}{}{}.
\newblock
{\BBOQ}\APACrefatitle {{Reference Management in Instructive Discourse}}
  {{Reference Management in Instructive Discourse}}.{\BBCQ}
\newblock
\APACjournalVolNumPages{Discourse Processes: A Multidisciplinary
  Journal}{37}{2}{117--144}.
\PrintBackRefs{\CurrentBib}

\bibitem [\protect \citeauthoryear {%
Marr%
}{%
Marr%
}{%
{\protect \APACyear {1982}}%
}]{%
marr1982}
\APACinsertmetastar {%
marr1982}%
\begin{APACrefauthors}%
Marr, D.%
\end{APACrefauthors}%
\unskip\
\newblock
\APACrefYear{1982}.
\newblock
\APACrefbtitle {{Vision: A Computational Investigation into the Human
  Representation and Processing of Visual Information}} {{Vision: A
  Computational Investigation into the Human Representation and Processing of
  Visual Information}}.
\newblock
\APACaddressPublisher{San Francisco}{W.H. Freeman}.
\PrintBackRefs{\CurrentBib}

\bibitem [\protect \citeauthoryear {%
Mitchell%
}{%
Mitchell%
}{%
{\protect \APACyear {2013}}%
}]{%
Mitchell2013}
\APACinsertmetastar {%
Mitchell2013}%
\begin{APACrefauthors}%
Mitchell, M.%
\end{APACrefauthors}%
\unskip\
\newblock
\APACrefYearMonthDay{2013}{}{}.
\newblock
{\BBOQ}\APACrefatitle {{Typicality and object reference}} {{Typicality and
  object reference}}.{\BBCQ}
\newblock
\APACjournalVolNumPages{Proceedings of the 35th Annual Meeting of the Cognitive
  Science Society}{}{}{3062--3067}.
\PrintBackRefs{\CurrentBib}

\bibitem [\protect \citeauthoryear {%
Murphy%
\ \BBA {} Smith%
}{%
Murphy%
\ \BBA {} Smith%
}{%
{\protect \APACyear {1982}}%
}]{%
murphy1982basic}
\APACinsertmetastar {%
murphy1982basic}%
\begin{APACrefauthors}%
Murphy, G\BPBI L.%
\BCBT {}\ \BBA {} Smith, E\BPBI E.%
\end{APACrefauthors}%
\unskip\
\newblock
\APACrefYearMonthDay{1982}{}{}.
\newblock
{\BBOQ}\APACrefatitle {Basic-level superiority in picture categorization}
  {Basic-level superiority in picture categorization}.{\BBCQ}
\newblock
\APACjournalVolNumPages{Journal of verbal learning and verbal
  behavior}{21}{1}{1--20}.
\PrintBackRefs{\CurrentBib}

\bibitem [\protect \citeauthoryear {%
Nadig%
\ \BBA {} Sedivy%
}{%
Nadig%
\ \BBA {} Sedivy%
}{%
{\protect \APACyear {2002}}%
}]{%
nadig2002}
\APACinsertmetastar {%
nadig2002}%
\begin{APACrefauthors}%
Nadig, A\BPBI S.%
\BCBT {}\ \BBA {} Sedivy, J\BPBI C.%
\end{APACrefauthors}%
\unskip\
\newblock
\APACrefYearMonthDay{2002}{}{}.
\newblock
{\BBOQ}\APACrefatitle {{Evidence of Perspective-Taking Constraints in
  Children's On-Line Reference Resolution}} {{Evidence of Perspective-Taking
  Constraints in Children's On-Line Reference Resolution}}.{\BBCQ}
\newblock
\APACjournalVolNumPages{Psychological Science}{13}{4}{329--336}.
\newblock
\begin{APACrefDOI} \doi{10.1111/j.0956-7976.2002.00460.x} \end{APACrefDOI}
\PrintBackRefs{\CurrentBib}

\bibitem [\protect \citeauthoryear {%
Oldfield%
\ \BBA {} Wingfield%
}{%
Oldfield%
\ \BBA {} Wingfield%
}{%
{\protect \APACyear {1965}}%
}]{%
oldfield1965response}
\APACinsertmetastar {%
oldfield1965response}%
\begin{APACrefauthors}%
Oldfield, R\BPBI C.%
\BCBT {}\ \BBA {} Wingfield, A.%
\end{APACrefauthors}%
\unskip\
\newblock
\APACrefYearMonthDay{1965}{}{}.
\newblock
{\BBOQ}\APACrefatitle {Response latencies in naming objects} {Response
  latencies in naming objects}.{\BBCQ}
\newblock
\APACjournalVolNumPages{Quarterly Journal of Experimental
  Psychology}{17}{4}{273--281}.
\PrintBackRefs{\CurrentBib}

\bibitem [\protect \citeauthoryear {%
Olson%
}{%
Olson%
}{%
{\protect \APACyear {1970}}%
}]{%
olson1970language}
\APACinsertmetastar {%
olson1970language}%
\begin{APACrefauthors}%
Olson, D\BPBI R.%
\end{APACrefauthors}%
\unskip\
\newblock
\APACrefYearMonthDay{1970}{}{}.
\newblock
{\BBOQ}\APACrefatitle {Language and thought: Aspects of a cognitive theory of
  semantics.} {Language and thought: Aspects of a cognitive theory of
  semantics.}{\BBCQ}
\newblock
\APACjournalVolNumPages{Psychological review}{77}{4}{257}.
\PrintBackRefs{\CurrentBib}

\bibitem [\protect \citeauthoryear {%
Osherson%
\ \BBA {} Smith%
}{%
Osherson%
\ \BBA {} Smith%
}{%
{\protect \APACyear {1981}}%
}]{%
Osherson1981}
\APACinsertmetastar {%
Osherson1981}%
\begin{APACrefauthors}%
Osherson, D\BPBI N.%
\BCBT {}\ \BBA {} Smith, E\BPBI E.%
\end{APACrefauthors}%
\unskip\
\newblock
\APACrefYearMonthDay{1981}{}{}.
\newblock
{\BBOQ}\APACrefatitle {{On the adequacy of prototype theory as a theory of
  concepts}} {{On the adequacy of prototype theory as a theory of
  concepts}}.{\BBCQ}
\newblock
\APACjournalVolNumPages{Cognition}{9}{}{35--58}.
\newblock
\begin{APACrefDOI} \doi{10.1016/0010-0277(81)90013-5} \end{APACrefDOI}
\PrintBackRefs{\CurrentBib}

\bibitem [\protect \citeauthoryear {%
Paraboni%
, van Deemter%
\BCBL {}\ \BBA {} Masthoff%
}{%
Paraboni%
\ \protect \BOthers {.}}{%
{\protect \APACyear {2007}}%
}]{%
Paraboni2007}
\APACinsertmetastar {%
Paraboni2007}%
\begin{APACrefauthors}%
Paraboni, I.%
, van Deemter, K.%
\BCBL {}\ \BBA {} Masthoff, J.%
\end{APACrefauthors}%
\unskip\
\newblock
\APACrefYearMonthDay{2007}{}{}.
\newblock
{\BBOQ}\APACrefatitle {{Generating Referring Expressions: Making Referents Easy
  to Identify}} {{Generating Referring Expressions: Making Referents Easy to
  Identify}}.{\BBCQ}
\newblock
\APACjournalVolNumPages{Computational Linguistics}{33}{2}{229--254}.
\newblock
\begin{APACrefDOI} \doi{10.1162/coli.2007.33.2.229} \end{APACrefDOI}
\PrintBackRefs{\CurrentBib}

\bibitem [\protect \citeauthoryear {%
Pechmann%
}{%
Pechmann%
}{%
{\protect \APACyear {1989}}%
}]{%
Pechmann1989}
\APACinsertmetastar {%
Pechmann1989}%
\begin{APACrefauthors}%
Pechmann, T.%
\end{APACrefauthors}%
\unskip\
\newblock
\APACrefYearMonthDay{1989}{}{}.
\newblock
{\BBOQ}\APACrefatitle {{Incremental speech production and referential
  overspecification}} {{Incremental speech production and referential
  overspecification}}.{\BBCQ}
\newblock
\APACjournalVolNumPages{Linguistics}{27}{1}{89--110}.
\PrintBackRefs{\CurrentBib}

\bibitem [\protect \citeauthoryear {%
Pennington%
, Socher%
\BCBL {}\ \BBA {} Manning%
}{%
Pennington%
\ \protect \BOthers {.}}{%
{\protect \APACyear {2014}}%
}]{%
pennington2014glove}
\APACinsertmetastar {%
pennington2014glove}%
\begin{APACrefauthors}%
Pennington, J.%
, Socher, R.%
\BCBL {}\ \BBA {} Manning, C.%
\end{APACrefauthors}%
\unskip\
\newblock
\APACrefYearMonthDay{2014}{}{}.
\newblock
{\BBOQ}\APACrefatitle {Glove: Global vectors for word representation} {Glove:
  Global vectors for word representation}.{\BBCQ}
\newblock
\BIn{} \APACrefbtitle {Proceedings of the 2014 conference on empirical methods
  in natural language processing (EMNLP)} {Proceedings of the 2014 conference
  on empirical methods in natural language processing (emnlp)}\ (\BPGS\
  1532--1543).
\PrintBackRefs{\CurrentBib}

\bibitem [\protect \citeauthoryear {%
Peters%
\ \protect \BOthers {.}}{%
Peters%
\ \protect \BOthers {.}}{%
{\protect \APACyear {2018}}%
}]{%
peters2018deep}
\APACinsertmetastar {%
peters2018deep}%
\begin{APACrefauthors}%
Peters, M\BPBI E.%
, Neumann, M.%
, Iyyer, M.%
, Gardner, M.%
, Clark, C.%
, Lee, K.%
\BCBL {}\ \BBA {} Zettlemoyer, L.%
\end{APACrefauthors}%
\unskip\
\newblock
\APACrefYearMonthDay{2018}{}{}.
\newblock
{\BBOQ}\APACrefatitle {Deep contextualized word representations} {Deep
  contextualized word representations}.{\BBCQ}
\newblock
\APACjournalVolNumPages{arXiv preprint arXiv:1802.05365}{}{}{}.
\PrintBackRefs{\CurrentBib}

\bibitem [\protect \citeauthoryear {%
Qing%
\ \BBA {} Franke%
}{%
Qing%
\ \BBA {} Franke%
}{%
{\protect \APACyear {2015}}%
}]{%
QingFranke2015}
\APACinsertmetastar {%
QingFranke2015}%
\begin{APACrefauthors}%
Qing, C.%
\BCBT {}\ \BBA {} Franke, M.%
\end{APACrefauthors}%
\unskip\
\newblock
\APACrefYearMonthDay{2015}{}{}.
\newblock
{\BBOQ}\APACrefatitle {{Bayesian Natural Language Semantics and Pragmatics}}
  {{Bayesian Natural Language Semantics and Pragmatics}}.{\BBCQ}
\newblock
\BIn{} H.~Zeevat\ \BBA {} H.~Schmitz\ (\BEDS), \APACrefbtitle {Bayesian Natural
  Language Semantics and Pragmatics} {Bayesian natural language semantics and
  pragmatics}\ (\BPGS\ 201--220).
\newblock
\APACaddressPublisher{Cham}{Springer}.
\newblock
\begin{APACrefDOI} \doi{10.1007/978-3-319-17064-0} \end{APACrefDOI}
\PrintBackRefs{\CurrentBib}

\bibitem [\protect \citeauthoryear {%
{R Core Team}%
}{%
{R Core Team}%
}{%
{\protect \APACyear {2017}}%
}]{%
R}
\APACinsertmetastar {%
R}%
\begin{APACrefauthors}%
{R Core Team}.%
\end{APACrefauthors}%
\unskip\
\newblock
\APACrefYearMonthDay{2017}{}{}.
\newblock
{\BBOQ}\APACrefatitle {R: A Language and Environment for Statistical Computing}
  {R: A language and environment for statistical computing}{\BBCQ}\
  [\bibcomputersoftwaremanual].
\newblock
\APACaddressPublisher{Vienna, Austria}{}.
\PrintBackRefs{\CurrentBib}

\bibitem [\protect \citeauthoryear {%
Rohde%
, Seyfarth%
, Clark%
, J{\"{a}}ger%
\BCBL {}\ \BBA {} Kaufmann%
}{%
Rohde%
\ \protect \BOthers {.}}{%
{\protect \APACyear {2012}}%
}]{%
rohde2012}
\APACinsertmetastar {%
rohde2012}%
\begin{APACrefauthors}%
Rohde, H.%
, Seyfarth, S.%
, Clark, B.%
, J{\"{a}}ger, G.%
\BCBL {}\ \BBA {} Kaufmann, S.%
\end{APACrefauthors}%
\unskip\
\newblock
\APACrefYearMonthDay{2012}{}{}.
\newblock
{\BBOQ}\APACrefatitle {{Communicating with Cost-based Implicature: a
  Game-Theoretic Approach to Ambiguity}} {{Communicating with Cost-based
  Implicature: a Game-Theoretic Approach to Ambiguity}}.{\BBCQ}
\newblock
\BIn{} \APACrefbtitle {Proceedings of the 16th Workshop on the Semantics and
  Pragmatics of Dialogue} {Proceedings of the 16th workshop on the semantics
  and pragmatics of dialogue}\ (\BPGS\ 107 -- 116).
\PrintBackRefs{\CurrentBib}

\bibitem [\protect \citeauthoryear {%
Rosch%
}{%
Rosch%
}{%
{\protect \APACyear {1973}}%
}]{%
Rosch1973}
\APACinsertmetastar {%
Rosch1973}%
\begin{APACrefauthors}%
Rosch, E\BPBI H.%
\end{APACrefauthors}%
\unskip\
\newblock
\APACrefYearMonthDay{1973}{}{}.
\newblock
{\BBOQ}\APACrefatitle {{Natural categories}} {{Natural categories}}.{\BBCQ}
\newblock
\APACjournalVolNumPages{Cognitive Psychology}{4}{3}{328--350}.
\newblock
\begin{APACrefDOI} \doi{10.1016/0010-0285(73)90017-0} \end{APACrefDOI}
\PrintBackRefs{\CurrentBib}

\bibitem [\protect \citeauthoryear {%
Rosch%
, Mervis%
, Gray%
, Johnson%
\BCBL {}\ \BBA {} Boyes-Braem%
}{%
Rosch%
\ \protect \BOthers {.}}{%
{\protect \APACyear {1976}}%
}]{%
Rosch1976}
\APACinsertmetastar {%
Rosch1976}%
\begin{APACrefauthors}%
Rosch, E\BPBI H.%
, Mervis, C\BPBI B.%
, Gray, W\BPBI D.%
, Johnson, D\BPBI M.%
\BCBL {}\ \BBA {} Boyes-Braem, P.%
\end{APACrefauthors}%
\unskip\
\newblock
\APACrefYearMonthDay{1976}{}{}.
\newblock
{\BBOQ}\APACrefatitle {{Basic objects in natural categories}} {{Basic objects
  in natural categories}}.{\BBCQ}
\newblock
\APACjournalVolNumPages{Cognitive Psychology}{8}{3}{382--439}.
\newblock
\begin{APACrefDOI} \doi{10.1016/0010-0285(76)90013-X} \end{APACrefDOI}
\PrintBackRefs{\CurrentBib}

\bibitem [\protect \citeauthoryear {%
Rothschild%
\ \BBA {} Segal%
}{%
Rothschild%
\ \BBA {} Segal%
}{%
{\protect \APACyear {2009}}%
}]{%
Rothschild2009}
\APACinsertmetastar {%
Rothschild2009}%
\begin{APACrefauthors}%
Rothschild, D.%
\BCBT {}\ \BBA {} Segal, G.%
\end{APACrefauthors}%
\unskip\
\newblock
\APACrefYearMonthDay{2009}{}{}.
\newblock
{\BBOQ}\APACrefatitle {{Indexical predicates}} {{Indexical predicates}}.{\BBCQ}
\newblock
\APACjournalVolNumPages{Mind and Language}{24}{4}{467--493}.
\newblock
\begin{APACrefDOI} \doi{10.1111/j.1468-0017.2009.01371.x} \end{APACrefDOI}
\PrintBackRefs{\CurrentBib}

\bibitem [\protect \citeauthoryear {%
Rubio-Fernandez%
}{%
Rubio-Fernandez%
}{%
{\protect \APACyear {2016}}%
}]{%
rubiofernandez2016}
\APACinsertmetastar {%
rubiofernandez2016}%
\begin{APACrefauthors}%
Rubio-Fernandez, P.%
\end{APACrefauthors}%
\unskip\
\newblock
\APACrefYearMonthDay{2016}{}{}.
\newblock
{\BBOQ}\APACrefatitle {{How redundant are redundant color adjectives? An
  efficiency-based analysis of color overspecification}} {{How redundant are
  redundant color adjectives? An efficiency-based analysis of color
  overspecification}}.{\BBCQ}
\newblock
\APACjournalVolNumPages{Frontiers in Psychology}{7}{153}{}.
\newblock
\begin{APACrefDOI} \doi{10.3389/fpsyg.2016.00153} \end{APACrefDOI}
\PrintBackRefs{\CurrentBib}

\bibitem [\protect \citeauthoryear {%
Scontras%
, Degen%
\BCBL {}\ \BBA {} Goodman%
}{%
Scontras%
\ \protect \BOthers {.}}{%
{\protect \APACyear {2017}}%
}]{%
scontras2017}
\APACinsertmetastar {%
scontras2017}%
\begin{APACrefauthors}%
Scontras, G.%
, Degen, J.%
\BCBL {}\ \BBA {} Goodman, N\BPBI D.%
\end{APACrefauthors}%
\unskip\
\newblock
\APACrefYearMonthDay{2017}{}{}.
\newblock
{\BBOQ}\APACrefatitle {{Subjectivity Predicts AdjectiveOrdering Preferences}}
  {{Subjectivity Predicts AdjectiveOrdering Preferences}}.{\BBCQ}
\newblock
\APACjournalVolNumPages{Open Mind: Discoveries in Cognitive
  Science}{1}{1}{53--65}.
\newblock
\begin{APACrefDOI} \doi{10.1162/opmi} \end{APACrefDOI}
\PrintBackRefs{\CurrentBib}

\bibitem [\protect \citeauthoryear {%
Sedivy%
}{%
Sedivy%
}{%
{\protect \APACyear {2003}}%
}]{%
sedivy2003a}
\APACinsertmetastar {%
sedivy2003a}%
\begin{APACrefauthors}%
Sedivy, J\BPBI C.%
\end{APACrefauthors}%
\unskip\
\newblock
\APACrefYearMonthDay{2003}{jan}{}.
\newblock
{\BBOQ}\APACrefatitle {{Pragmatic versus form-based accounts of referential
  contrast: evidence for effects of informativity expectations.}} {{Pragmatic
  versus form-based accounts of referential contrast: evidence for effects of
  informativity expectations.}}{\BBCQ}
\newblock
\APACjournalVolNumPages{Journal of psycholinguistic research}{32}{1}{3--23}.
\PrintBackRefs{\CurrentBib}

\bibitem [\protect \citeauthoryear {%
Smith%
\ \BBA {} Levy%
}{%
Smith%
\ \BBA {} Levy%
}{%
{\protect \APACyear {2013}}%
}]{%
Smith2013}
\APACinsertmetastar {%
Smith2013}%
\begin{APACrefauthors}%
Smith, N\BPBI J.%
\BCBT {}\ \BBA {} Levy, R.%
\end{APACrefauthors}%
\unskip\
\newblock
\APACrefYearMonthDay{2013}{}{}.
\newblock
{\BBOQ}\APACrefatitle {{The effect of word predictability on reading time is
  logarithmic}} {{The effect of word predictability on reading time is
  logarithmic}}.{\BBCQ}
\newblock
\APACjournalVolNumPages{Cognition}{128}{3}{302--319}.
\newblock
\begin{APACrefDOI} \doi{10.1016/j.cognition.2013.02.013} \end{APACrefDOI}
\PrintBackRefs{\CurrentBib}

\bibitem [\protect \citeauthoryear {%
Sproat%
\ \BBA {} Shih%
}{%
Sproat%
\ \BBA {} Shih%
}{%
{\protect \APACyear {1991}}%
}]{%
sproat1991}
\APACinsertmetastar {%
sproat1991}%
\begin{APACrefauthors}%
Sproat, R.%
\BCBT {}\ \BBA {} Shih, C.%
\end{APACrefauthors}%
\unskip\
\newblock
\APACrefYearMonthDay{1991}{}{}.
\newblock
{\BBOQ}\APACrefatitle {{The cross-linguistic distribution of adjective ordering
  restrictions}} {{The cross-linguistic distribution of adjective ordering
  restrictions}}.{\BBCQ}
\newblock
\BIn{} \APACrefbtitle {Interdisciplinary approaches to language}
  {Interdisciplinary approaches to language}\ (\BPGS\ 565--593).
\newblock
\APACaddressPublisher{}{Springer Netherlands}.
\PrintBackRefs{\CurrentBib}

\bibitem [\protect \citeauthoryear {%
Szabo%
}{%
Szabo%
}{%
{\protect \APACyear {2001}}%
}]{%
Szabo2001}
\APACinsertmetastar {%
Szabo2001}%
\begin{APACrefauthors}%
Szabo, Z\BPBI G.%
\end{APACrefauthors}%
\unskip\
\newblock
\APACrefYearMonthDay{2001}{}{}.
\newblock
{\BBOQ}\APACrefatitle {{Adjectives in context}} {{Adjectives in
  context}}.{\BBCQ}
\newblock
\BIn{} I.~Kenesei\ \BBA {} R.~Harnish\ (\BEDS), \APACrefbtitle {Perspectives on
  Semantics, Pragmatics, and Discourse} {Perspectives on semantics, pragmatics,
  and discourse}\ (\BPGS\ 119--146).
\newblock
\APACaddressPublisher{Amsterdam}{John Benjamins}.
\newblock
\begin{APACrefDOI} \doi{10.1075/pbns.90.12gen} \end{APACrefDOI}
\PrintBackRefs{\CurrentBib}

\bibitem [\protect \citeauthoryear {%
Tanaka%
\ \BBA {} Taylor%
}{%
Tanaka%
\ \BBA {} Taylor%
}{%
{\protect \APACyear {1991}}%
{\protect \APACexlab {{\protect \BCnt {1}}}}}]{%
TanakaTaylor91_BasicLevelAndExpertise}
\APACinsertmetastar {%
TanakaTaylor91_BasicLevelAndExpertise}%
\begin{APACrefauthors}%
Tanaka, J\BPBI W.%
\BCBT {}\ \BBA {} Taylor, M.%
\end{APACrefauthors}%
\unskip\
\newblock
\APACrefYearMonthDay{1991{\protect \BCnt {1}}}{}{}.
\newblock
{\BBOQ}\APACrefatitle {Object categories and expertise: Is the basic level in
  the eye of the beholder?} {Object categories and expertise: Is the basic
  level in the eye of the beholder?}{\BBCQ}
\newblock
\APACjournalVolNumPages{Cognitive Psychology}{23}{3}{457--482}.
\PrintBackRefs{\CurrentBib}

\bibitem [\protect \citeauthoryear {%
Tanaka%
\ \BBA {} Taylor%
}{%
Tanaka%
\ \BBA {} Taylor%
}{%
{\protect \APACyear {1991}}%
{\protect \APACexlab {{\protect \BCnt {2}}}}}]{%
Tanaka1991}
\APACinsertmetastar {%
Tanaka1991}%
\begin{APACrefauthors}%
Tanaka, J\BPBI W.%
\BCBT {}\ \BBA {} Taylor, M.%
\end{APACrefauthors}%
\unskip\
\newblock
\APACrefYearMonthDay{1991{\protect \BCnt {2}}}{}{}.
\newblock
{\BBOQ}\APACrefatitle {{Object categories and expertise: Is the basic-level in
  the eye of the beholder?}} {{Object categories and expertise: Is the
  basic-level in the eye of the beholder?}}{\BBCQ}
\newblock
\APACjournalVolNumPages{Cognitive Psychology}{23}{}{457--482}.
\newblock
\begin{APACrefDOI} \doi{10.1016/0010-0285(91)90016-H} \end{APACrefDOI}
\PrintBackRefs{\CurrentBib}

\bibitem [\protect \citeauthoryear {%
van Gompel%
, Gatt%
, Krahmer%
\BCBL {}\ \BBA {} van Deemter%
}{%
van Gompel%
\ \protect \BOthers {.}}{%
{\protect \APACyear {2014}}%
}]{%
VanGompel2014}
\APACinsertmetastar {%
VanGompel2014}%
\begin{APACrefauthors}%
van Gompel, R\BPBI P.%
, Gatt, A.%
, Krahmer, E.%
\BCBL {}\ \BBA {} van Deemter, K.%
\end{APACrefauthors}%
\unskip\
\newblock
\APACrefYearMonthDay{2014}{}{}.
\newblock
{\BBOQ}\APACrefatitle {{Overspecification in reference: modelling size contrast
  effects}} {{Overspecification in reference: modelling size contrast
  effects}}.{\BBCQ}
\newblock
\BIn{} \APACrefbtitle {Poster presented at AMLaP 2014.} {Poster presented at
  amlap 2014.}
\newblock
\APACaddressPublisher{Edinburgh, UK}{}.
\PrintBackRefs{\CurrentBib}

\bibitem [\protect \citeauthoryear {%
van Gompel%
, van Deemter%
, Gatt%
, Snoeren%
\BCBL {}\ \BBA {} Krahmer%
}{%
van Gompel%
\ \protect \BOthers {.}}{%
{\protect \APACyear {2019}}%
}]{%
VanGompel2019}
\APACinsertmetastar {%
VanGompel2019}%
\begin{APACrefauthors}%
van Gompel, R\BPBI P.%
, van Deemter, K.%
, Gatt, A.%
, Snoeren, R.%
\BCBL {}\ \BBA {} Krahmer, E\BPBI J.%
\end{APACrefauthors}%
\unskip\
\newblock
\APACrefYearMonthDay{2019}{}{}.
\newblock
{\BBOQ}\APACrefatitle {{Conceptualization in reference production:
  Probabilistic modeling and experimental testing}} {{Conceptualization in
  reference production: Probabilistic modeling and experimental
  testing}}.{\BBCQ}
\newblock
\APACjournalVolNumPages{Psychological Review}{126}{3}{345--373}.
\newblock
\begin{APACrefDOI} \doi{10.1037/rev0000138} \end{APACrefDOI}
\PrintBackRefs{\CurrentBib}

\bibitem [\protect \citeauthoryear {%
van Miltenburg%
, Koolen%
\BCBL {}\ \BBA {} Krahmer%
}{%
van Miltenburg%
\ \protect \BOthers {.}}{%
{\protect \APACyear {2018}}%
}]{%
VanMiltenburg2018}
\APACinsertmetastar {%
VanMiltenburg2018}%
\begin{APACrefauthors}%
van Miltenburg, E.%
, Koolen, R.%
\BCBL {}\ \BBA {} Krahmer, E.%
\end{APACrefauthors}%
\unskip\
\newblock
\APACrefYearMonthDay{2018}{}{}.
\newblock
{\BBOQ}\APACrefatitle {{Varying image description tasks : spoken versus written
  descriptions}} {{Varying image description tasks : spoken versus written
  descriptions}}.{\BBCQ}
\newblock
\BIn{} \APACrefbtitle {Proceedings of the Fifth Workshop on NLP for Similar
  Languages, Varieties and Dialects} {Proceedings of the fifth workshop on nlp
  for similar languages, varieties and dialects}\ (\BPGS\ 88--100).
\PrintBackRefs{\CurrentBib}

\bibitem [\protect \citeauthoryear {%
Viethen%
, van Vessem%
, Goudbeek%
\BCBL {}\ \BBA {} Krahmer%
}{%
Viethen%
\ \protect \BOthers {.}}{%
{\protect \APACyear {2017}}%
}]{%
Viethen2017}
\APACinsertmetastar {%
Viethen2017}%
\begin{APACrefauthors}%
Viethen, J.%
, van Vessem, T.%
, Goudbeek, M.%
\BCBL {}\ \BBA {} Krahmer, E.%
\end{APACrefauthors}%
\unskip\
\newblock
\APACrefYearMonthDay{2017}{}{}.
\newblock
{\BBOQ}\APACrefatitle {{Color in Reference Production: The Role of Color
  Similarity and Color Codability}} {{Color in Reference Production: The Role
  of Color Similarity and Color Codability}}.{\BBCQ}
\newblock
\APACjournalVolNumPages{Cognitive Science}{41}{}{1493--1514}.
\newblock
\begin{APACrefDOI} \doi{10.1111/cogs.12387} \end{APACrefDOI}
\PrintBackRefs{\CurrentBib}

\bibitem [\protect \citeauthoryear {%
Walker%
}{%
Walker%
}{%
{\protect \APACyear {1993}}%
}]{%
Walker1993}
\APACinsertmetastar {%
Walker1993}%
\begin{APACrefauthors}%
Walker, M\BPBI A.%
\end{APACrefauthors}%
\unskip\
\newblock
\APACrefYear{1993}.
\unskip\
\newblock
\APACrefbtitle {{Informational Redundancy and Resource Bounds in Dialogue}}
  {{Informational Redundancy and Resource Bounds in Dialogue}}\
  \APACtypeAddressSchool {\BUPhD}{}{}.
\unskip\
\newblock
\APACaddressSchool {}{University of Pennsylvania}.
\PrintBackRefs{\CurrentBib}

\bibitem [\protect \citeauthoryear {%
Westerbeek%
, Koolen%
\BCBL {}\ \BBA {} Maes%
}{%
Westerbeek%
\ \protect \BOthers {.}}{%
{\protect \APACyear {2015}}%
}]{%
Westerbeek2015}
\APACinsertmetastar {%
Westerbeek2015}%
\begin{APACrefauthors}%
Westerbeek, H.%
, Koolen, R.%
\BCBL {}\ \BBA {} Maes, A.%
\end{APACrefauthors}%
\unskip\
\newblock
\APACrefYearMonthDay{2015}{}{}.
\newblock
{\BBOQ}\APACrefatitle {{Stored object knowledge and the production of referring
  expressions: The case of color typicality}} {{Stored object knowledge and the
  production of referring expressions: The case of color typicality}}.{\BBCQ}
\newblock
\APACjournalVolNumPages{Frontiers in Psychology}{6}{}{}.
\PrintBackRefs{\CurrentBib}

\bibitem [\protect \citeauthoryear {%
Xu%
\ \BBA {} Tenenbaum%
}{%
Xu%
\ \BBA {} Tenenbaum%
}{%
{\protect \APACyear {2007}}%
}]{%
Xu2007}
\APACinsertmetastar {%
Xu2007}%
\begin{APACrefauthors}%
Xu, F.%
\BCBT {}\ \BBA {} Tenenbaum, J\BPBI B.%
\end{APACrefauthors}%
\unskip\
\newblock
\APACrefYearMonthDay{2007}{}{}.
\newblock
{\BBOQ}\APACrefatitle {{Word learning as Bayesian inference.}} {{Word learning
  as Bayesian inference.}}{\BBCQ}
\newblock
\APACjournalVolNumPages{Psychological review}{114}{2}{245--72}.
\newblock
\begin{APACrefDOI} \doi{10.1037/0033-295X.114.2.245} \end{APACrefDOI}
\PrintBackRefs{\CurrentBib}

\bibitem [\protect \citeauthoryear {%
Zadeh%
}{%
Zadeh%
}{%
{\protect \APACyear {1965}}%
}]{%
zadeh1965fuzzy}
\APACinsertmetastar {%
zadeh1965fuzzy}%
\begin{APACrefauthors}%
Zadeh, L\BPBI A.%
\end{APACrefauthors}%
\unskip\
\newblock
\APACrefYearMonthDay{1965}{}{}.
\newblock
{\BBOQ}\APACrefatitle {Fuzzy sets} {Fuzzy sets}.{\BBCQ}
\newblock
\APACjournalVolNumPages{Information and control}{8}{3}{338--353}.
\PrintBackRefs{\CurrentBib}

\end{thebibliography}

\end{document}